\documentclass{article}

\newif\ifarxiv
\arxivtrue

\usepackage{microtype}
\usepackage{graphicx}
\usepackage{framed}
\usepackage{booktabs} 

\usepackage{hyperref}
\usepackage{listings}
\usepackage{xcolor}

\definecolor{codegreen}{rgb}{0,0.6,0}
\definecolor{codegray}{rgb}{0.5,0.5,0.5}
\definecolor{codepurple}{rgb}{0.58,0,0.82}
\definecolor{backcolour}{rgb}{0.95,0.95,0.92}
\lstdefinestyle{mystyle}{
    backgroundcolor=\color{backcolour},   
    commentstyle=\color{codegreen},
    keywordstyle=\color{magenta},
    numberstyle=\tiny\color{codegray},
    stringstyle=\color{codepurple},
    basicstyle=\ttfamily\footnotesize,
    breakatwhitespace=false,         
    breaklines=true,                 
    captionpos=b,                    
    keepspaces=true,                 
    numbers=left,                    
    numbersep=5pt,                  
    showspaces=false,                
    showstringspaces=false,
    showtabs=false,                  
    tabsize=2
}

\ifarxiv
\usepackage[accepted]{icml2020_arxiv}
\else
\usepackage{icml2021}
\fi

\usepackage[utf8]{inputenc} 
\usepackage[T1]{fontenc}    
\usepackage{hyperref}       
\hypersetup{colorlinks=true,linkcolor=black,anchorcolor=black,citecolor=black,filecolor=black,menucolor=black,runcolor=black,urlcolor=black}
\usepackage{booktabs}       
\usepackage{amsfonts}       
\usepackage{nicefrac}       
\usepackage{microtype}      
\usepackage{tcolorbox}
\usepackage{subfig}
\usepackage{algorithm}
\usepackage{graphicx}
\usepackage[percent]{overpic}
\usepackage{multirow}
\usepackage{float}
\usepackage{amsmath}
\usepackage{amssymb}
\usepackage{amsthm}
\usepackage{xcolor}
\pdfoutput=1

\newtheorem*{theoreminfo*}{Theorem (Informal)}

\newtheorem*{conjecture*}{Main point}

\usepackage{natbib}
\usepackage{xcolor}

\icmltitlerunning{How to decay your learning rate}

\begin{document}

\twocolumn[

\icmltitle{How to decay your learning rate}

\begin{icmlauthorlist}

\icmlauthor{Aitor Lewkowycz}{goo}

\end{icmlauthorlist}

\icmlaffiliation{goo}{Blueshift, Alphabet}
\icmlcorrespondingauthor{Aitor Lewkowycz}{alewkowycz@google.com}

\vskip 0.3in


]
\printAffiliationsAndNotice{}
\begin{abstract}
Complex learning rate schedules have become an integral part of deep learning. We find empirically that common fine-tuned schedules decay the learning rate after the weight norm bounces. This leads to the proposal of ABEL: an automatic scheduler which decays the learning rate by keeping track of the weight norm. ABEL's performance matches that of tuned schedules and is more robust with respect to its parameters. Through extensive experiments in vision, NLP, and RL, we show that if the weight norm does not bounce, we can simplify schedules even further with no loss in performance. In such cases, a complex schedule has similar performance to a constant learning rate with a decay at the end of training. 
\end{abstract}

\section{Introduction}

Learning rate schedules play a crucial role in modern deep learning. They were originally proposed with the goal of reducing noise to ensure the convergence of SGD in convex optimization \citep{Bottou98onlinelearning}. A variety of tuned schedules are often used, some of the most common being step-wise, linear or cosine decay. Each schedule has its own advantages and disadvantages and they all require hyperparameter tuning. Given this heterogeneity, it would be desirable to have a coherent picture of when schedules are useful and to come up with good schedules with minimal tuning.  

While we do not expect dependence on the initial learning rate in convex optimization, large learning rates behave quite different from small learning rates in deep learning \citep{li2020explaining,lewkowycz2020large}. We expect the situation to be similar for learning rate schedules, the non-convex landscape makes it desirable to reduce the learning rate as we evolve our models. The goal of this paper is to study empirically (a) in which situations schedules are beneficial and (b) when during training one should decay the learning rate. Given that stochastic gradients are used in deep learning, we will use a \emph{simple} schedule as our baseline: a constant learning rate with one decay close to the end of training. Training with this smaller learning rate for a short time is expected to reduce the noise without letting the model explore the landscape too much. This is corroborated by the fact that the minimum test error often occurs almost immediately after decaying the learning rate. Part of the paper focuses on comparing the \emph{simple} schedule with standard \emph{complex} schedules used in the literature, studying the situations in which these complex schedules are advantageous. We find that complex schedules are considerably helpful whenever the weight norm bounces, which happens often in the usual, optimal setups. In the presence of a bouncing weight norm, we propose an automatic scheduler which performs as well as fine tuned schedules.

\subsection{Our contribution}
\label{sec:our_contr}

The goal of the paper is to study the benefits of learning rate schedules and when the learning rate should be decayed. We focus on the dynamics of the weight norm \ifarxiv  , defined as the sum over the squared $L_2$-norm of the weight in each layer \ $|w|^2 \equiv \sum_{\text{layer}} ||w(\text{layer})||^2_{2} $. \fi

\begin{figure}[]
  \centering
  \subfloat[Resnet-50 on ImageNet]{
    \includegraphics[width=0.24\textwidth]{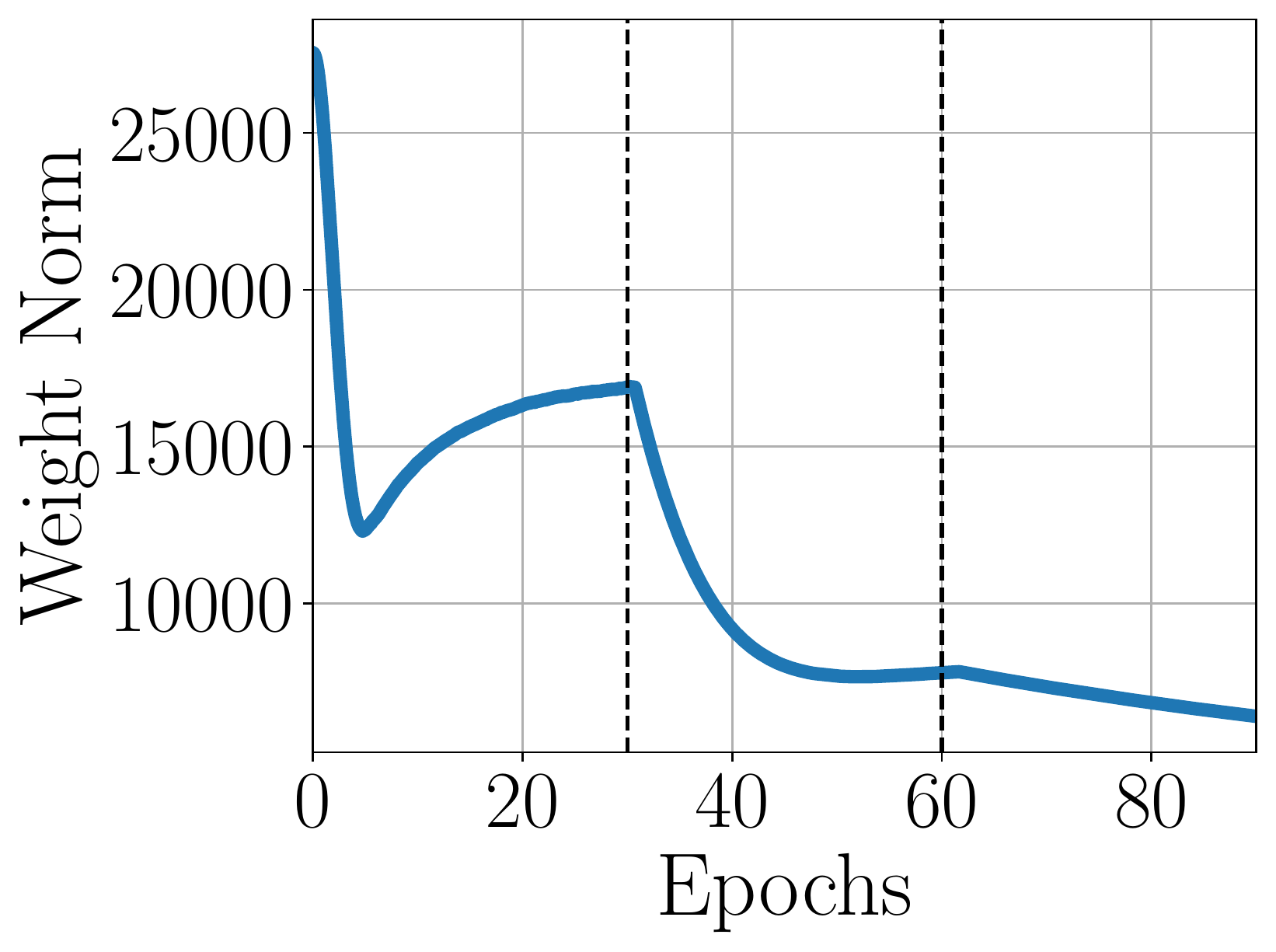}
    \label{fig:fig1a}
  }
  \subfloat[WRN28-10 on CIFAR-100]{
    \includegraphics[width=0.24\textwidth]{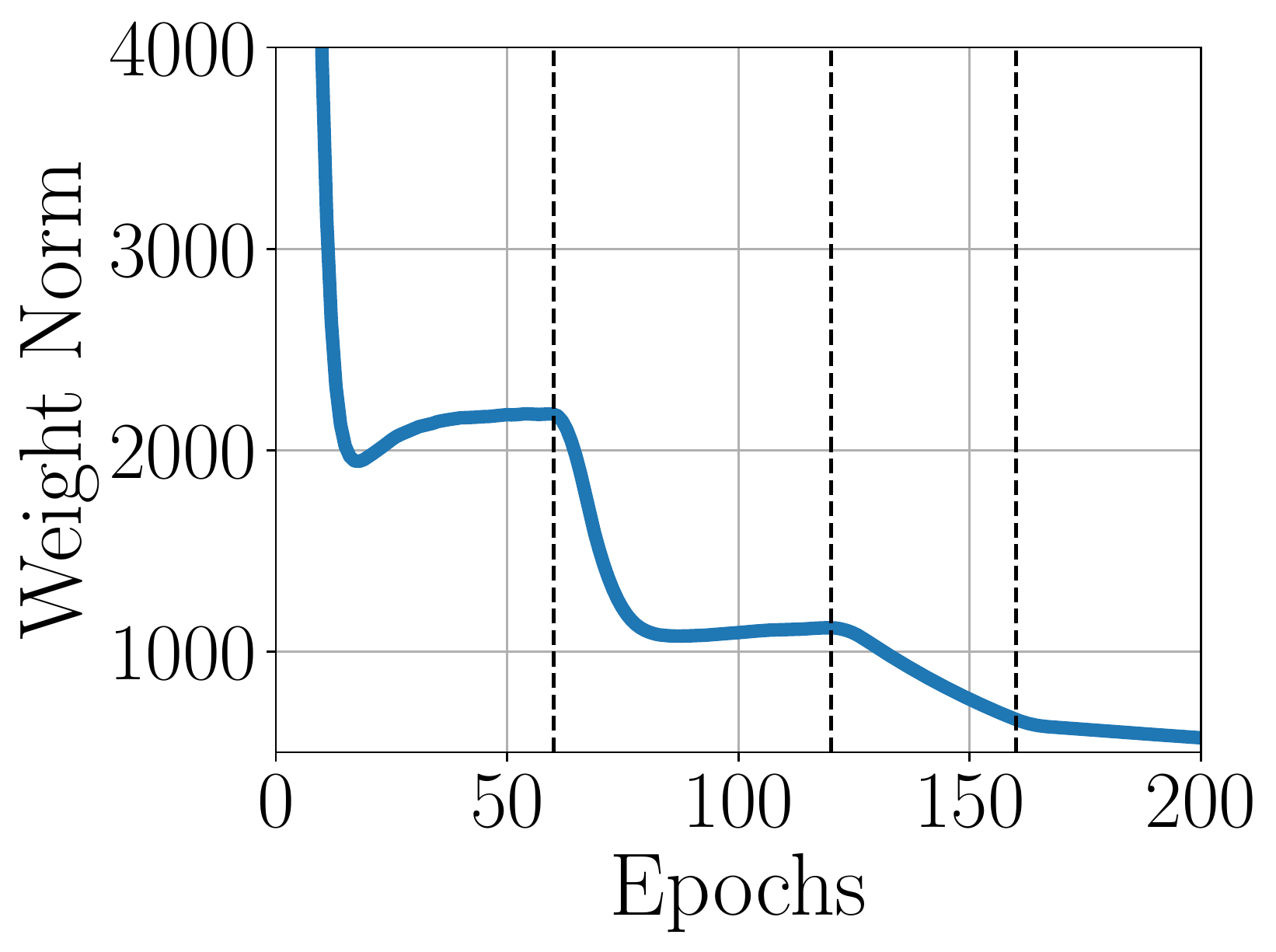}
    \label{fig:fig1b}
  }

  \caption{ Evolution of the weight norm when training with step-wise decay (decay times marked by black dashed lines). The learning rate is decayed after the weight norm bounces, towards its convergence. Models were evolved in optimal settings whose tuning did not use the weight norm as input.
  }
  \label{fig:fig1}
\end{figure}

We make the following observations based on extensive empirical results which include Vision, NLP and RL tasks.
\begin{enumerate}
\item We observe that tuned step-wise schedules decay the learning rate after the weight norm bounces\footnote{We define bouncing by the monotonic decrease of the weight norm followed by a monotonic increase which occurs for a fixed learning rate as can be seen in figure \ref{fig:fig1}.}, when the weight norm starts to converge. Towards the end of training, a last decay decreases the noise. See figure \ref{fig:fig1}. 
\item We propose an \emph{Automatic, Bouncing into Equilibration Learning rate scheduler} (ABEL). ABEL is competitive with fine tuned schedules and needs less tuning (see table \ref{table:table1} and discussion in section \ref{sec:abel}). 
\item A bouncing weight norm seems necessary for non-trivial learning rate schedules to  outperform the simple decay baseline. $L_2$ regularization is required for the weight norm to bounce and in its absence (which is common in NLP and RL) we don't see a benefit from complex schedules.  This is explored in detail in section \ref{sec:nobounce} and the results are summarized in table \ref{table:nobounce}.
\end{enumerate}

\paragraph{The origin of weight bouncing.} There is a simple heuristic for why the weight norm bounces. Without $L_2$ regularization, the weight norm usually increases for learning rate values used in practice. In the presence of $L_2$ regularization, we expect the weight norm to decrease initially. As the weight norm decrease slows down, the natural tendency for the weight norm to increase in the absence of regularization will eventually dominate. This is explained in more detail in section \ref{sec:understand}.

\paragraph{Weight bouncing and performance.} Generally speaking, weight bouncing occurs when we have non-zero $L_2$ regularization and large enough learning rates. While $L_2$ regularization is crucial in vision tasks, it is not found to be that beneficial in NLP or Reinforcement Learning tasks (for example \citet{vaswani2017attention} does not use $L_2$). If the weight norm does not bounce, ABEL yields the "simple" learning rate schedule that we expect naively from the noise reduction picture: decay the learning rate once towards the end of training. We confirm that, in the absence of bouncing, such a simple schedule is competitive with more complicated ones across a variety of tasks and architectures,  see table \ref{table:nobounce}. 
We also see that the well known advantage of momentum compared with Adam for image classification in ImageNet (see \citet{adagraft} for example) seems to disappear in the absence of bouncing, when we turn off $L_2$ regularization.
Weight norm bouncing thus seems empirically a necessary condition for non-trivial schedules to provide a benefit, but it is not sufficient: we observe that when the datasets are easy enough that simple schedules can get zero training error, schedules do not make a difference.

\subsection{Related works}
We do not know of any explicit discussion of weight bouncing in the literature. The dynamics of deep networks with $L_2$ regularization has drawn recent attention, see for example \citet{vanlaarhoven2017l2,lewkowycz2020training,li2020reconciling,wan2020spherical,kunin2020neural}.   The recent paper \citet{wan2020spherical} observes that the weight norm equilibration is a dynamical process (the weight norm still changes even if the equilibrium conditions are approximately satisfied) which happens soon after the bounce. 

The classic justification for schedules comes from reducing the noise in a quadratic potential \citep{Bottou98onlinelearning}. Different schedules do not provide an advantage in convex optimization unless there is a substantial mismatch between train and test landscape  \citep{nakkiran2020learning}, however this is not the effect that we are observing in our setup: when schedules are beneficial, their training performance is substantially different, see for example figure \ref{fig:noaug}. The work of \citet{li2020explaining} could be helpful to understand better the theory behind our phenomena, although it is not clear to us how their mechanism can generalize to multiple decays (or other complex schedules).
There has been lots of empirical work trying to learn schedules/optimizers see for example \citet{maclaurin2015gradientbased,li2016learning,li2017metasgd,wichrowska2017learned,rolinek2018l4,qi2020mlrsnet}. Our approach does not have an outer loop: the learning rate is decayed depending on the weight norm, which is conceptually similar to the \texttt{ReduceLROnPlateau} scheduler, where the learning rate is decayed when the loss plateaus which is present in most deep learning libraries. However, \texttt{ReduceLROnPlateau} does not perform well across our tasks. A couple of papers which thoroughly compare the performance of learning rate schedules are  \citet{shallue2019measuring,kaplan2020scaling}.

\section{An automatic learning rate schedule based on the weight norm}
\label{sec:abel}
\subsection*{ABEL and its motivation}

From the two setups in figure \ref{fig:fig1} it seems that optimal schedules tend to decay the learning rate after bouncing, when the weight norm growth slows down. We can use this observation to propose ABEL (Automatic Bouncing into Equilibration Learning rate scheduler): a schedule which implements this behaviour automatically, see algorithm \ref{alg:ABEL}. In words, we keep track of the changes in weight norm between subsequent epochs, $\Delta |w_t|^2 \equiv |w_t|^2-|w_{t-1}|^2$. When the sign of $\Delta |w_t|^2$ flips, it necessarily means that it has gone through a local minimum: because $\Delta |{w}_t|^2<0$ initially) if $\Delta |w_{t+1}|^2 \cdot \Delta |w_{t}|^2<0,\Delta |w_t|^2<0$, then $|w_t|$ is  a minimum: $|w_{t+1}|>|w_t|<|w_{t-1}|$. After this, the weight norm grows and slows down until at some point $\Delta |w_t|^2$ is noise dominated. In this regime, $\Delta |w_t|^2$ will become negative, which we will take as our decaying condition. In order to reduce SGD noise, near the end of training we decay it one last time. In practice we do this last decay at around $85\%$ of the total training time and as we can see in the SM \ref{sec:last_decay_epoch}, this particular value does not really matter.
\begin{algorithm}[tb]
   \caption{ABEL Scheduler}
   \label{alg:ABEL}
\begin{algorithmic}
   \IF{$(|w_t|^2-|w_{t-1}|^2) \cdot  (|w_{t-1}|^2-|w_{t-2}|^2) < 0 \ $}
   \IF{reached\_minimum}
   \STATE{learning\_rate = decay\_factor $\cdot$ learning\_rate}
   \STATE{reached\_minimum=False}
   \ELSE
   \STATE{reached\_minimum=True}
   \ENDIF
   \ENDIF

   \IF{t = last\_decay\_epoch}
   \STATE{learning\_rate = decay\_factor  $\cdot$ learning\_rate}
   \ENDIF

\end{algorithmic}
\end{algorithm}

Algorithm \ref{alg:ABEL} is an implementation of the idea with the base learning rate and the decay factor as the main hyperparameters. While alternative implementations could be more explicit about the weight norm slowing down after reaching the minimum, they would likely require more hyperparameters. 

We have decided to focus on the total weight norm, but one might ask what happens with the layer-wise weight norm. In SM \ref{sec:different_layers}, we study the evolution of the weight norm in different layers. We focus on the $10$ layers which contribute the most to the weight norm (these layers account for $50\%$ of the weight norm). We see that most layers exhibit the same bouncing plus slowing down pattern as the total weight norm and this happens at roughly the same time scale. 

\subsection*{Performance comparison across setups.}

\begin{figure}[ht]
  \centering
  \subfloat[Resnet-50 on ImageNet]{
    \includegraphics[width=0.24\textwidth]{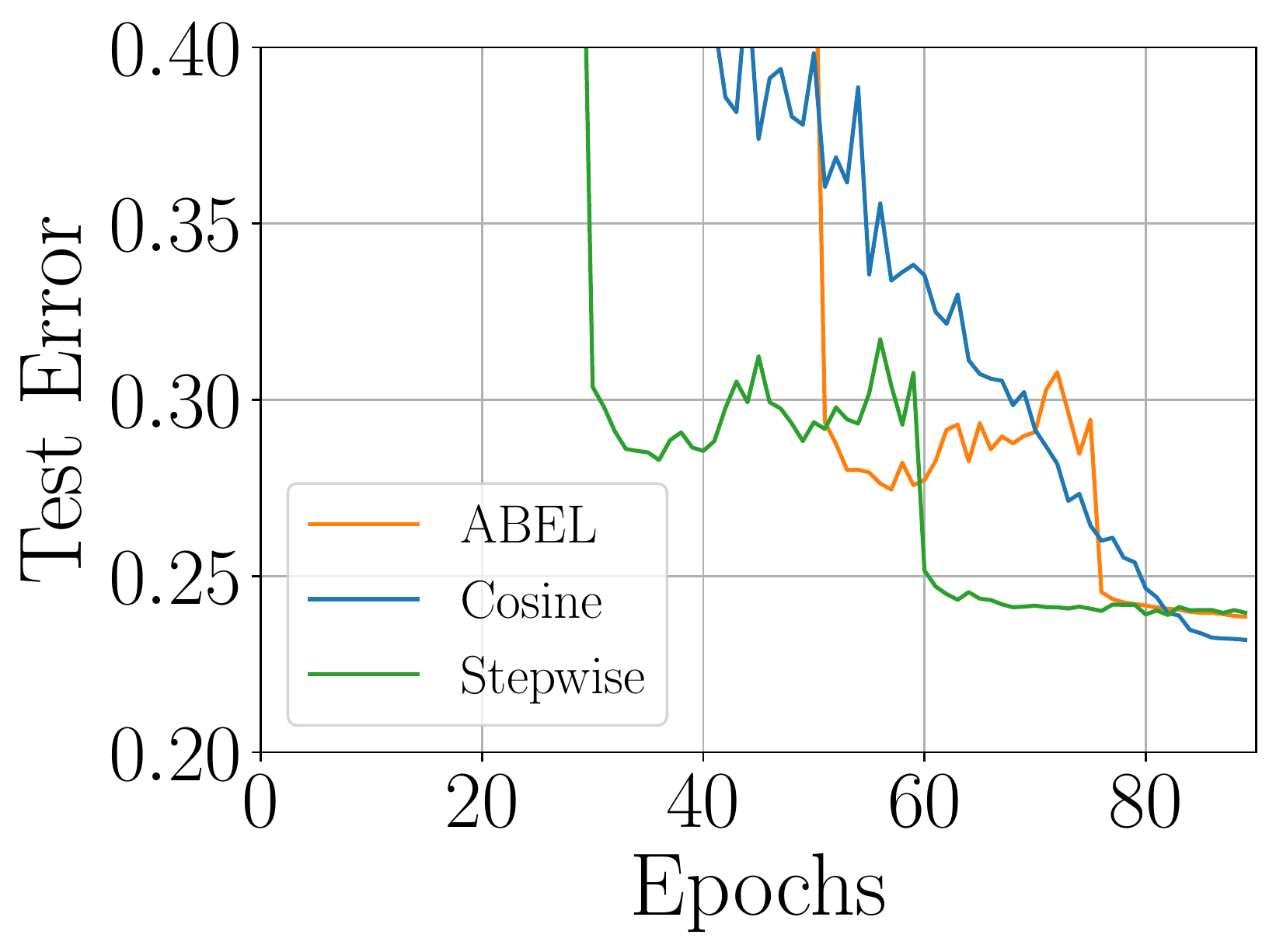}
    \label{fig:fig2a}
  }
  \subfloat[WRN28-10 on CIFAR-100]{
    \includegraphics[width=0.24\textwidth]{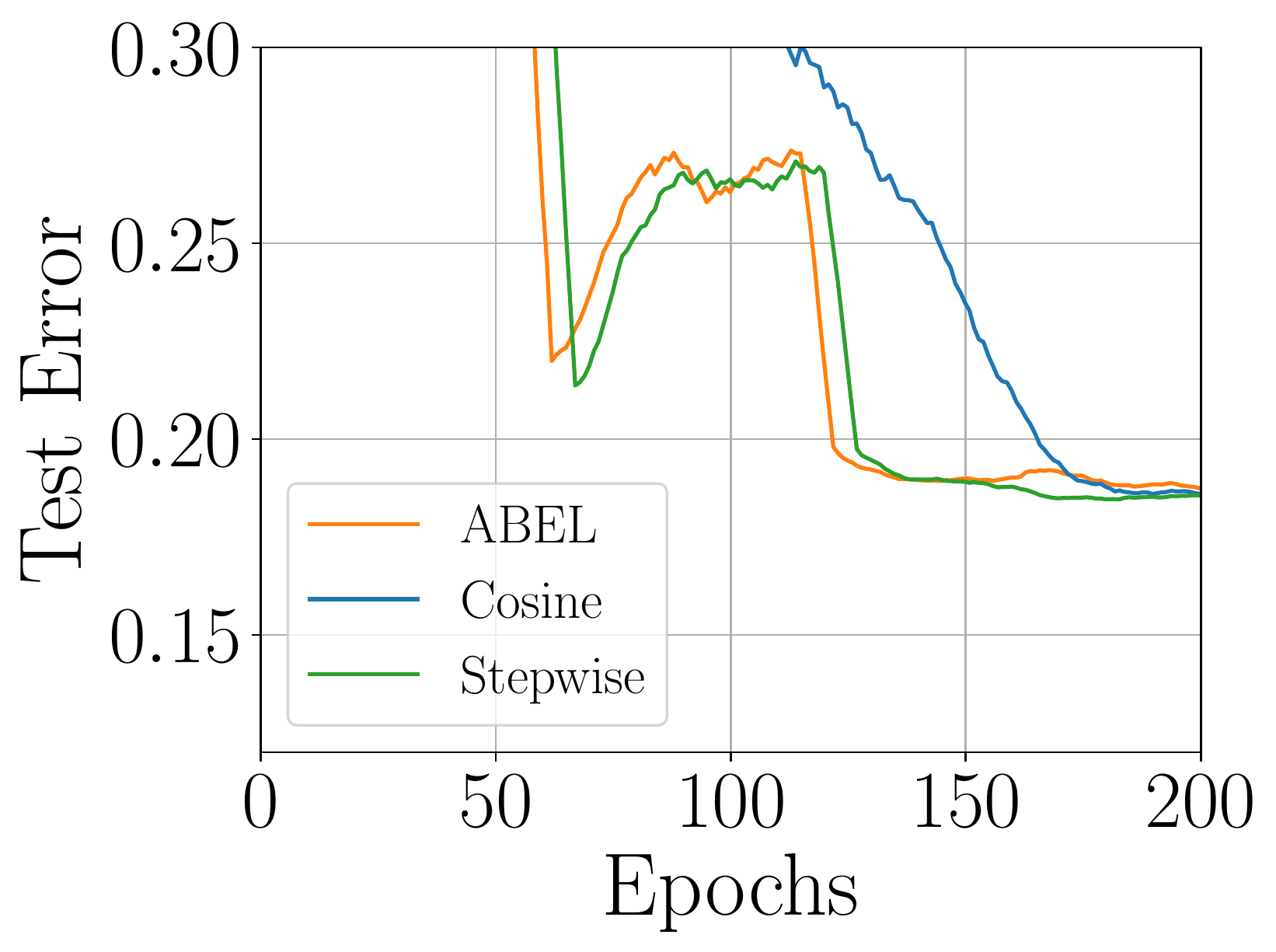}
    \label{fig:fig2d}
  } 
  \\
  \subfloat[Resnet-50 on ImageNet]{
    \includegraphics[width=0.24 \textwidth]{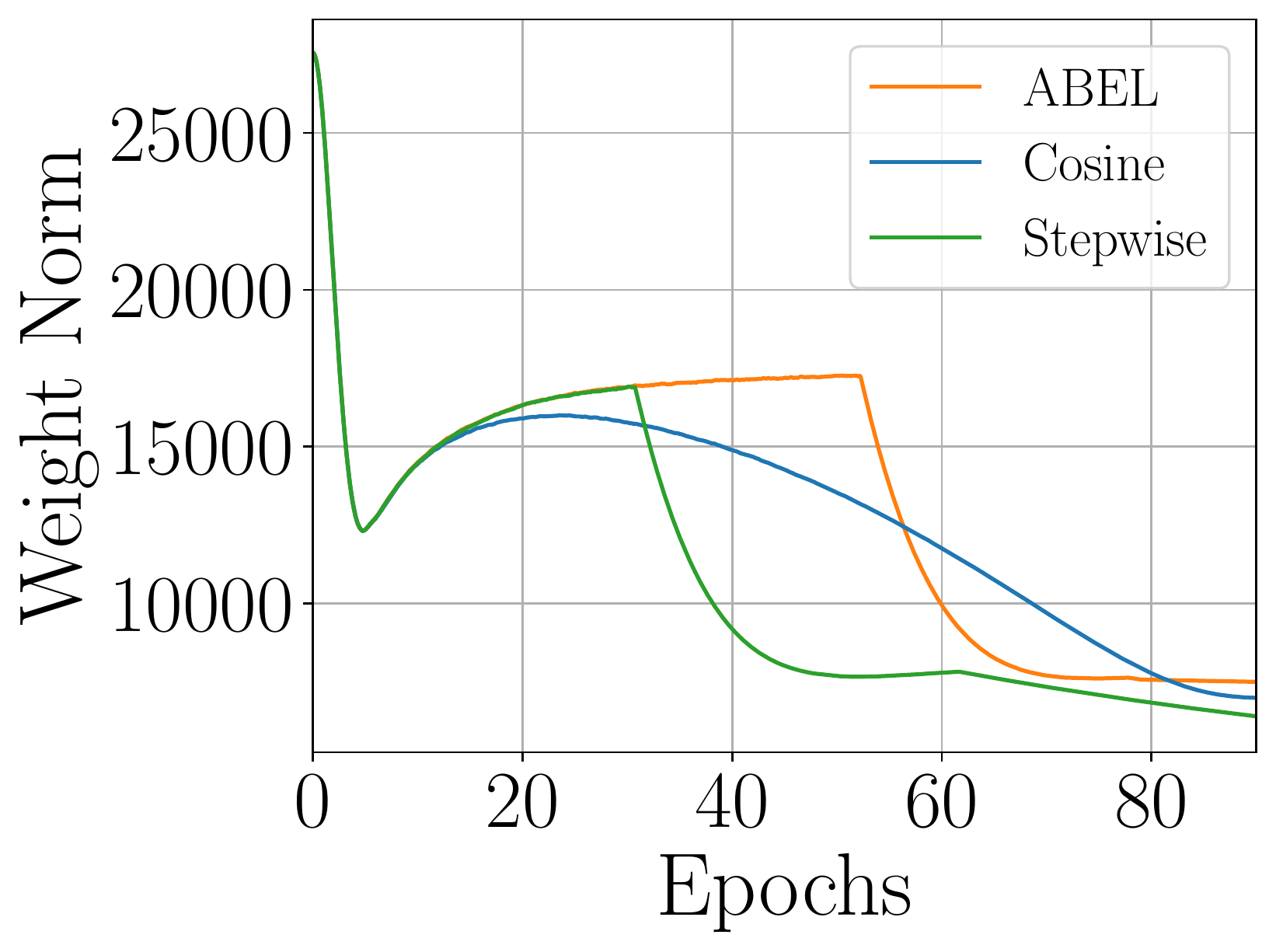}
    \label{fig:fig2b}
  }
  \subfloat[WRN28-10 on CIFAR-100]{
    \includegraphics[width=0.24 \textwidth]{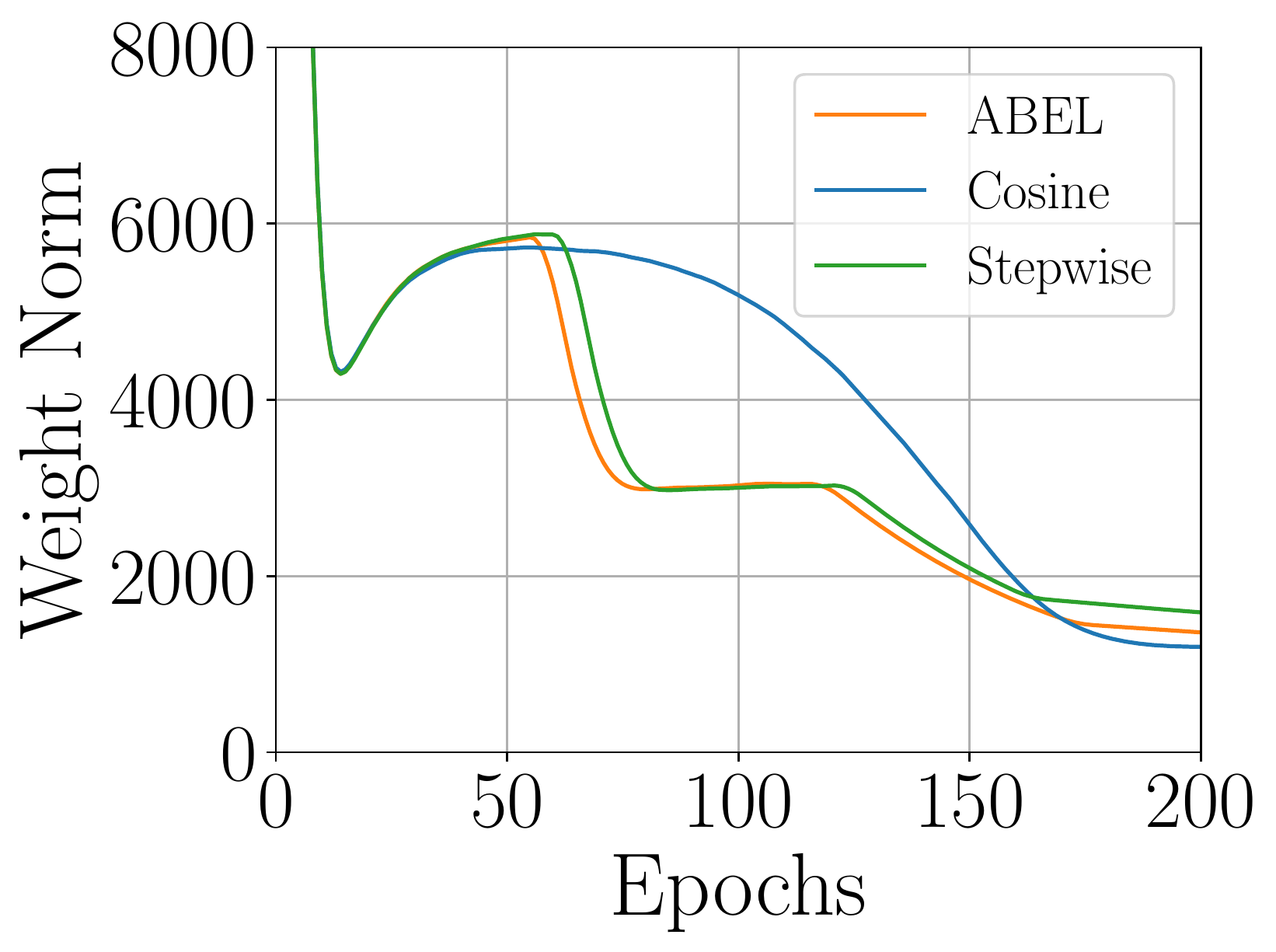}
    \label{fig:fig2e}
  }
  \\
      \subfloat[Resnet-50 on ImageNet]{
    \includegraphics[width=0.24 \textwidth]{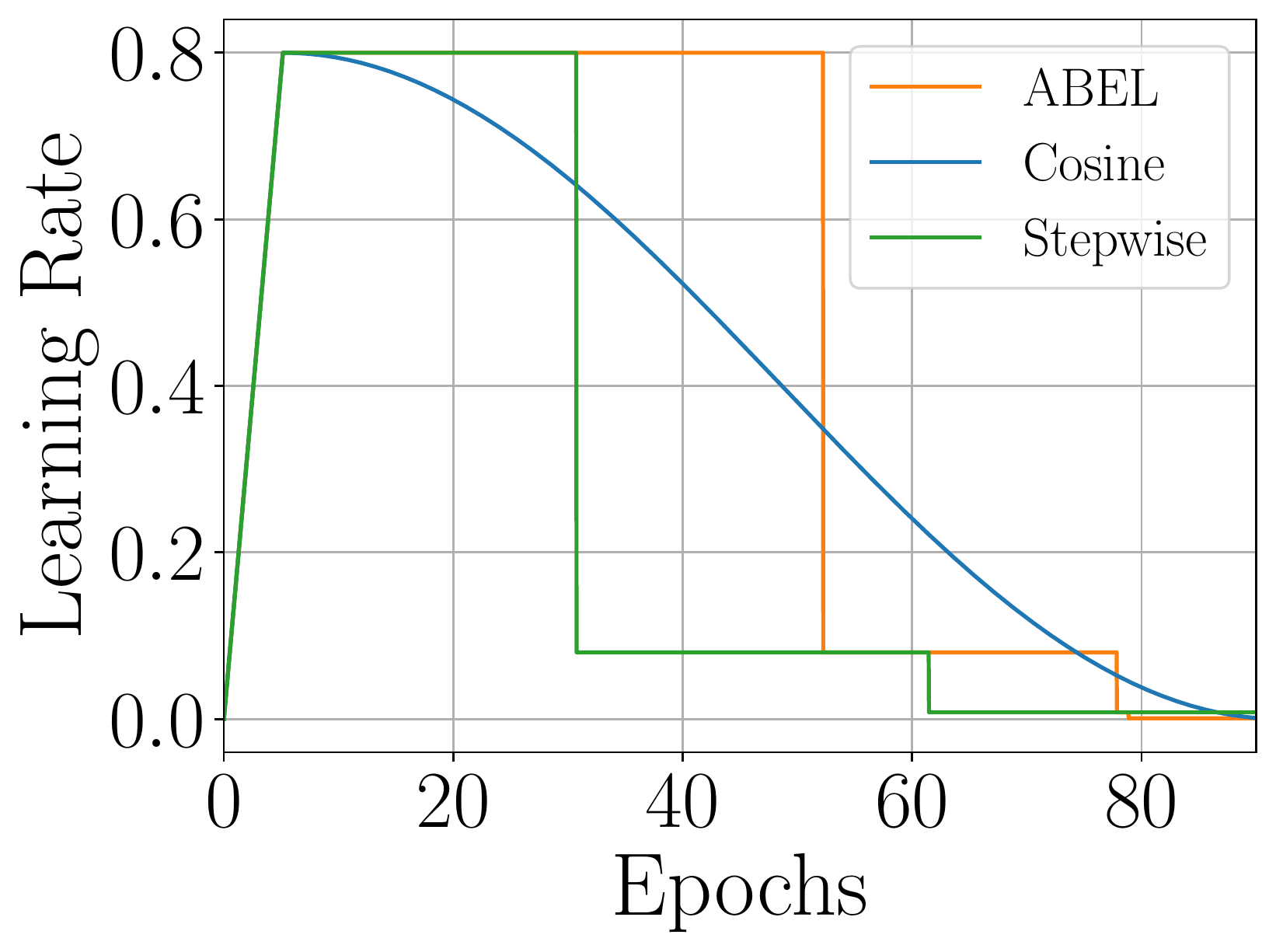}
    \label{fig:fig2c}
  }
  \subfloat[WRN28-10 on CIFAR-100]{
    \includegraphics[width=0.24 \textwidth]{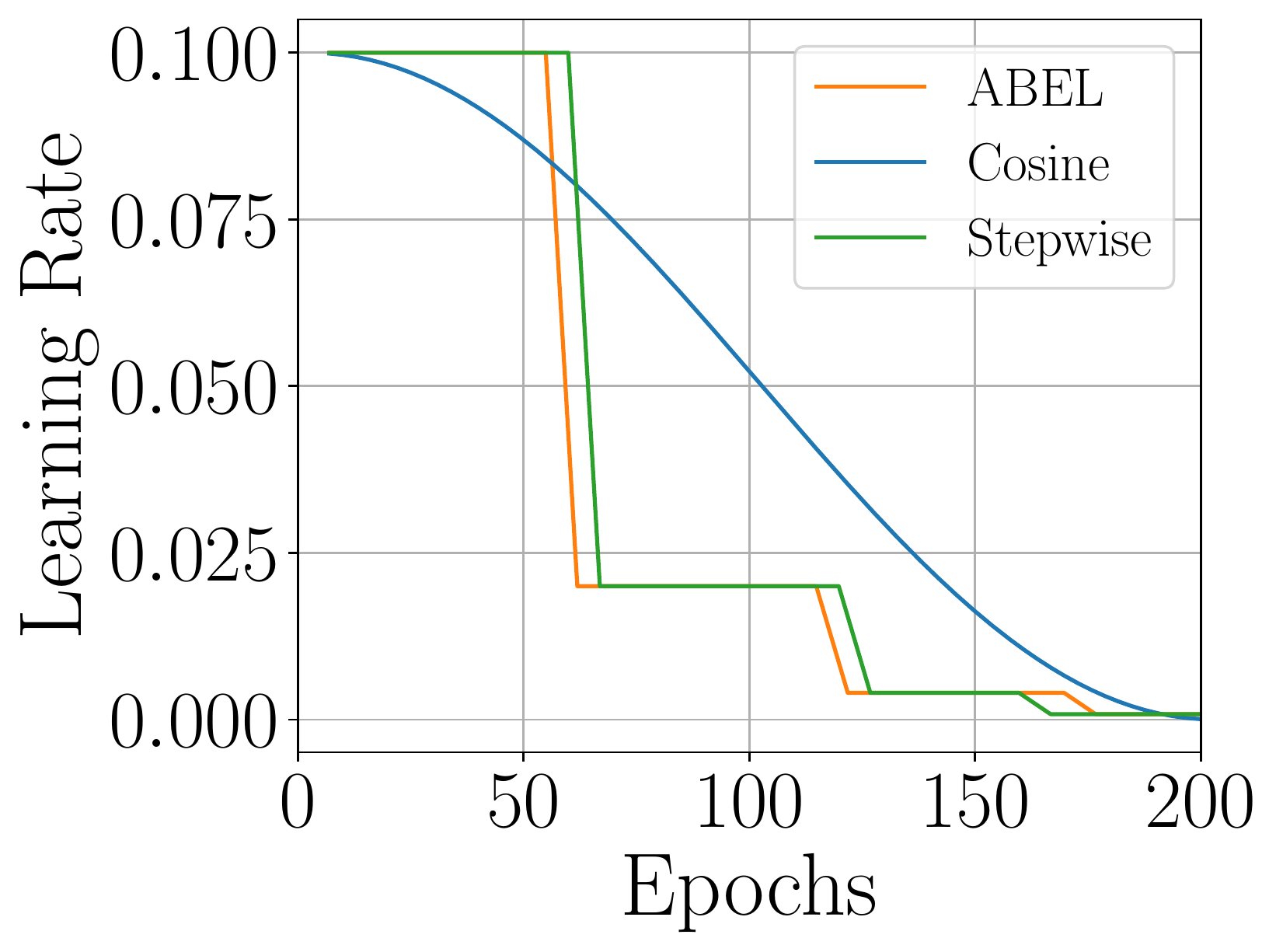}
    \label{fig:fig2f}
    
  }

  \caption{Training curves of two experiments from table \ref{table:table1}.  }
  \label{fig:abel}
\end{figure}

We have run a variety of experiments comparing learning rate schedules with ABEL, see table \ref{table:table1} for a summary and figure \ref{fig:abel} for some selected training curves ( rest of the training curves are in SM \ref{SM:training}). We use ABEL without hyperparameter tuning it: we are plugging the base learning rate and the decay factor of the reference step-wise schedule (these reference decay factors are $0.2$ for CIFAR and $0.1$ for other datasets). We see that ABEL is competitive with existing fine tuned learning rate schedules and slightly outperforms step-wise decay on ImageNet. Cosine often beats step-wise schedules, however as we will discuss shortly, such decay has several drawbacks.  

\newcommand\x{2.2cm}
\newcommand\xb{1.25cm}
\newcommand\y{1.7cm}
\newcommand\z{2.7cm}

\begin{table}

{ \centering
\begin{tabular}{ |l|l||r|r|r|  }
\hline
 \multicolumn{2}{|c||}{Setup} &
 \multicolumn{3}{c|}{Test error} \\
 \hline
 Dataset & Architecture & Step-wise & ABEL & Cosine    \\ \hline
 ImageNet & Resnet-50    & 24.0  & 23.8 & 23.2  \\
CIFAR-10 & WRN 28-10   & 3.7  & 3.8 & 3.5 \\
CIFAR-10 & VGG-16    & -  & 7.1 & 6.9 \\
CIFAR-100 & WRN 28-10    & 18.5  & 18.7 & 18.4  \\
CIFAR-100 & PyramidNet    & -  & 10.8 & 10.8 \\
SVHN & WRN 16-8    & 1.77  & 1.79 & 1.89  \\
 \hline
 
\end{tabular}

}
\hspace*{0.5cm}
\caption{Comparison of test error at the end of training for different setups and learning rate schedules. We see that ABEL has very similar performance to the fine tuned step-wise schedule without the need to tune when to decay. ABEL uses the baseline values of learning rates and decay factors and we have not fine tuned these. The cells denoted by - refer to setups for which we do not have reference step-wise decays. The experimental details can be found in the SM. }
\label{table:table1}
\end{table}

\subsection*{Robustness of ABEL}
\begin{figure}[ht!]

  \subfloat[]{
    \includegraphics[width=0.24\textwidth]{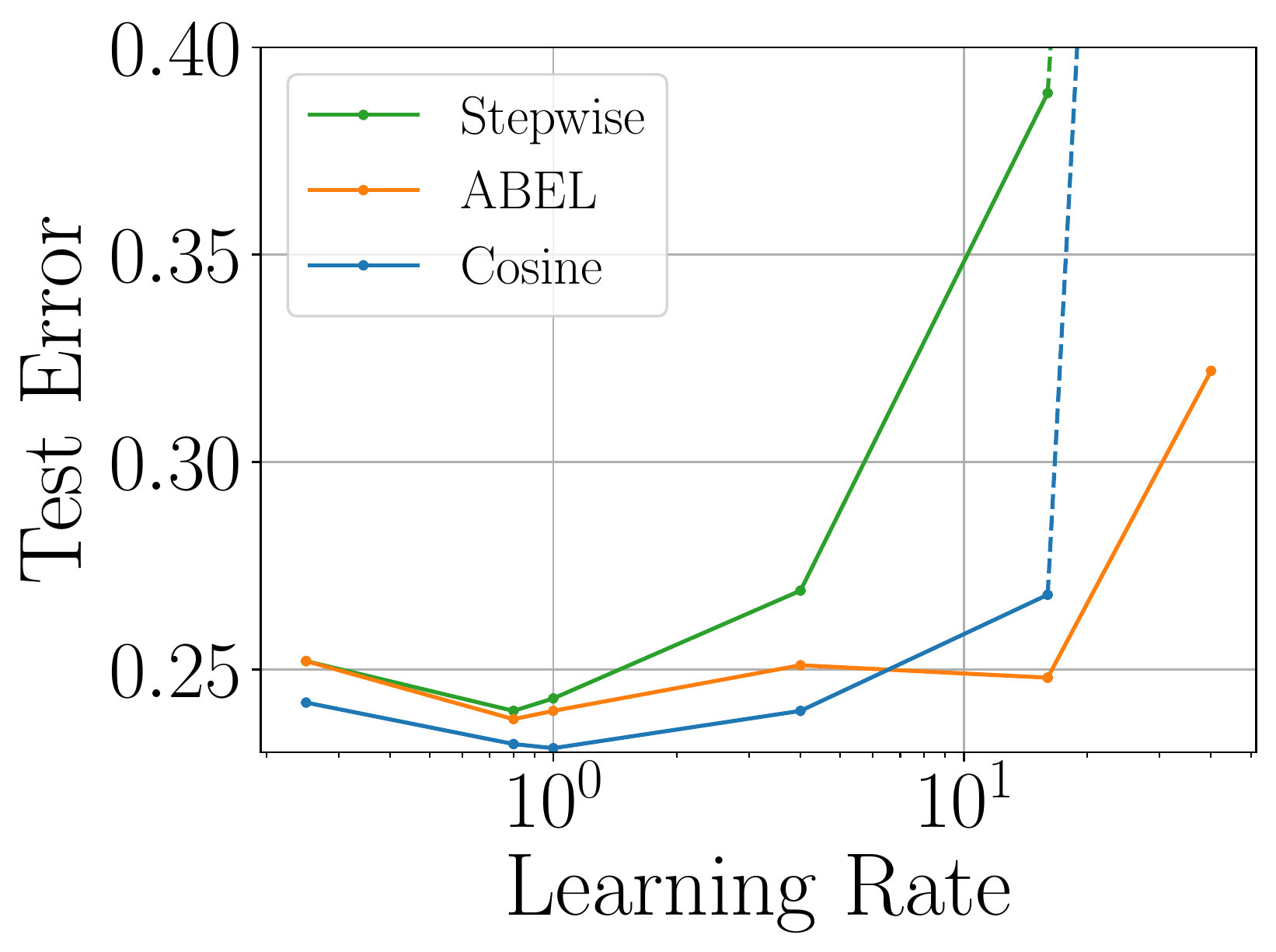}
    \label{fig:fig3a}
  }
  \subfloat[]{
    \includegraphics[width=0.24 \textwidth]{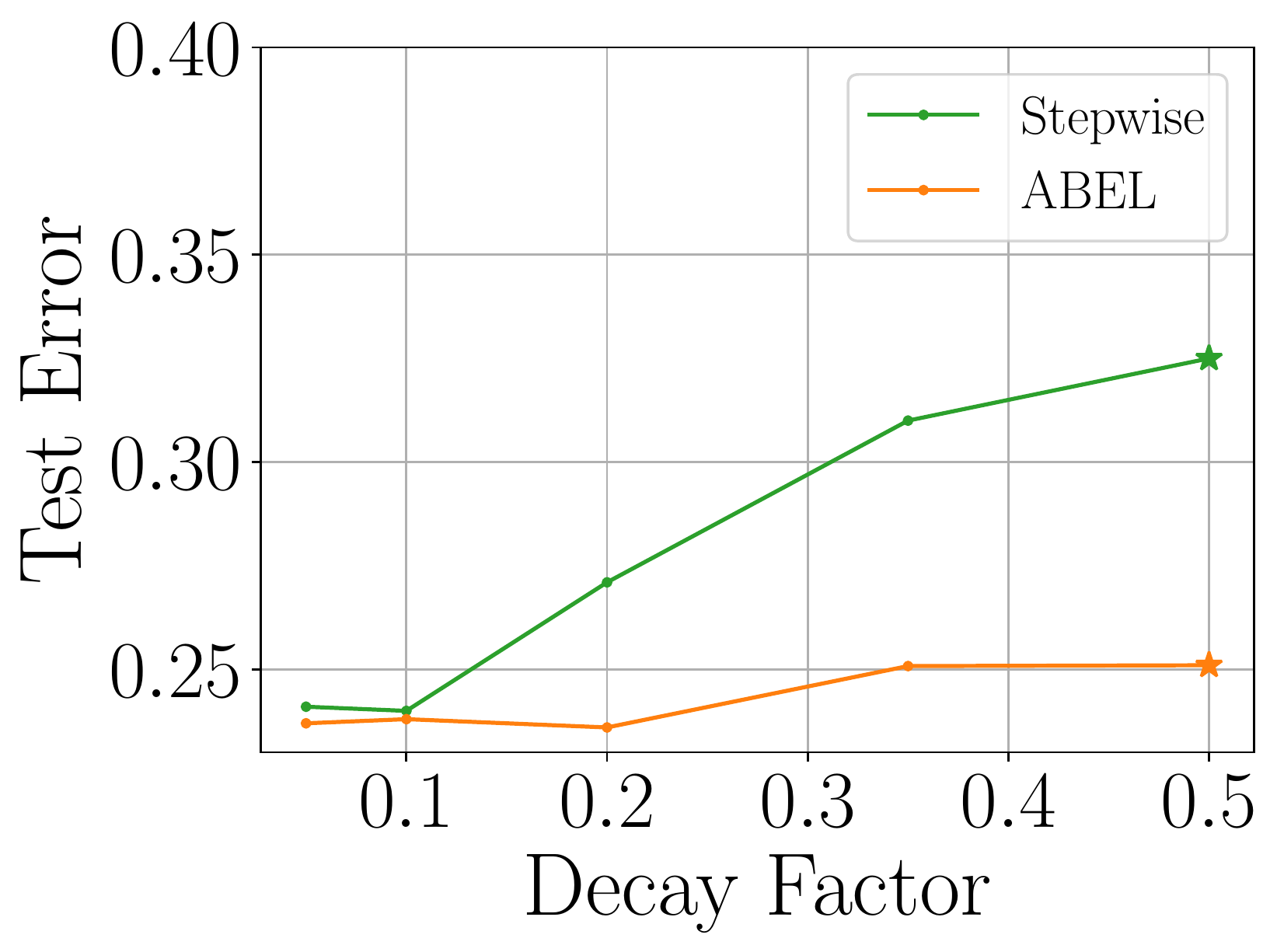}
    \label{fig:fig3b}
  }

  \caption{ResNet-50 trained on ImageNet for different learning rates and decay factors. (a) ABEL beats others schedules when using non-optimal learning rates. At learning rate $40$, only ABEL converges. (b) ABEL is robust with respect to changes in the decay factor, its performance does not depend too much on the decay factor because it adjusts the number of decays accordingly. Note: when the decay factor is $0.5$, $90$ epochs is too short of a time for ABEL to adapt properly: the weight norm is still is bouncing, so we evolved that point for $120$ epochs. If evolved for $90$ epochs like the other points, the standard decay performance does not change much, but ABEL has an error closer to $30\%$. }
  \label{fig:comparison}
\end{figure}

ABEL is quite robust with respect to the learning rate and the decay factor. Since it depends implicitly on the natural time scales of the system, it will adapt to when to decay the learning rate. We can illustrate this by repeating the ImageNet experiment with different base learning rates or decay factors. The results are shown in figure \ref{fig:comparison}. 

We would like to highlight the mild dependence of performance in the learning rate: if the learning rate is too high, the weight norm will bounce faster and ABEL will adapt to this by quickly decaying the learning rate. This can be seen quite clearly in the learning rate $=16$ training curves, see SM \ref{SM:largelr}.

ABEL also has the `last\_decay\_epoch' hyperparameter, which determines when to perform the last decay in order to reduce noise. Performance depends very weakly on this hyperparameter (see SM more) and for all setups in table \ref{table:table1} we have chosen it to be at $85\%$ of the total training time. The most natural way to think about this would be to run ABEL for a fixed amount of time and after decay the learning rate for a small number of epochs in order to get the performance with less SGD noise.

\subsection*{Comparison of ABEL with other schedules}

It is very natural to compare ABEL with step-wise decay. Step-wise decay is complicated to use in new settings because on top of the base learning rate and the decay factor, one has to determine when to decay the learning rate. ABEL, takes care of the `when' automatically without hurting performance. Because when to decay depends strongly on the system and its current hyperparameters, ABEL is much more robust to the choices of base learning rate and decay factor. 

A second class of schedules are those which depend explicitly in the number of training epochs ($T$), like cosine or linear decay. This strongly determines the decay profile: with cosine decay, the learning rate will not decay by a factor of $10$ with respect to its initial value until $93\%$ of training!  Having $T$ as a determining hyperparameter is problematic:
it takes a long time  for these schedules to have comparable error rates to step-wise decays, as can be seen in figures \ref{fig:abel}, \ref{fig:more_training_curves}. This implies that until very late in training one can not tell whether $T$ is too short, in which case there is no straightforward way to resume training (if we want to evolve the model with the same decay for a longer time, we have to start training from the beginning). This is part of the reason why large models in NLP and vision use schedules which can be easily resumed like rsqrt decay \citep{vaswani2017attention}, "clipped" cosine decay  \citep{kaplan2020scaling,brown2020language} or exponential decay \citep{tan2020efficientnet}. In contrast, for ABEL the learning rate at any given time is independent of the total training budget ( while there is the last\_decay\_epoch parameter, it can easily be evolved for longer if we load the model before the last decay).

We have decided to compare ABEL with the previous two schedules because they are the most commonly used ones. There are a lot of automatic/learnt learning rate schedules (or optimizers), see \citet{maclaurin2015gradientbased,li2016learning,li2017metasgd,wichrowska2017learned,yaida2018fluctuationdissipation,rolinek2018l4,qi2020mlrsnet} and to our knowledge most of them require either significant change in the code (like the addition of non-trivial measurements) or outer loops and also add hyperparameter of their own, so these are never completely hyperparameter free. Compared with these algorithms ABEL is simple, interpretable (it can be easily compared with fine tuned step-wise decays) and performs as well as tuned schedules. It is also quite robust compared with other automatic methods because it relies in the weight norm which is mostly noise free through training (compared with other batched quantities like gradients or losses).

An algorithm similar in simplicity and interpretability is \texttt{ReduceLROnPlateau} which is one of the basic optimizers of PyTorch or TensorFlow and decays the learning rate whenever the loss equilibrates. We train  a Resnet-50 ImageNet model and a WRN 28-10 CIFAR-10 model with this algorithm , see SM \ref{SM:losseq} for details. We use the default hyperparameters and for the ImageNet experiment, the learning rate does not decay at all, yielding a test error of $47.4$. For CIFAR-10, \texttt{ReduceLROnPlateau} does fairly well, test error of $3.9$, however the learning rate decays without bound rather fast. These two experiments suggest that  \texttt{ReduceLROnPlateau} can not really compete with the schedules described above.

\subsection*{ABEL does not require a fixed train budget}

From the empirical studies, the drop in the test error after decaying the learning rate is upper bounded by the previous drops in the test error, a reason for this is that this drop can be attributed to a reduction of the SGD noise and smaller learning rates have less SGD noise. This provides an automatic way of prescribing the train budget: if the improvement of accuracy after a decay is smaller than some threshold, exit training by after a small number of epochs (to process the last decay). This approach does not have a manual decay at the end of training. Such a training setup would not possible for cosine/linear decay by construction since they depend on the training budget. This seems hard for step-wise decay since there is no way to predetermine how to decay the learning rate automatically.

\section{Schedules and performance in the absence of a bouncing weight norm} 
\label{sec:nobounce}

In this section, we study settings where the weight norm does not bounce to understand the impact of learning rate schedules on performance. Setups without $L_2$ regularization are the most common situation with no bouncing, these setups often present a monotonically increasing weight norm. It is not clear to us what characteristics of a task make $L_2$ regularization beneficial but it seems that Vision benefits considerably more from it than NLP or RL.

We conduct an extensive set of experiments in the realms of vision, NLP and RL and the results are summarized in table \ref{table:nobounce}. In these experiments, we compare complex learning rate schedules with a simple schedule where the model is evolved with a constant learning rate and decayed once towards the end of training. This simple schedule mainly reduces noise: the error decreases considerably immediately after decaying and it does not change much afterwards (it often increases). Across these experiments, we observe that complicated learning rate schedules are not significantly better than the simple ones.  For a couple of tasks (like ALBERT finetuning or WRN on CIFAR-100), complex schedules are slightly better ($\sim 0.3\%$) than the simple decay but this small advantage is nothing compared with the substantial advantage that schedules have in vision tasks with $L_2$. Another situation where there is no bouncing weight norm is for small learning rates, for example VGG-16 with learning rate $0.01$, in such case there is also no benefit from using complex schedules, see SM \ref{SM:small_lr} for more details. Note that in this paper we are using $L_2$ regularization and weight decay interchangeably: what matters is that there is explicit weight regularization. These experiments also show that the well known advantage of momentum versus Adam for vision tasks is only significant in the presence of $L_2$. In the absence of a benefit from $L_2$ regularization/ weight decay it seems like Adam is a better optimizer, \citet{adagraft} suggested that this is because it can adjusts the learning rate of each layer appropriately and it would be interesting to understand whether there is any connection between that and bouncing.

These experiments have a growing weight norm as can be seen in SM \ref{SM:L20}. While the weight norm does not have to be always increasing in the absence of $L_2$ regularization, this is a function of the learning rate (see section \ref{sec:theorysh}) , and the learning rates used in practice exhibit this property. Homogeneous networks with cross entropy loss will have an increasing weight norm at late times, see \citet{lyu2020gradient}. Even if a simple schedule is competitive this does not imply that other features of convex optimization like the independence of performance in the learning rate carry over. We  repeat the CIFAR-100 experiments for a fixed small learning rate of $0.02$ (the same as the final learning rate for the simple schedule) and the error with $L_2=0$ is $23.8$ while with $L_2 \not =0$ is $29.7$, we see that while there is a performance gap between a small and large learning rates, this gap is much smaller if there is no bouncing (difference in error rate of $1.2 \%$ for $L_2=0$ vs $7.5\%$ for $L_2\not =0$). For a fair comparison with small learning rates, we evolved these experiments for $5$ times longer than the large learning rates, but this did not give any benefit.

While the experiments presented in table \ref{table:nobounce} do not have $L_2$ regularization, some NLP architectures like \citet{devlin2019bert,brown2020language} have weight decays of $0.01, 0.1$ respectively. We tried adding weight decay to our translation models and while performance did not change substantially, we were not able to get a bouncing weight norm.

The effect of different learning rate schedules in NLP was also studied thoroughly in appendix D.6 of \citet{kaplan2020scaling} with the similar conclusion that as long as the learning rate is not small and is decayed near the end of training, performance stays roughly the same.

\begin{figure}[ht!]
  \centering
  \subfloat[Wide Resnet on CIFAR-10]{
    \includegraphics[width=0.24 \textwidth]{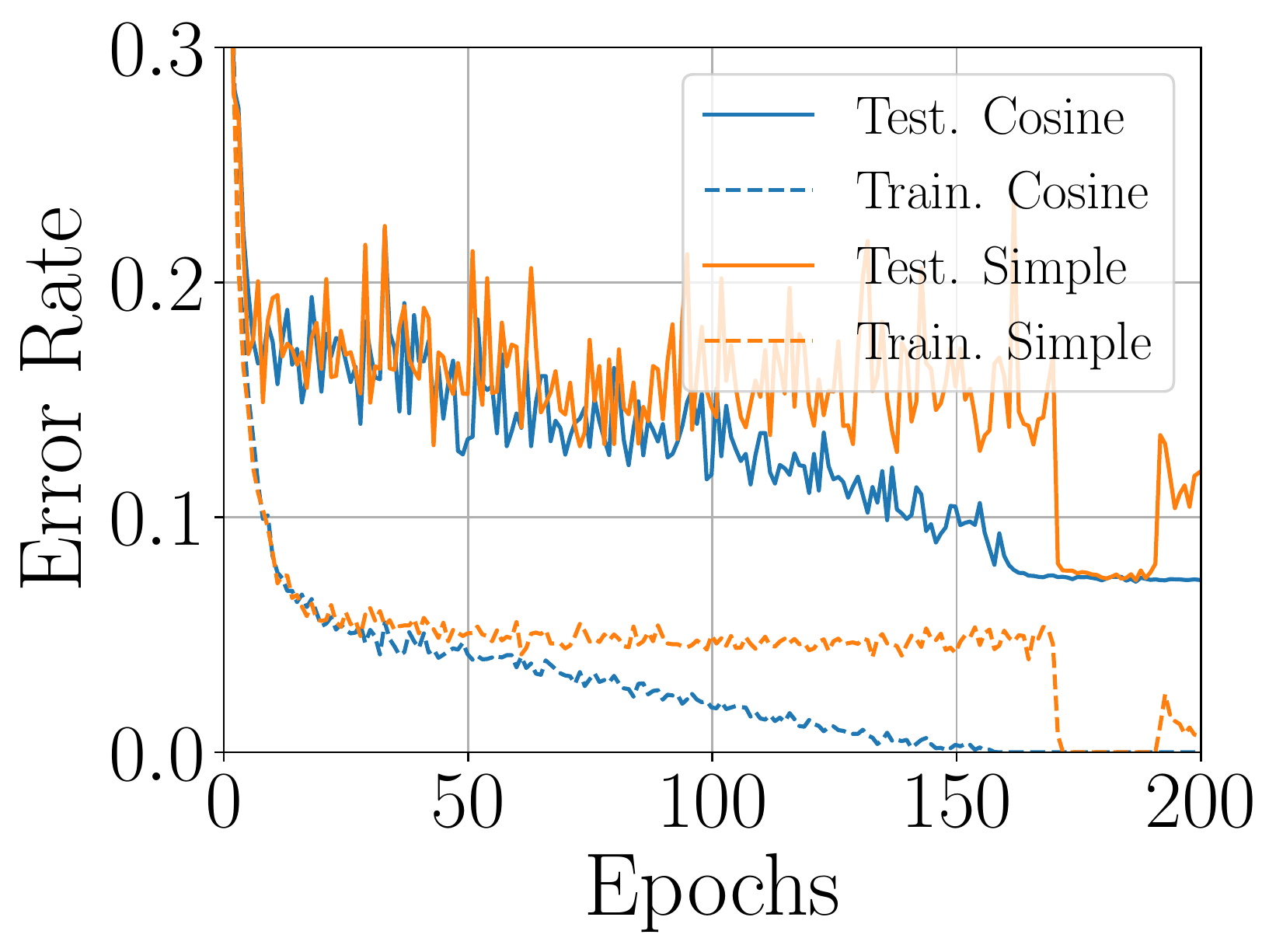}
  }
  \subfloat[Wide Resnet on CIFAR-10]{
    \includegraphics[width=0.24 \textwidth]{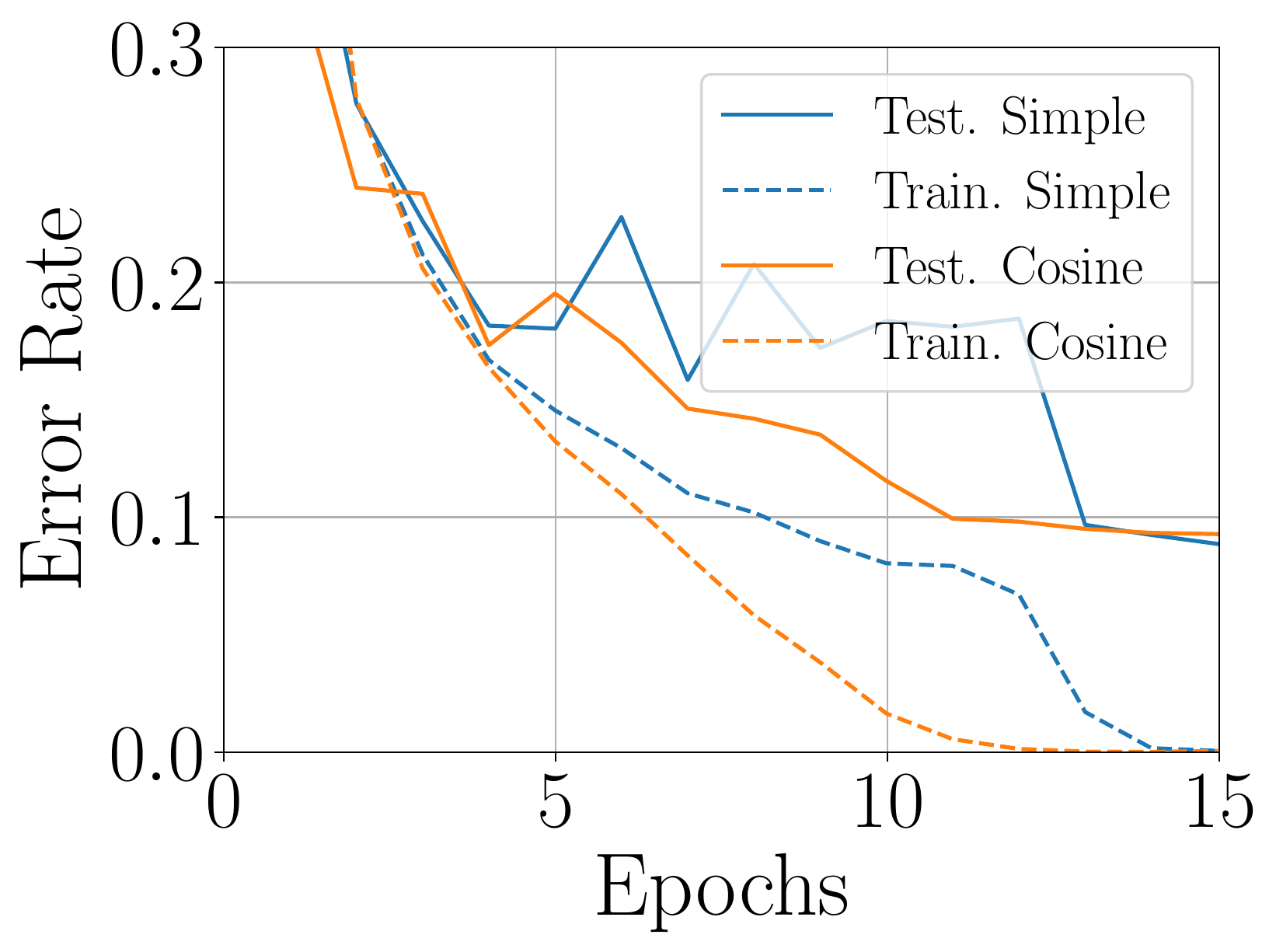}
    }
  \caption{Wide Resnet on CIFAR-10 without data augmentation evolved for $200$ epochs (left) and $15$ epochs (right). In this setup, the weight norm bounces at around $15$ epochs. a) Both schedules reach training error $0$ and their performance is the same (error of $7.3$). b) If we evolve the model for only $15$ epochs, both schedules can still get training error $0$ without a weight norm bounce, we think this is the reason why there is no performance difference in a).  }
  \label{fig:noaug_main}
\end{figure}

\paragraph{The presence of a bouncing weight norm does not guarantee that schedules are beneficial.} From this section, a bouncing weight norm seems to be a necessary condition for learning rate schedules to matter, but it is not a sufficient condition. Learning rate schedules seem only advantageous if the training task is hard enough. In our experience, if the training data can be memorized with a simple learning rate schedule before the weight norm has bounced, then more complex schedules are not useful. This can be seen by removing data augmentation in our Wide Resnet CIFAR experiments, see figure \ref{fig:noaug_main}. In the presence of data augmentation, simple schedules can not reach training error $0$ even when evolved for $200$ epochs, see SM. 

\begin{table*}

{ \centering
\begin{tabular}{ |l|l|l||r|r|  }
 \hline
 \multicolumn{3}{|c||}{Setup} &
 \multicolumn{2}{|c|}{Performance for different schedules} \\
 \hline
 
 Type & Task and metric& Architecture & Complex Decay & Simple Decay   \\ \hline
 \multirow{3}{*}{NLP}  & EN-DE, BLEU & Transformer    & 29.0 & 28.9   \\
 & EN-FR, BLEU & Transformer    & 43.0 & 43.0   \\
 & GLUE, Average score & ALBERT finetuning    & 83.1 & 82.9  \\ \hline
  \multirow{3}{*}{RL} & Qbert, Score & PPO     & 1750 & 1850  \\
  & Seaquest, Score & PPO     & 21.0 & 20.7  \\
  & Pong, Score & PPO    & 22300 & 23000  \\ \hline
   \multirow{4}{*}{Vision $L_2=0$} 
 & ImageNet, Test Error & Resnet-50    & 28.1 & 27.8  \\
 & ImageNet, Test Error & Resnet-50 + Adam    & 28.2 & 28.9  \\
 & CIFAR-10, Test Error & Wide Resnet 28-10     & 5.0 & 5.0  \\
& CIFAR-100, Test Error & Wide Resnet 28-10 & 21.8 & 22.1  \\ \hline \hline
 
& ImageNet, Test Error & Resnet-50    & 23.2 & 28.5  \\
Vision $L_2\not=0$& ImageNet, Test Error & Resnet-50 + Adam    & 24.7 & 26.0 \\
(has bounce) & CIFAR-10, Test Error & Wide Resnet 28-10     & 3.5 & 4.9  \\
&  CIFAR-100, Test Error & Wide Resnet 28-10     & 17.8 & 22.2  \\
 \hline
 
\end{tabular}

}
\hspace*{0.5cm}
\caption{Comparison of performance between a simple learning rate decay and a "complex" decay among tasks: "complex" means cosine decay for vision tasks and linear decay for NLP and RL. For NLP and RL tasks higher metrics imply better performance, while for vision tasks, lower error denotes better performance. None of these tasks (except for the vision task with $L_2$ used as a reference) have weight norm bouncing nor an advantage from non-simple schedules. We have averaged the RL tasks over $3$ runs and their difference is compatible with noise. See \ref{table:Gluetasks} for the individual GLUE scores, as it is common we have omitted the problematic WNLI. }
\label{table:nobounce}
\end{table*}

\section{Understanding weight norm bouncing}
\label{sec:understand}
In this section, we will pursue some first steps towards understanding the mechanism behind the phenomena that we found empirically in the previous sections.

\subsection{Intuition behind bouncing behaviour}
\label{sec:theorysh}
We can build intuition about the dynamics of the weight norm by studying its dynamics under SGD updates:
\begin{eqnarray}
     \Delta |w_{t+1}|^2 = \eta^2 |g_t|^2-2 \eta \lambda |w_t|^2-2 \eta  g_t \cdot w_t +O(\eta^2 \lambda) ~~ 
    \label{eq:weight_norm}
\end{eqnarray}
where $\eta,\lambda$ are the learning rate and $L_2$ regularization coefficient, $g_t \equiv \frac{dL_t}{dw_t}$ is the gradient with respect to the loss (in the absence of the $L_2$ term) and we have used that empirically $\eta \lambda \ll 1$. This equation holds layer by layer, see SM for more details about it. In the absence of $L_2$ regularization, for large enough learning rates ($\eta > \frac{|g_t|^2}{2 g_t.w_t}$), this suggests that the weight norm will be increasing. 

Equation \ref{eq:weight_norm} can be further simplified for scale invariant networks, which satisfy $g_t\cdot w_t=0$, see for example \citet{vanlaarhoven2017l2} \footnote{Scale invariant networks are defined by network functions which are independent of the weight norm: $f(\alpha w)=f(w)$. The weight norm of these functions still affects its dynamics \citep{li2019exponential}.}. In the absence of such term,
 we see that the updates of the weight norm are determined by the relative values of the gradient and weight norm. If $\lambda=0$ or $\eta$ is very small, the weight norm updates will have a fixed sign and thus there will not be bouncing. More generally, we expect that in the initial stages of training, the weight norm is large and its dynamics are dominated by the decay term. As it shrinks, the relative value of the gradient norm term becomes larger and it seems natural that at some point, it will dominate, making the weight norm bounce. This is also studied in \citet{wan2020spherical}, where it is shown that after the bounce, the two terms in equation \ref{eq:weight_norm} are the same order and the weight norm "dynamically equilibrates" (although it can not stay constant because the gradient norm changes with time). While we expect the $g_t \cdot w_t$ to be non-zero in our setups, only layers which are not scale invariant would contribute to this term and  roughly any layer before a BatchNorm layer is scale invariant so we expect this term to be smaller than the other two. 

In our experience, the only necessary condition for a model to have a bouncing weight norm is that is has $L_2$ regularization (or weight decay) and the learning rate is large enough. We expect the previous intuition to apply to other optimizers with weight decay. Empirically, we have seen that different optimizers, losses and batch sizes can have a bouncing weight norm.

\subsection{Towards understanding the benefits of bouncing and schedules}

While it is still unclear why bouncing is correlated with the benefit of schedules, we would like to point in some directions which could help provide a more complete picture.

To better understand this phenomenon it would be useful to distill its properties and find the simplest model that captures it. The bouncing of the weight norm appears to be generic, as long as we have $L_2$ regularization and the learning rate is big enough. We believe that learning rate schedules being only advantageous for hard tasks (as we discussed in section \ref{sec:nobounce}) is the principal roadblock to find theoretically tractable models of this phenomena. 

 For bouncing setups, decaying the learning rate when the weight norm is equilibrating allows the weight decay term in equation \ref{eq:weight_norm} to dominate, causing the weight norm to bounce again. However, from equation \ref{eq:weight_norm}, in the absence of $L_2$ decaying the learning rate can only slow down the weight norm equilibration process and this implies that the weights change more slowly, see SM. It seems like the combination of weight bouncing and decaying the learning rate might be beneficial because it allows the model to explore a larger portion of the landscape.  Exploring this direction further might yield better insights to this phenomenon, perhaps building on the results of \citet{wan2020spherical,kunin2020neural}.

We now conduct several experiments with models with bouncing weight norms in order to understand better its properties.

\begin{figure}[ht!]
  \centering
    \subfloat[]{
\includegraphics[width=0.24\textwidth]{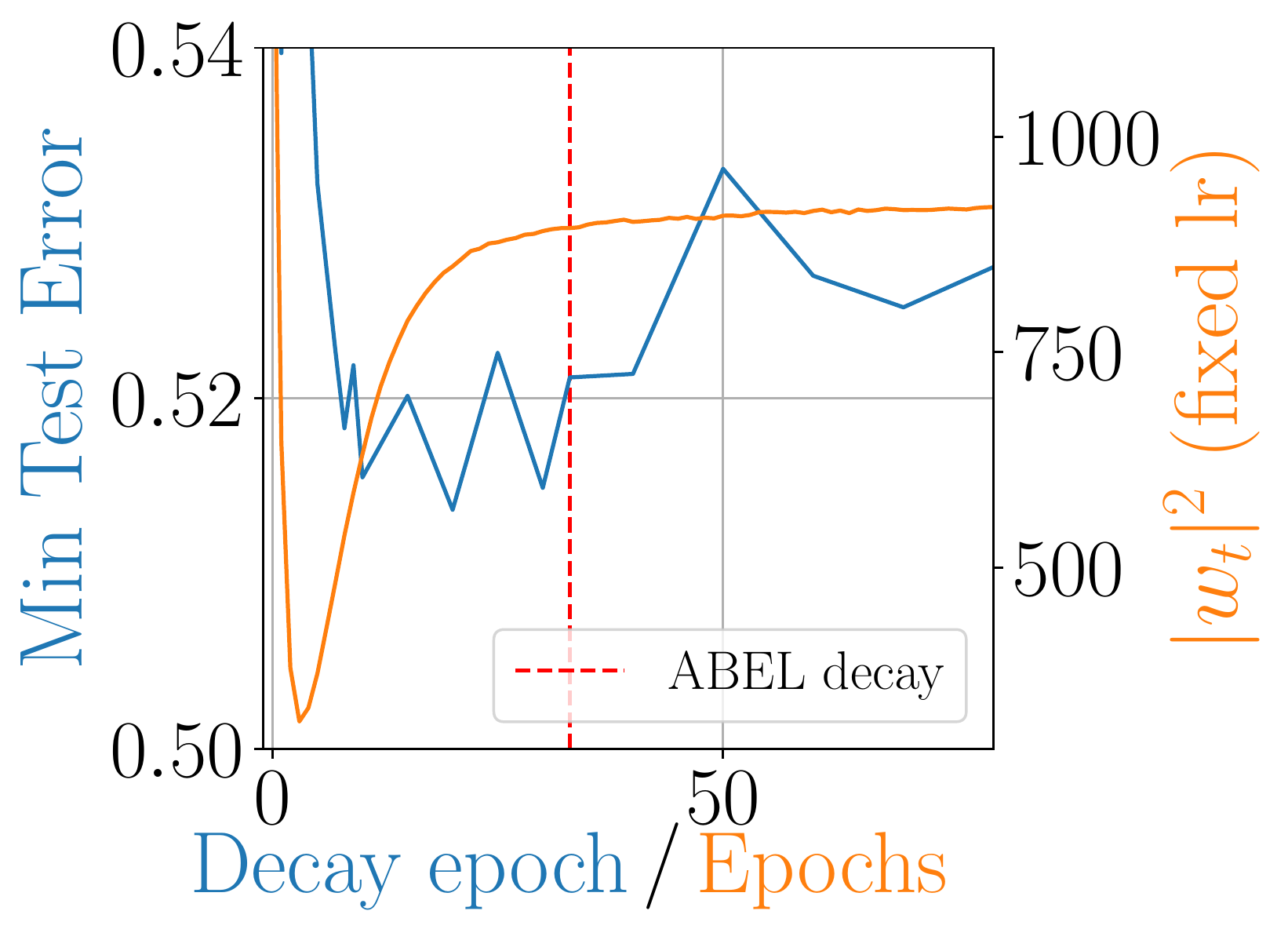}
    \label{fig:vgg5}
  }
  \subfloat[]{
    \includegraphics[width=0.24 \textwidth]{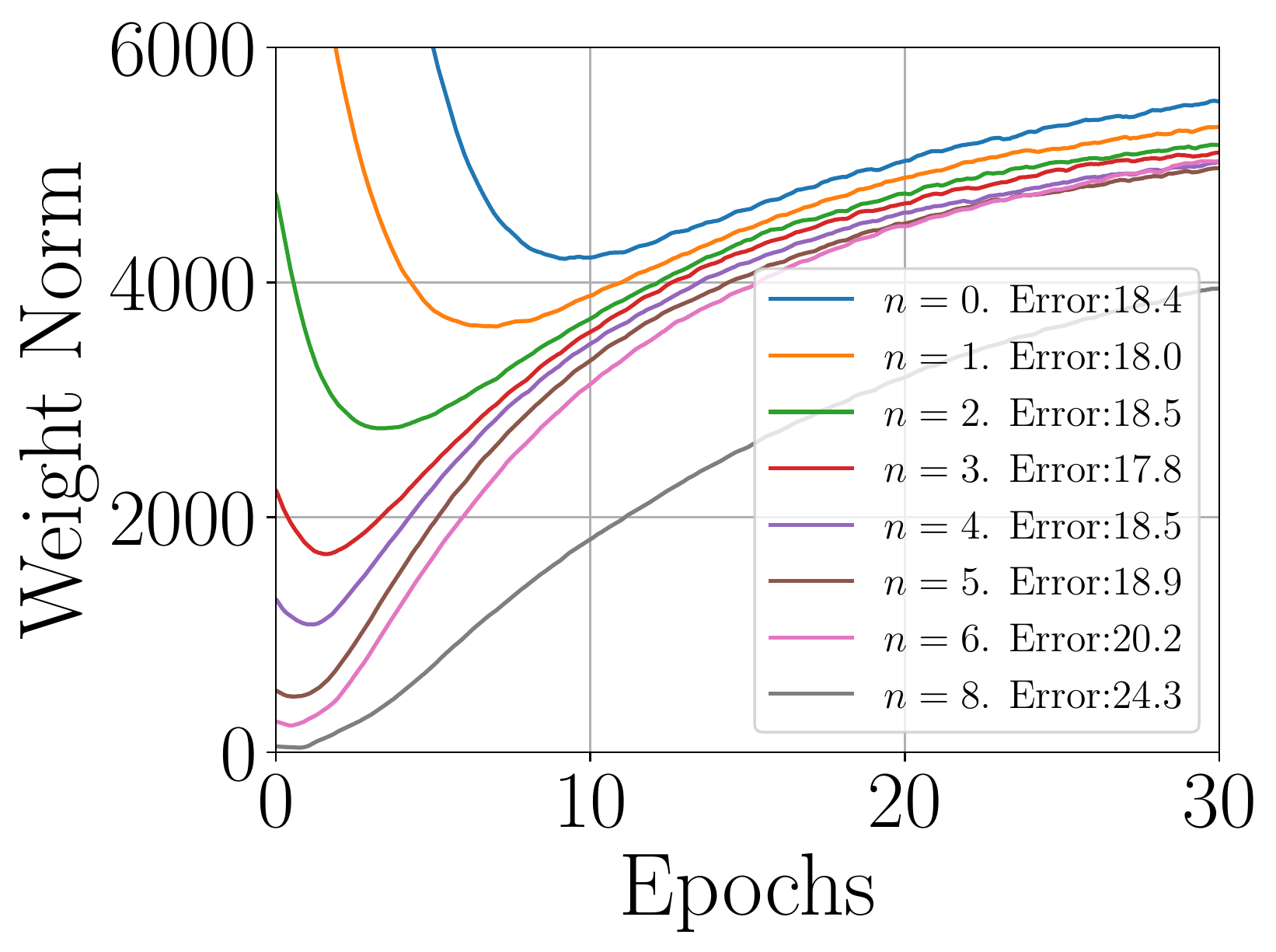}
    \label{fig:init_scale}
  }

  \caption{Experiments exploring features of weight bouncing. a) Minimum test error for VGG5 models decayed at different times (all are evolved for $80$ epochs) and weight norm for fixed learning rate. Models with the best performance are decayed after the bounce, soon before the weight norm dramatically slows down its growth. b) WRN 28-10 models trained with different weight initialization scales ($\sigma_w=\frac{1}{2^n}$) and cosine decay. The weight norm keeps bouncing even if the initialization scale is small and when it stops bouncing, performance is degraded. }
  
\end{figure}

\paragraph{The disadvantage of decaying too early or too late.} Waiting for the weight norm to bounce seems key to get good performance. Decaying too late might be harmful because the weight norm does not have enough time to bounce again, but it is not clear if it is bad by itself. In this section, we run a VGG-5 experiment on CIFAR-100 for $80$ epochs and we decay the learning rate once (by a factor of 10) at different times and compare the minimum test error, see fig \ref{fig:vgg5}. We see that decaying too early significantly hurts performance and the best time to decay is after the weight norm has started slowing down its growth, before it is fully equilibrated. We can compare this sweep over decay times with ABEL: ABEL would be equivalent to a simple schedule decayed at the time marked by the red line (here there is no benefit from decaying it again at the end of training). Given the limitation of the experiment, we can not conclude that decaying to late is hurtful and from the success of cosine decay we expect it is not bad.

\paragraph{Dependence on initialization scale.} One could wonder if the bounce would disappear if we change the initialization of the weights so that the initial weight norm is smaller than the minimum of the bounce with the original normalization. We studied this in figure \ref{fig:init_scale} (see SM for more details) and we see how even for very small initialization scales, there is a bouncing weight norm. If the initialization scale is too small, the bouncing weight norm disappears and the performance gets significantly degraded.

\section{Conclusions}
In this work we have studied the connections between learning rate schedules and the weight norm. We have made the empirical observation that a bouncing weight norm is a necessary condition for complex learning rate schedules to be beneficial, and we have observed that the step-wise schedules tend to decay the learning rates when the weight norm equilibrates after bouncing. We have checked these observations across architectures and datasets and have proposed ABEL: a learning rate scheduler which automatically decays the learning rate depending on the weight norm,  performs as well as fine tuned schedules and is more robust than standard schedules with respect to its initial learning rate and the decay factor. In the absence of weight bouncing, complex schedules do not seem to matter too much.
\ifarxiv

Now we would like to briefly mention on some important points and future directions:

\paragraph{Practical implications.} In \fi vision tasks with $L_2$ regularization, learning rate schedules have a substantial impact on performance and tuning schedules or using ABEL should be beneficial. In other setups, training with a constant learning rate and decay it at the end of training should not hurt performance and might be a preferable, simpler method. 
\ifarxiv
\paragraph{ ABEL's hyperparameters.} The main hyperparameters of ABEL are the base learning rate and decay factor: while our schedule is not hyperparameter free, ABEL is more robust than other schedules to these parameters because of its adaptive nature.

\paragraph{Warmup.} Our discussion focuses on learning rate decays and we have not studied the effects of warmup. This is an interesting problem but seems unrelated to bouncing weight norms and the benefit of learning rate schedules at late times. In this way, whenever warmup is beneficial and widely used (such as large batch Imagenet or Transformer models) we added it on top of our schedule.

\paragraph{Comparison between ABEL and other schedules.} We have mainly compared ABEL with step-wise decays and cosine decay. It can easily be compared with step-wise decay and is objectively better because it does not require to fine tune when to decay the learning rate. Compared with cosine decay, the learning rate in ABEL does not depend explicitly on the total training budget and it is thus preferred for situations where we do not know for how long we want to train or might want to pick up training where it was left.  Compared with other automatic methods, it does not require an outer loop nor adding complicated measurements to training, it only depends on the weight norm which is pretty stable through training (compared with quantities like the training loss or gradients).

\paragraph{The role of simple schedules is to reduce noise.} We used simple schedules as a baseline to reduce SGD noise. This picture is justified by the fact that the most dramatic drop in test error for such schedules is always immediately after the decay, as can be seen in the training curves of the SM.

\paragraph{Memory and compute considerations.} ABEL only depends on the current and past weight norm (two scalars) so it does not add any significant compute nor memory cost.

\paragraph{Weight norm bounce with Adam.} Despite of Adam having an implicit schedule, it still exhibits weight norm bouncing as can be seen in SM . 

\paragraph{What is the behaviour of the layerwise weight norm?} As discussed previously and in the SM \ref{sec:different_layers}, most layers exhibit the same pattern as the total weight norm.

\paragraph{Understanding the source of the generalization advantage of learning rate schedules.} It would be nice to understand if the bouncing of the weight norm is a proxy for some other phenomena. While we have tried tracking other simple quantities, the weight norm seems the best predictor for when to decay the learning rate. In order to make theoretical progress the two significant roadblocks that we identify are that this phenomena requires large learning rates and hard datasets, both of which are complicated to study theoretically. 
\fi

\newcommand{\acktext}{
The authors would like to thank Anders Andreassen, Yasaman Bahri, Ethan Dyer, Orhan Firat, Pierre Foret, Guy Gur-Ari, Jaehoon Lee, Behnam Neyshabur and Vinay Ramasesh for useful discussions.
}

\ifarxiv
\section*{Acknowledgments}
\acktext

\fi

\newpage
\setcounter{equation}{0}
\setcounter{figure}{0}
\setcounter{table}{0}
\setcounter{page}{1}
\setcounter{section}{0}

\renewcommand{\theequation}{S\arabic{equation}}
\renewcommand{\thefigure}{S\arabic{figure}}
\renewcommand{\thetable}{S\arabic{table}}

\newpage
\onecolumn
\section*{Supplementary material}
\appendix
\section{Experimental settings}

\label{sec:expdetails}
We are using Flax ( \url{https://github.com/google/flax}) which is based on JAX \citep{jax2018github}. The implementations are based on the Flax examples and the repository \url{https://github.com/google-research/google-research/tree/master/flax\_models/cifar}  \citep{foret2020sharpnessaware}. 
\ifarxiv  Our simple Flax implementation can be found in SM \ref{SM:flax}. \fi All datasets are downloaded for tensorflow-datasets and we used their training/test split.

All experiments use the same seed for the weights at initialization and we consider only one such initialization unless otherwise stated. We have not seen much variance in the phenomena described, see Table $1,2$ description. All experiments use cross-entropy loss.

We run all experiments in v3-8 TPUs. The WideResnet 28-10 CIFAR experiments take roughly $1$h per run, the ResNet-50 Imagenet experiments take roughly $5$h per run, the PyramidNet CIFAR experiments take roughly $50$h per run, the Transformer experiments take roughly $24$h per run, the ALBERT finetuning experiments take around $10-20$minutes per task.

Here we describe the particularities of each experiment in the figure/tables.

\textbf{Table \ref{table:table1}}.
All models use momentum with a momentum parameter of  $0.9$ and none of them uses dropout. ABEL considerations: learning rate decay $0.1$ for ImageNet and SVHN and $0.2$ for CIFAR-100. We force ABEL to decay at $85\%$ of the total epoch budget by the same decay factor. We run the CIFAR-10 and CIFAR-100 WideResnet experiments with three different seeds and report the average accuracy, the standard deviation across experiments for all schedules and CIFAR-10, CIFAR-100 is $ \le 0.1\%$, except for ABEL and CIFAR-100 which has a standard deviation of $0.2\%$ (the standard error of the mean is $0.1\%)$, from these observations we only consider a single run for the other experiments. For the CIFAR and WRN experiments, the gradient norm is clipped to $5$.
\begin{itemize}
    \item Resnet-50 on Imagenet: learning rate of $0.8$, $L_2$ regularization of $0.0001$ (we do not decay the batch-norm parameters here), label smoothing of $0.1$ and $90$ epochs. All experiments include a linear warmup of $5$ epochs. Standard data augmentation: horizontal flips and random crops. Step-wise decay: multiply learning rate by $\alpha=0.1$ at $30,60$ epochs.  
    
    \item Wide Resnet 16-8 on SVHN: learning rate of $0.01$, $L_2$ regularization of $0.0005$, batch size of $128$, no dropout and evolved for $160$ epochs. Training data includes the "extra" training data and no data augmentation. Step-wise decay: multiply learning rate by $\alpha=0.1$ at $80,120$ epochs.  
    
    \item Wide Resnet 28-10 on CIFAR10/100: learning rate of $0.1$, $L_2$ regularization of $0.0005$, batch size of $128$ and evolved for $200$ epochs. Standard data augmentation: horizontal flips and random crops. Step-wise decay: multiply learning rate by $\alpha=0.2$ at $60,120,160$ epochs. 
 
    \item Shake-Drop PyramidNet on CIFAR100: learning rate of $0.1$, $L_2$ regularization of $0.0005$, batch size $256$, evolved for $1800$ epochs. Uses AutoAugment and cutout as data augmentation. Because training this model takes a long time and weight bounces generally take longer at smaller learning rates, we found it convenient to average the weight norm every five epochs when using ABEL in order to avoid the decays from being noise dominated (without this, the third decay happens too early and the test error increases by $0.2\%$) .
    
    \item VGG-16 on CIFAR10: learning rate of $0.05$, $L_2$ regularization of $0.0005$, batch size $128$, evolved for $200$ epochs. Basic data augmentation and no batch norm.
    
\end{itemize}

\textbf{Table \ref{table:nobounce} NLP, RL.}
All these experiments use ADAM, no dropout nor weight decay. 
\begin{itemize} 
\item Base Transformer trained for translation: learning rate: $1.98 \cdot 10^{-3}$, learning rate warmup of $10$k steps, evolved for 100k steps, batch size $256$ uses reverse translation. Simple decay: decay by $0.1$ at $90$k steps. The translation tasks correspond to WMT'17.  
\item ALBERT and Glue finetuning. Fine tuned from the base ALBERT model of https://github.com/google-research/albert.    Tasks: 
All tasks were trained for 10k steps (including 1k linear warmup steps) for four learning rates: $\lbrace 10^{-5}, 2\cdot 10^{-5},5 \cdot 10^{-5}, 10^{-6} \rbrace$, reported best learning rate, batch size $16$. See table \ref{table:Gluetasks} for the specific scores. Simple decay: decay by $0.1$ at $0.8$ of training. The individual scores are summarized in \ref{table:Gluetasks}.
\item PPO: learning rate $2.5 \cdot 10^{-4}$, batch size $256$, $8$ agents.  Evolved for 40M frames. Simple decay: decay by $0.1$ at $0.9$ of training. Rest of configuration is the default configuration of https://github.com/google/flax/examples. Reported score is the average score over the last $100$ episodes.  We have three runs per task and schedule. The scores with a 95\% confidence interval are: Seaquest: Linear: $1750 \pm 260$, Simple: $1850 \pm 40$; Pong: Linear $21.0 \pm 0.01$, Simple $20.7 \pm 0.5$; Qbert: Linear $22300 \pm 2950$, Simple $23000 \pm 1800$.
\end{itemize}
\textbf{Table \ref{table:nobounce} Vision.}

These experiments are the same as table \ref{table:table1} but without $L_2$. The simple decay is defined by evolving with a fixed learning rate during $85\%$ of the time and then decay it (with decay factor $0.1$ for ImageNet and $0.2$ for CIFAR). In the case of ImageNet+Adam, the learning rate was reduced to $0.0008$. For simple decay with $L_2 \not = 0$, the test error would often increase after decaying, in such case, we choose the test error at the minimum of the training loss, see fig \ref{fig:L20_training_curves}a for an experiemnt with this behaviour.

\textbf{Fig \ref{fig:vgg5}}: VGG-5 trained with $\eta=0.25,\lambda=0.005$ and decayed by a factor of $10$ at the respective time.

\textbf{Rest of figures.} Small modifications or further plots of previous experiments, changes are specified explicitly. 

\begin{table}

 \centering
\begin{tabular}{ |l||r|r|r|r|r|r|r|r|r|r|r|  }
 \hline
 Task & MNLI(m/mm) & QQP & QNLI & SST-2 & CoLA & STS-B & MRPC & RTE  & Average    \\
 \hline
Linear Decay & 82.6/83.4 & 88.1 & 92.1 & 92.8 & 58.5 & 91.0 & 88.7 & 70.8 & 83.1 \\
Simple Decay & 82.9/83.5 & 87.6 & 91.1 & 92.1 & 57.0 & 91.2 & 88.2 & 73.3 & 82.9 \\
 \hline
\end{tabular}

\hspace*{0.5cm}
\caption{Evaluation metric is accuracy except for CoLA (Matthew correlation), MRPC (F1) and STS-B (Pearson correlation).}
\label{table:Gluetasks}
\end{table} 
\clearpage
\section{Experimental details}
\subsection{The mild dependence on the last decay epoch}
\label{sec:last_decay_epoch}
\begin{figure}[ht!]
  \centering
 {
    \includegraphics[width=0.33 \textwidth]{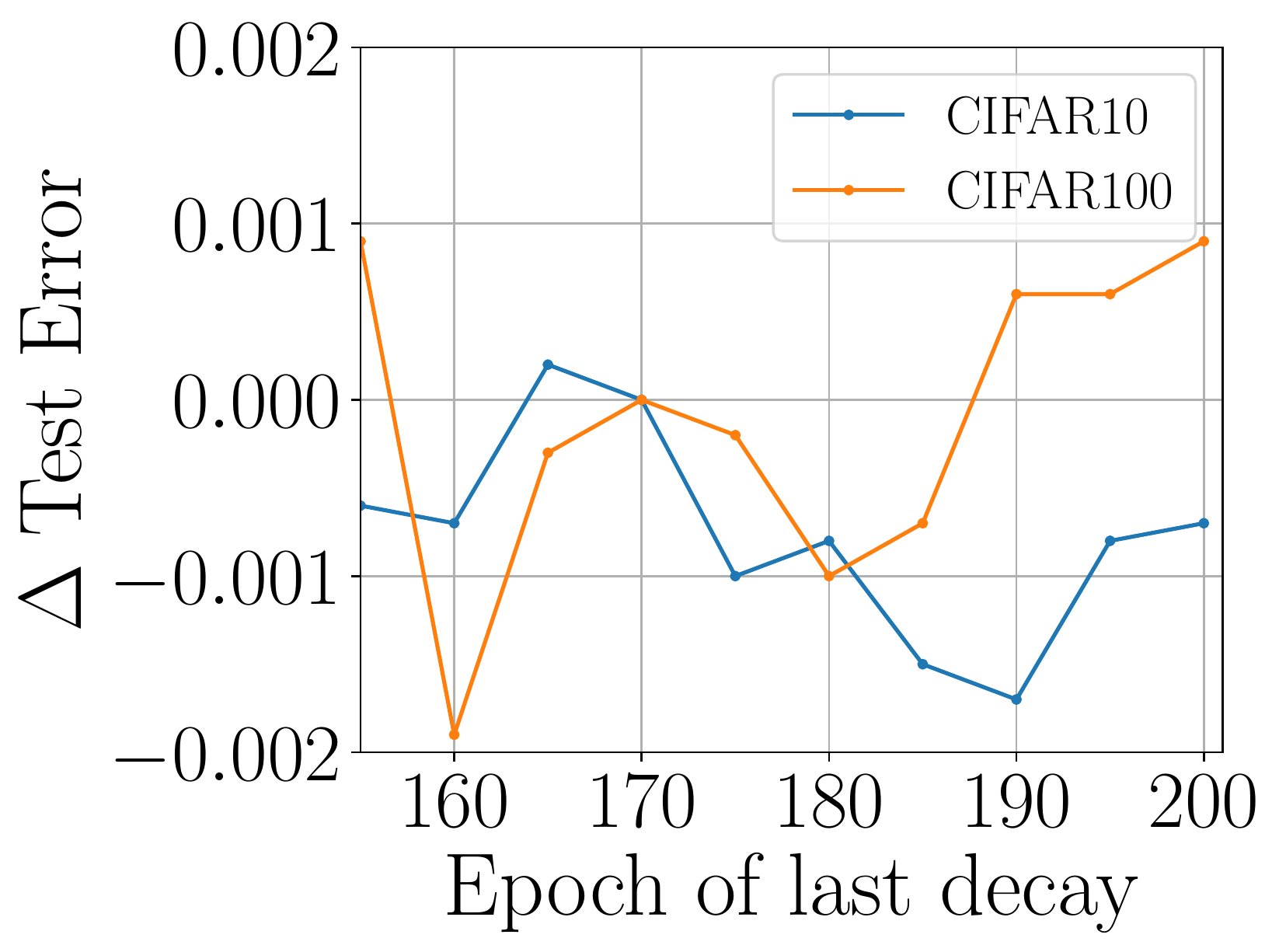}
  }
  
  \caption{Models trained with ABEL with different values for `last\_decay\_epoch`. Difference of test error with respect to $85\%$ of training time ($170$ epochs). We see that the test error does not change too much, most changes are less than $0.2\%$.   }
  \label{fig:decay_time}
\end{figure}

In figure \ref{fig:decay_time} we run again the CIFAR experiments with ABEL for different values of `last\_decay\_epoch`. We see how the final performance does not depend too much on this value.

\subsection{Layer-wise dependence dynamics of weight norm.}
\label{sec:different_layers}
\begin{figure}[ht!]
  \centering

  \subfloat[Resnet-50 on Imagenet]{
    \includegraphics[width=0.33 \textwidth]{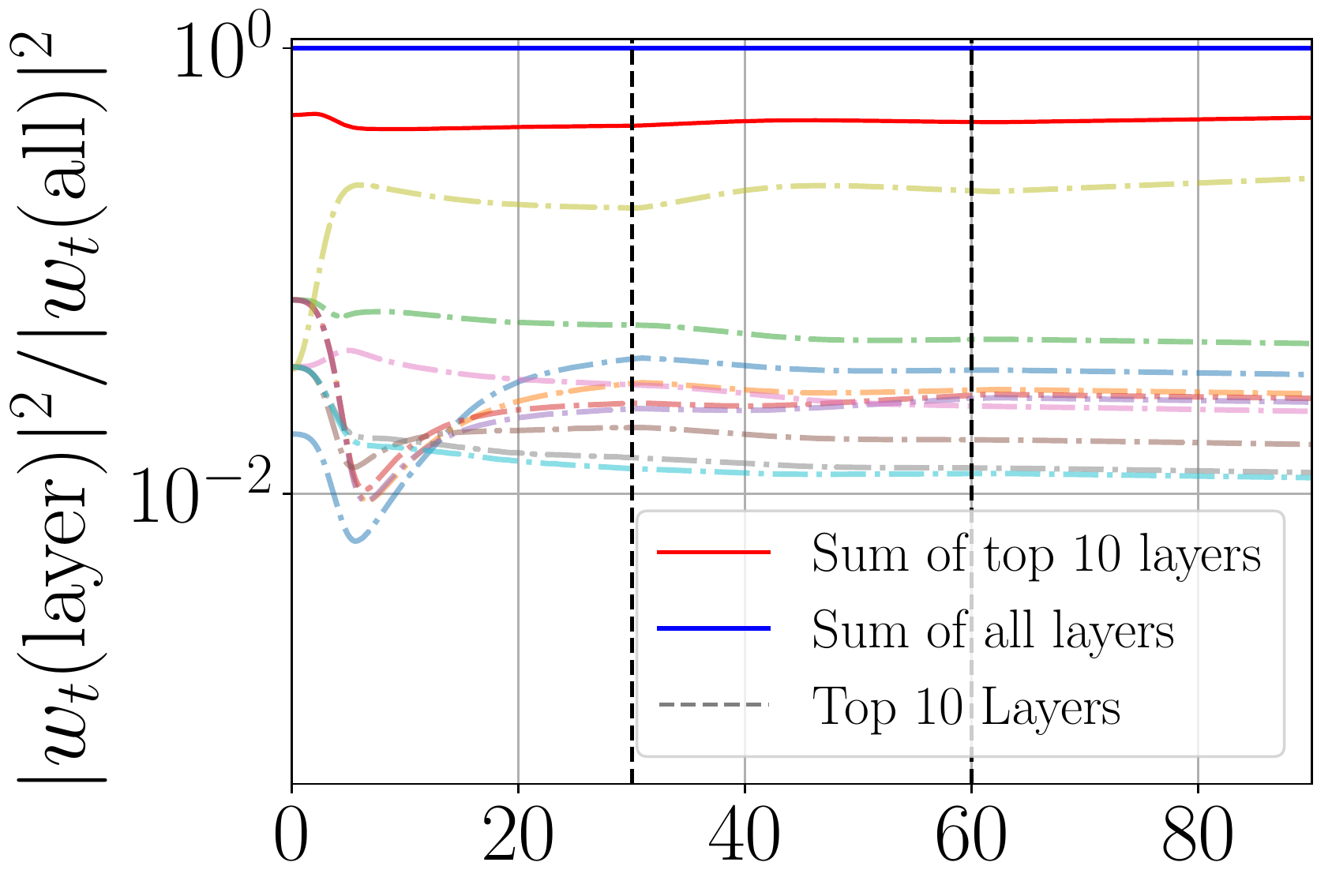}
  }
      \subfloat[Resnet-50 on Imagenet]{
    \includegraphics[width=0.33 \textwidth]{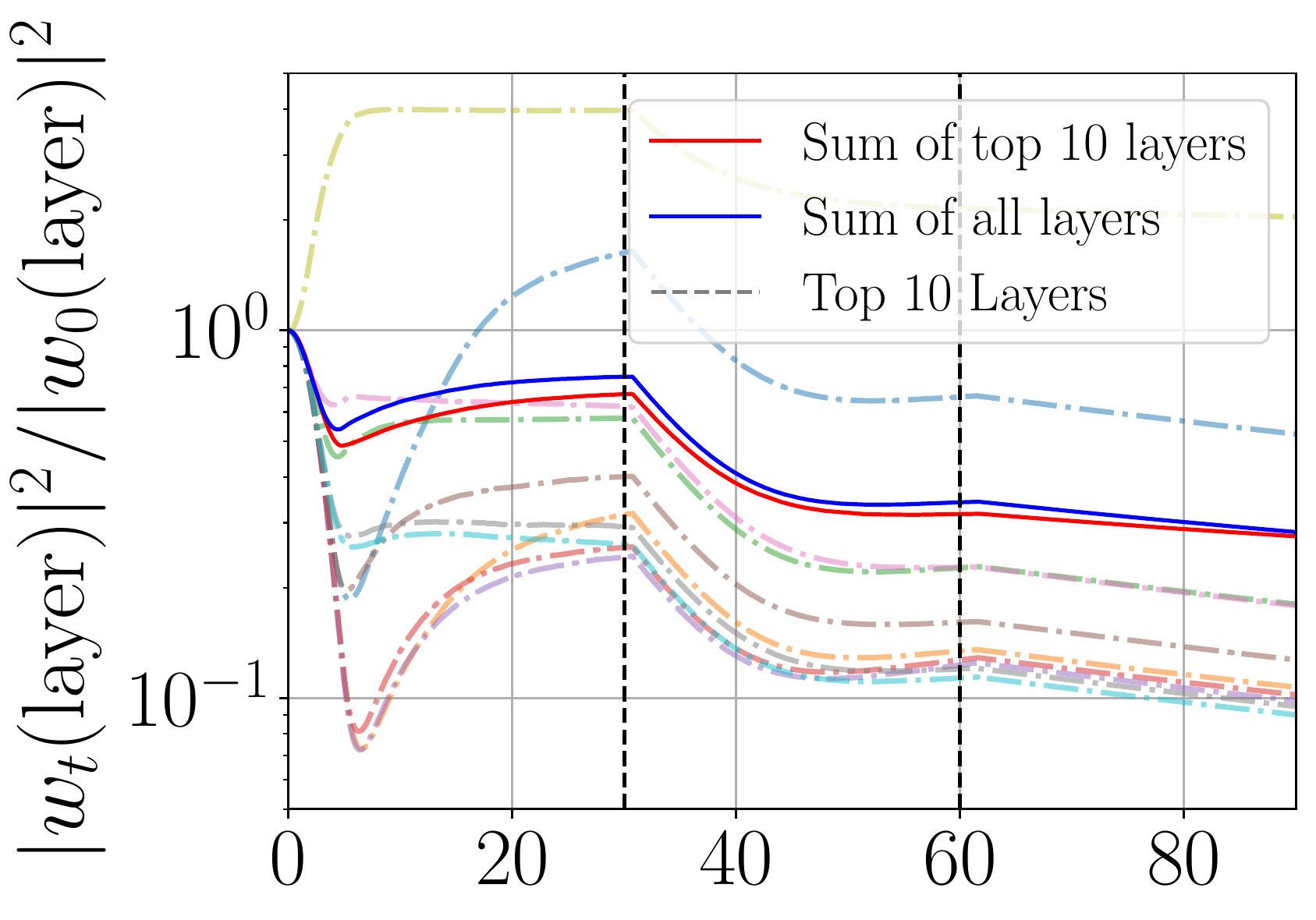}
  }
  \caption{a) Evolution of the quotient between the weight norm of different layers and the total weight norm. b) Evolution of the weight norm of different layers, normalized by its value at initialization. }
  \label{fig:different_layers}
\end{figure}

In this section, we study the dynamics of the layer-wise weight norm. We consider the Resnet-50 Imagenet setup of table \ref{table:table1} with step-wise decay. For better visualization, we focuse on the top $10$ layers, we rank layers by their maximum weight norm during training. These top $10$ layers are mostly intermediate but there are also early and late layers. In figure \ref{fig:different_layers}a, we plot the contribution the quotient between the weight norm of each layer (or the sum of the top $10$ layers) and the total weight norm. We see these top layers account for $\sim 50\%$ of the total weight norm and they exhibit the same dynamics: the red line is straight. At the individual layer level, we see that several layers become constant really fast while there are $\sim 3$ which have a deeper bounce but then have the same dynamics as the total weight norm. We can illustrate this in fig \ref{fig:different_layers}b, where we compare the evolution of the different weights (normalized by their value at initialization). Most layers have the behavour of the total weight norm: a bounce and slow down of growth before decaying.

\newpage
\subsection{Decay LR on Loss Plateau}
\label{SM:losseq}

  We implemented a simple version of this scheduler in FLAX with the default values for the hyperparameters from the pytorch implementation , except for the decay factor (called factor) which we set to $0.1$ for ImageNet and $0.2$ for CIFAR. The other relevant hyperparameters which we did not change are   patience$=10$ (Number of epochs with no improvement after which learning rate will be reduced), threshold$=0.0001$ (Relative Threshold for measuring the new optimum, to only focus on significant changes.). See the  \url{https://pytorch.org/docs/stable/optim.html} for more details. The test errors for WRN 28-10 on CIFAR-10 are $3.9$ and $47.4$ for Resnet-50 on Imagenet. 
  For Imagenet, the loss still decreases slightly during training so it does not seem like the logic behind this schedule could yield good results.
  
  For CIFAR-10, the first decay is caused by a real plateau, but after that the loss increases for more than $10$ epochs, so the learning rate decays again. One decay after that, the loss basically stays in a plateau and the learning rate decays without bound.
  It is surprising that this schedule does so well on CIFAR-10 despite of the learning rate becoming so low so early. 
  
  These two examples exhibit the opposite issues: too much vs too few decay. It seems unrealistic that proper hyperparameter tuning can fix both problems, among other things because the ImageNet Resnet loss is slowly decaying. 
  \begin{figure}[ht!]
  \centering
  \subfloat[Wide Resnet on CIFAR-10]{
    \includegraphics[width=0.33 \textwidth]{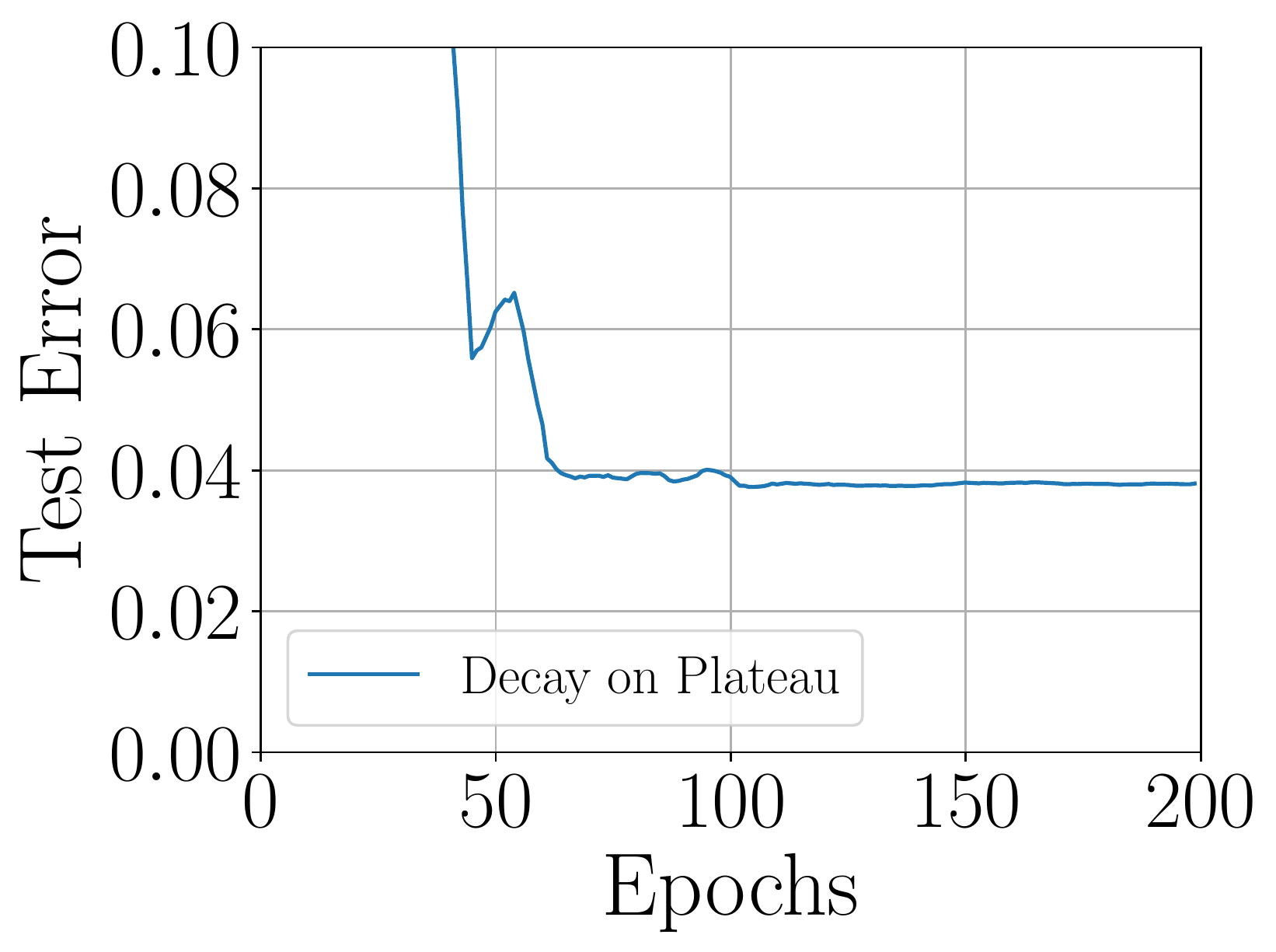}
  }
  \subfloat[Wide Resnet on CIFAR-10]{
    \includegraphics[width=0.33 \textwidth]{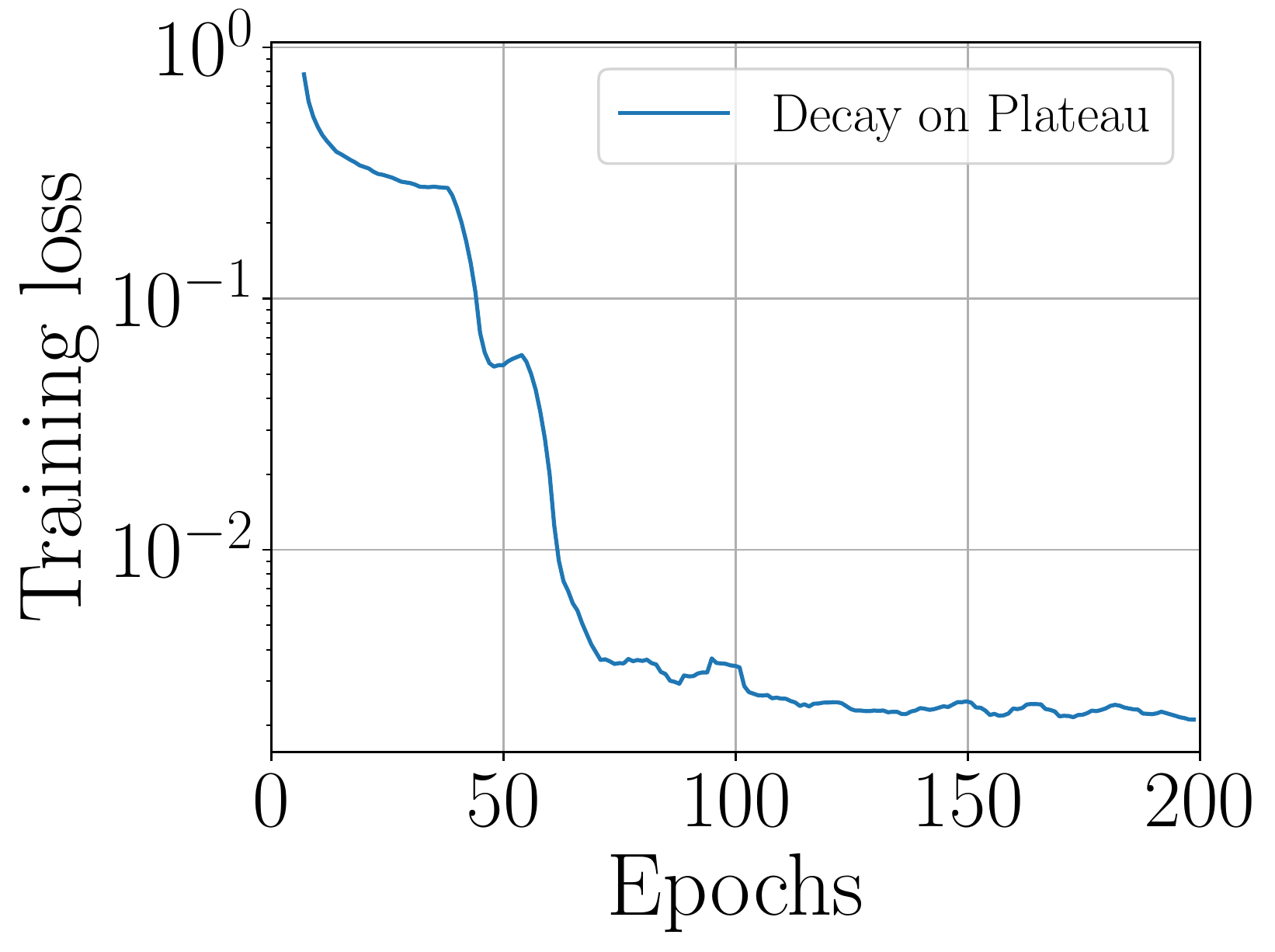}
    
  }
    \subfloat[Wide Resnet on CIFAR-10]{
    \includegraphics[width=0.33 \textwidth]{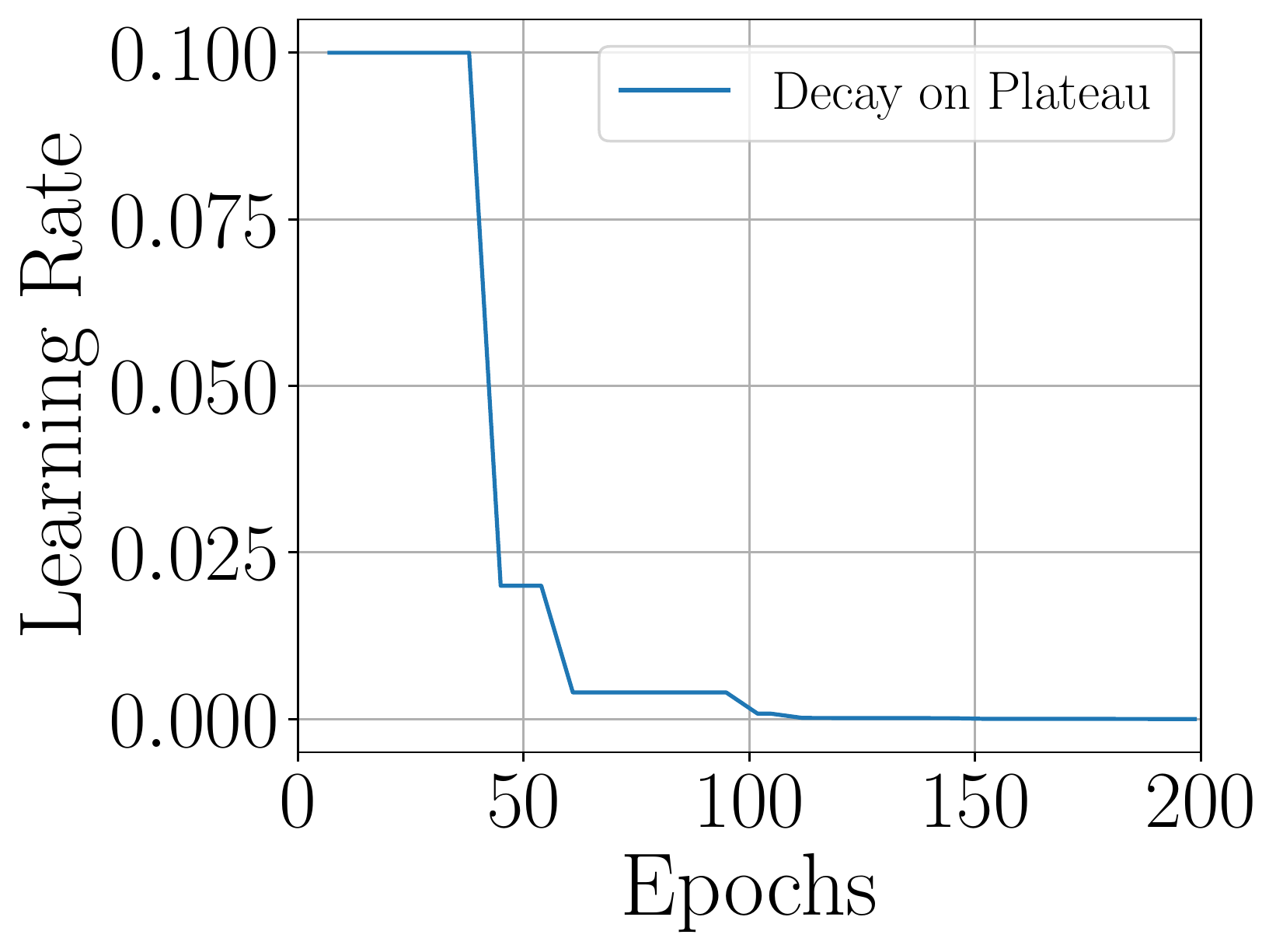}
    
  }
  \\
  \subfloat[Resnet-50 on Imagenet]{
    \includegraphics[width=0.33 \textwidth]{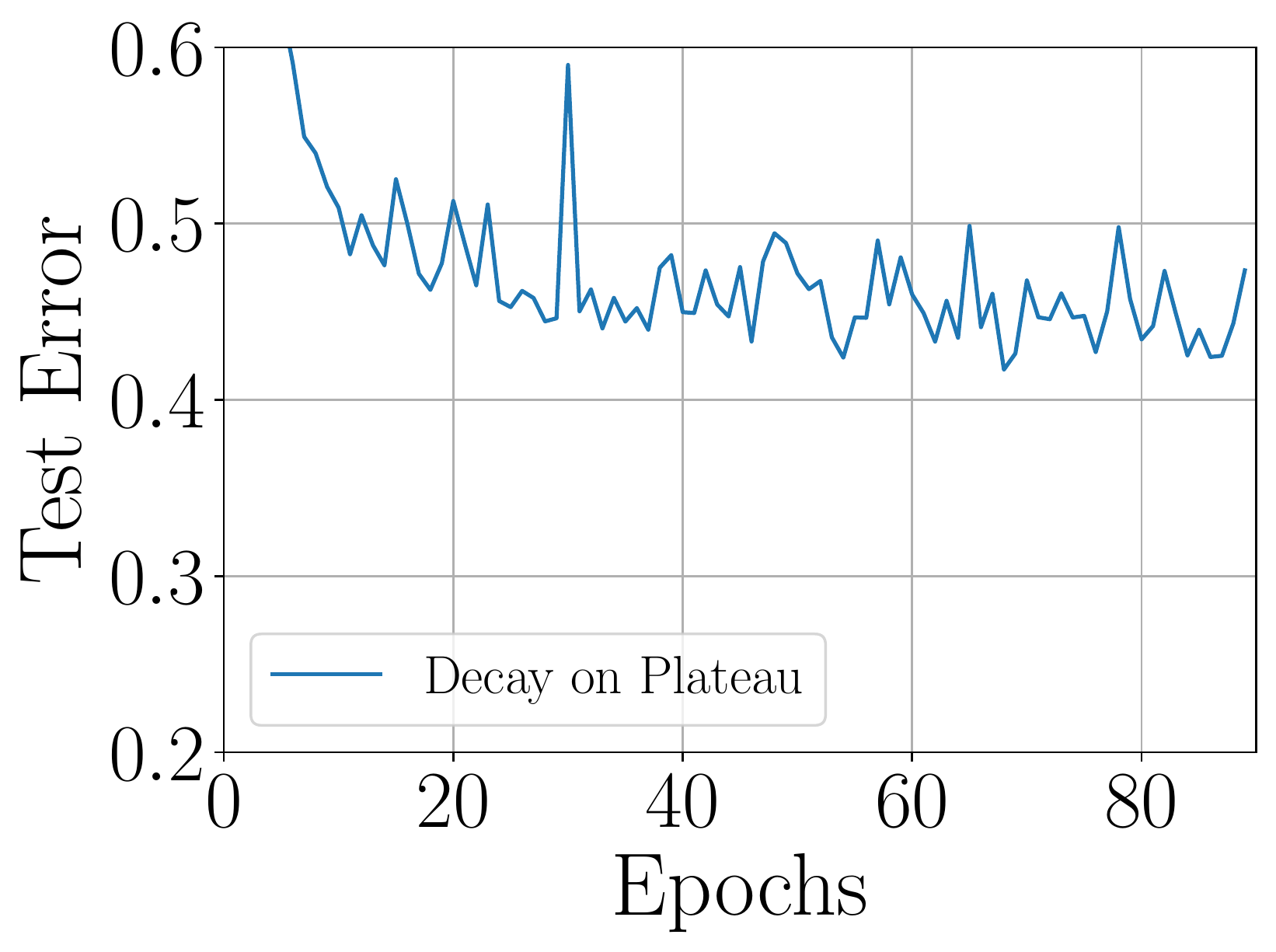}
  }
  \subfloat[Resnet-50 on Imagenet]{
    \includegraphics[width=0.33 \textwidth]{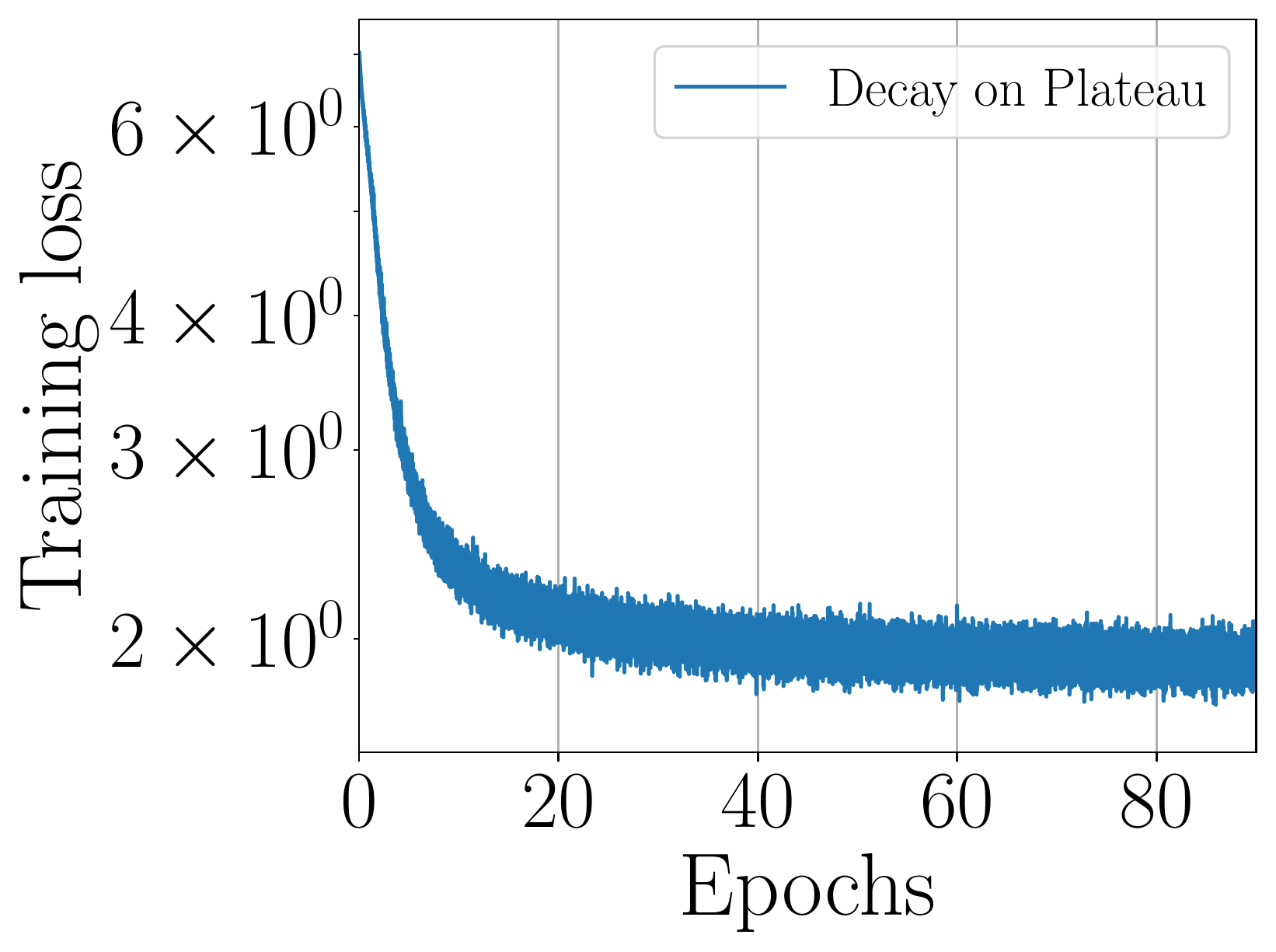}
  }
      \subfloat[Resnet-50 on Imagenet]{
    \includegraphics[width=0.33 \textwidth]{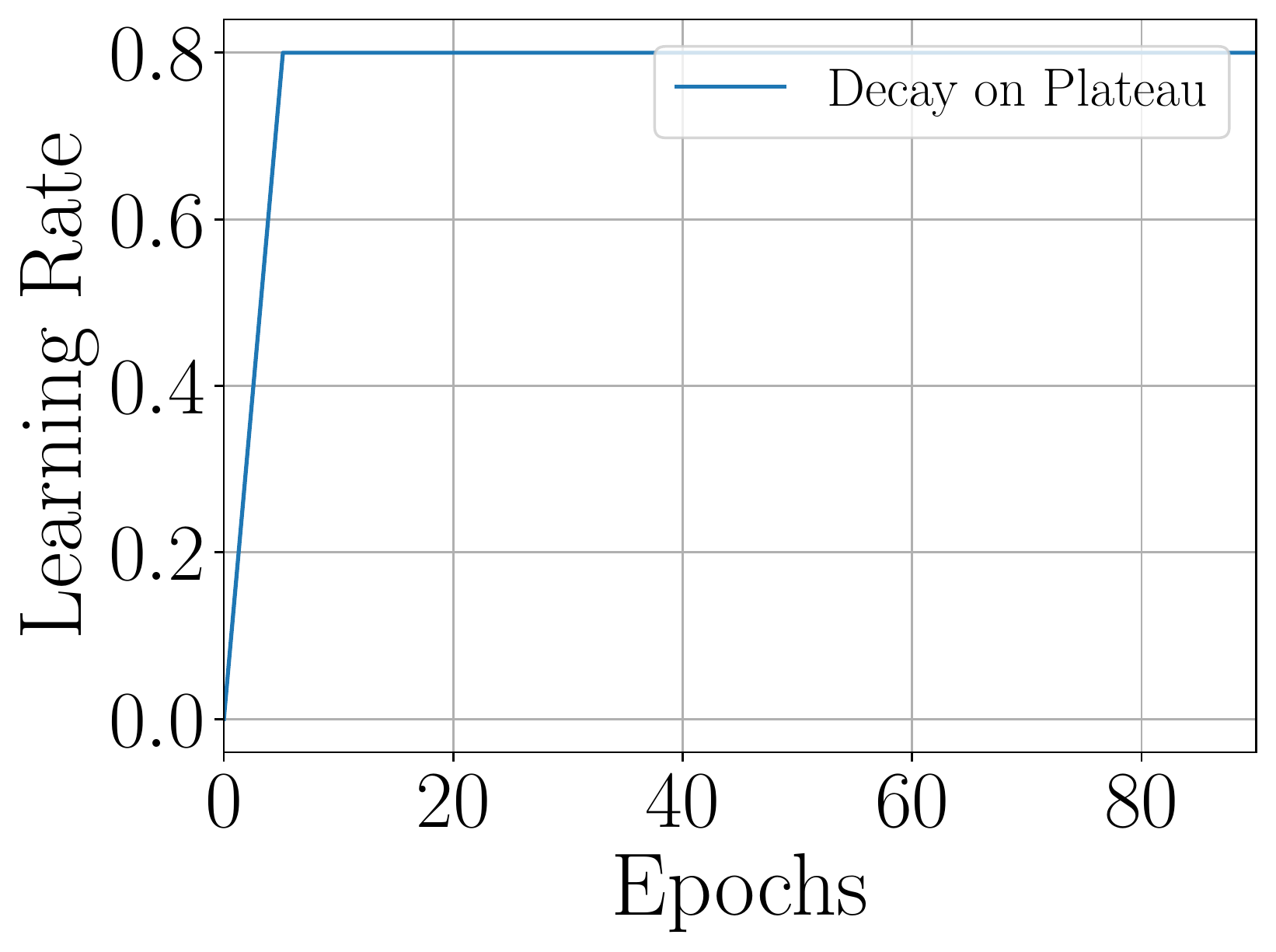}
  }
  \caption{Experiments from the main section with a \texttt{ReduceLROnPlateau} schedule.  }
  \label{fig:loss_eq}
\end{figure}
\newpage
\subsection{No advantage for easy tasks}
\label{sec:easy_tasks}
We expand on the discussion of the main text. We train the same Wide Resnet models of table \ref{table:table1} but without data augmentation. We see that in this case, simple decay reaches training error $0$ at the end of training and there is no advantage of cosine decay. The test error without data augmentation for CIFAR100 and  $29.7$ (cosine) and $29.4$ (simple) and for CIFAR10 it is $7.3$ for both schedules. We attribute this to the fact that when evolved for a small number of epochs, before the weight norm has time to bounce,  the simple decay schedule can already reach training error $0$ and thus it can not benefit from the bounce. 

  \begin{figure}[h!]
  \centering
  \subfloat[Wide Resnet on CIFAR-10]{
    \includegraphics[width=0.33 \textwidth]{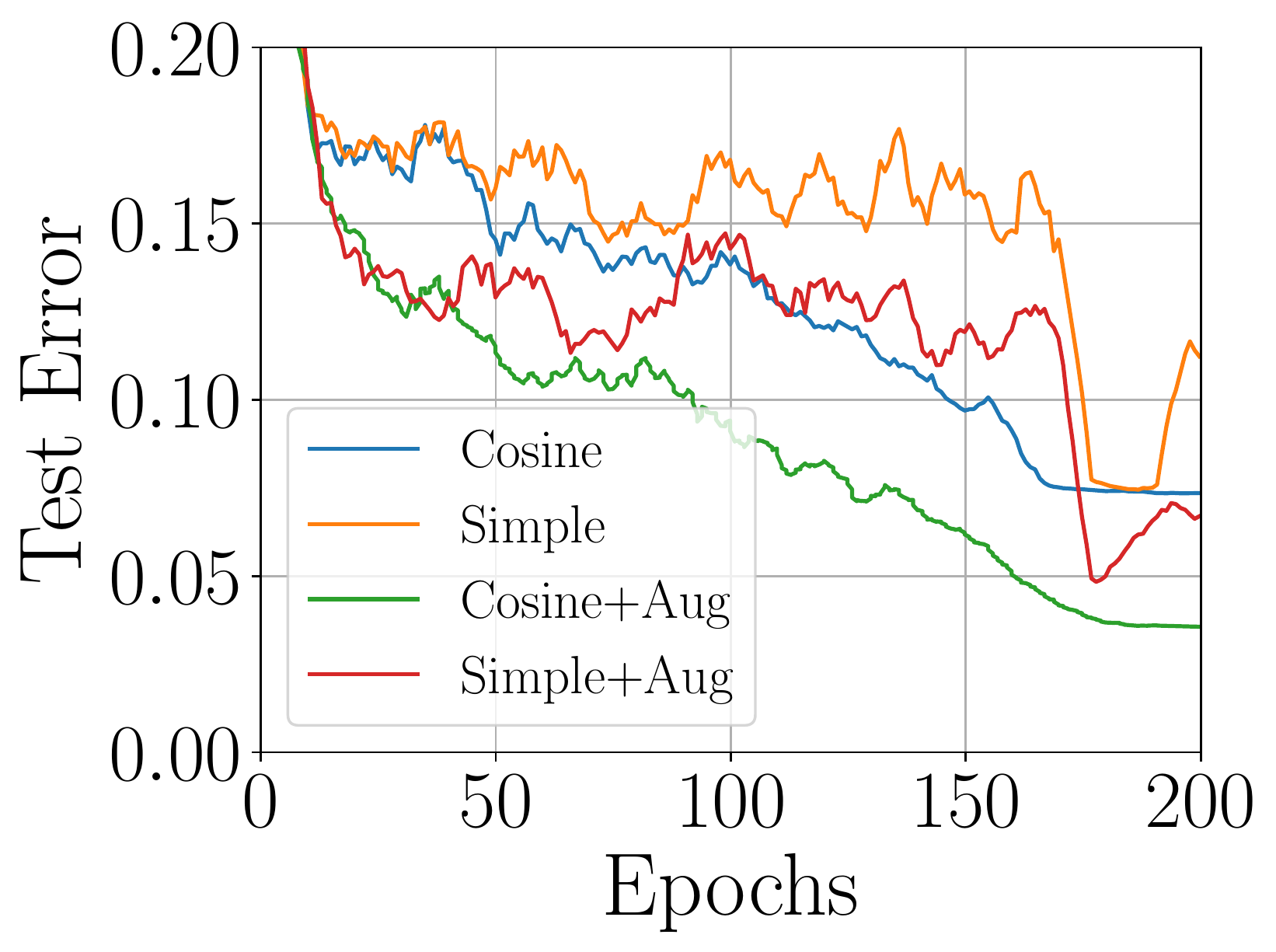}
  }
  \subfloat[Wide Resnet on CIFAR-10]{
    \includegraphics[width=0.33 \textwidth]{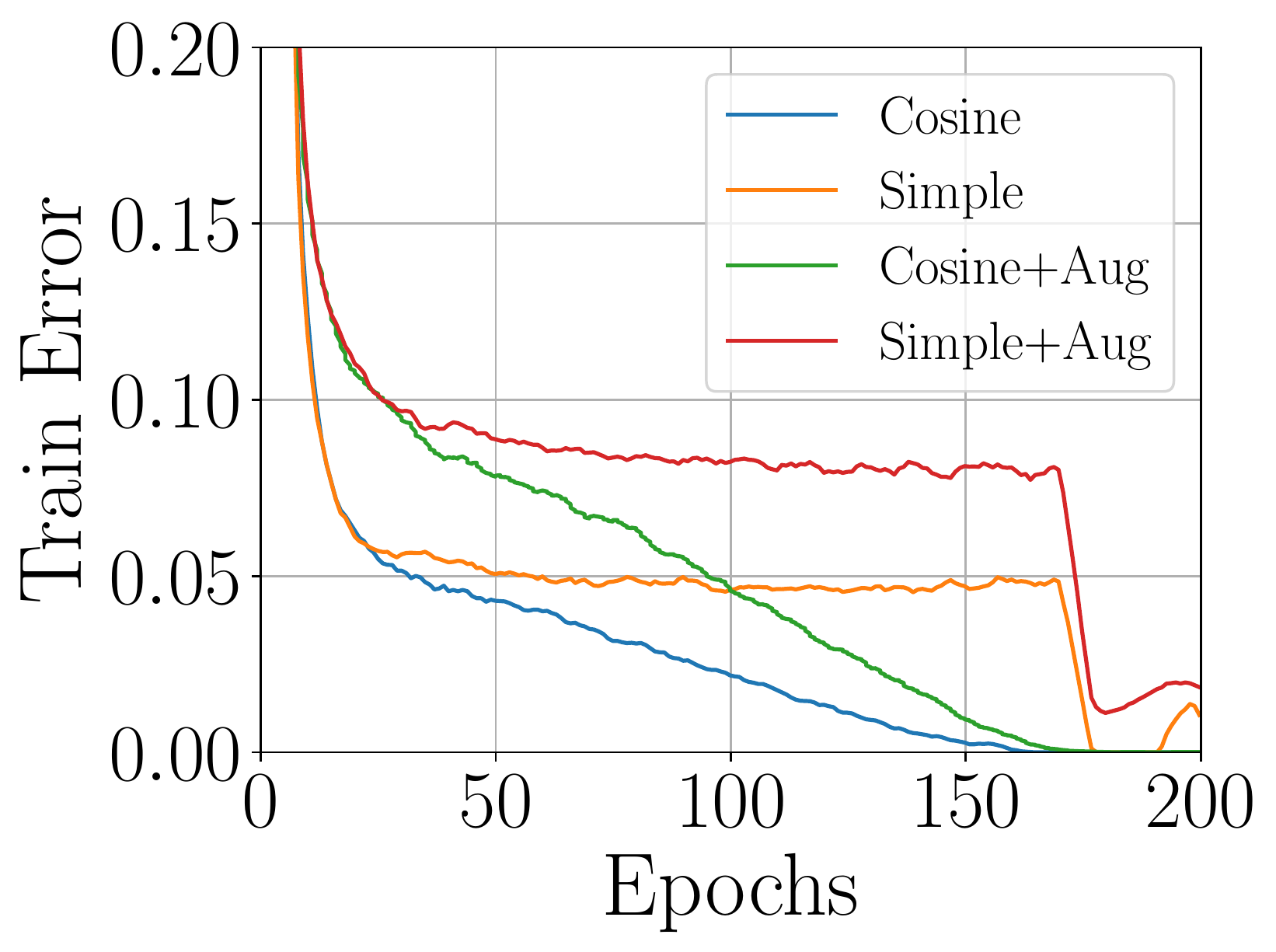}
    
  }
    \subfloat[Wide Resnet on CIFAR-10]{
    \includegraphics[width=0.33 \textwidth]{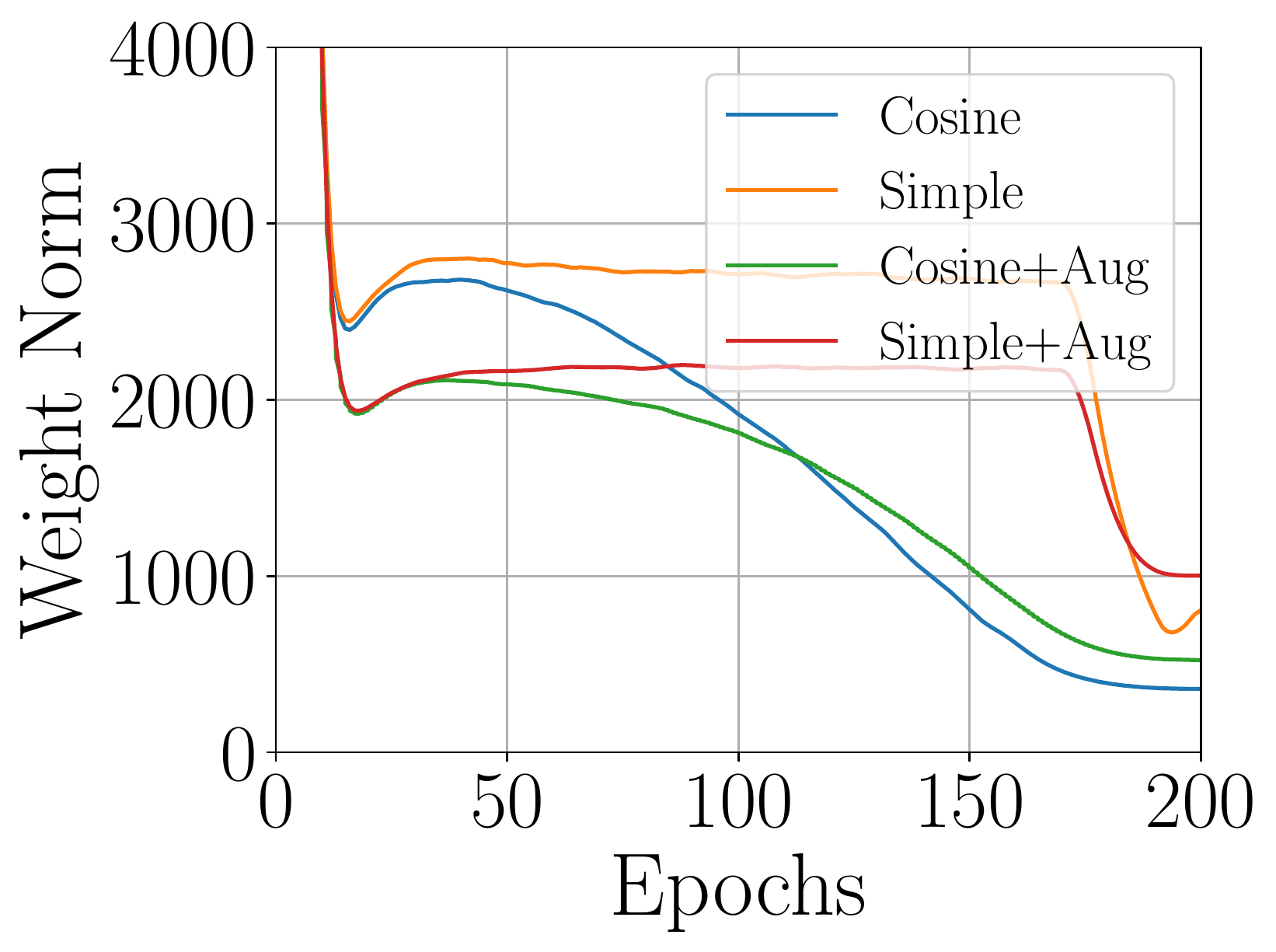}
    
  }
  \\
  \subfloat[Wide Resnet on CIFAR-100]{
    \includegraphics[width=0.33 \textwidth]{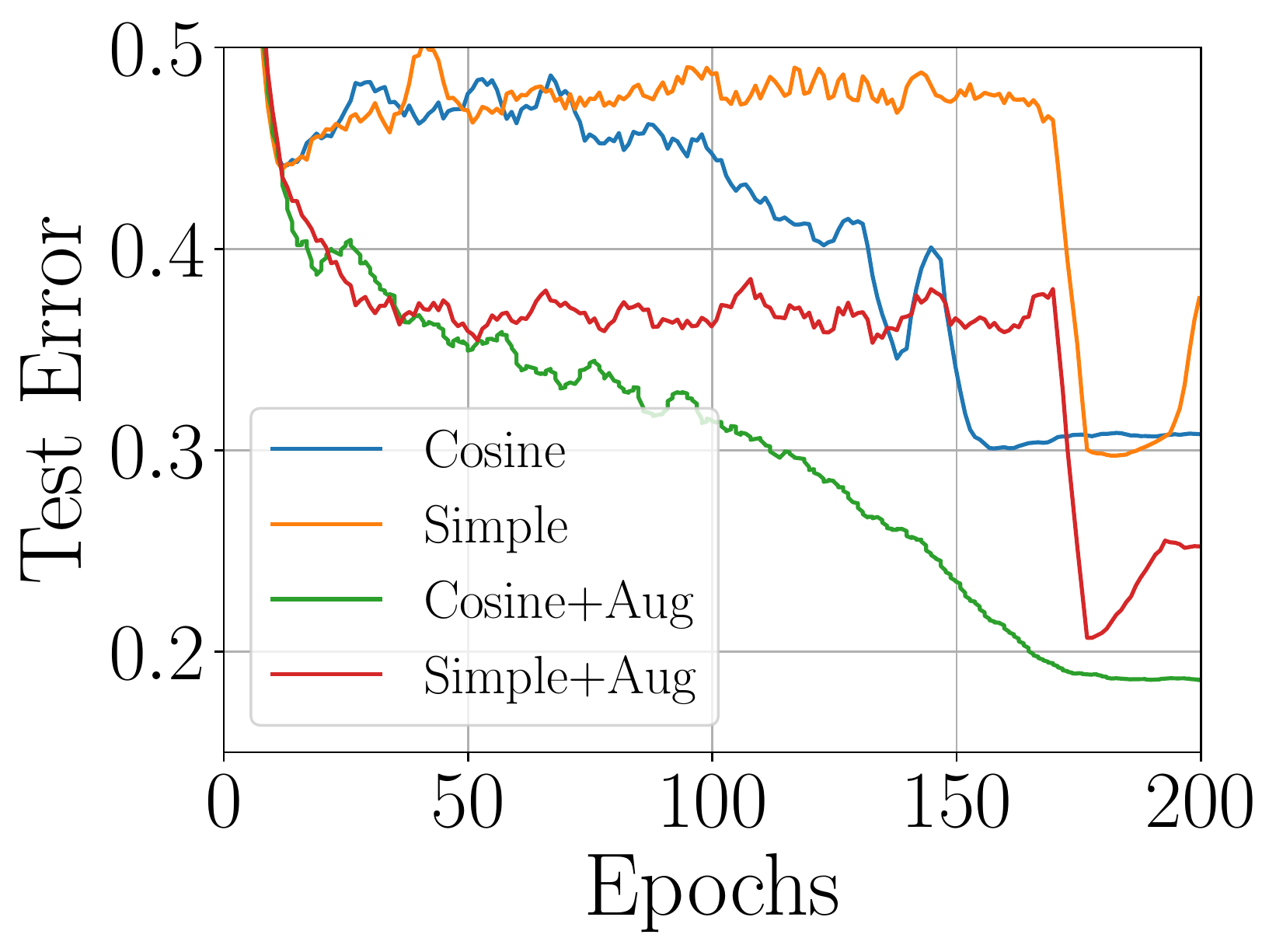}
  }
  \subfloat[Wide Resnet on CIFAR-100]{
    \includegraphics[width=0.33 \textwidth]{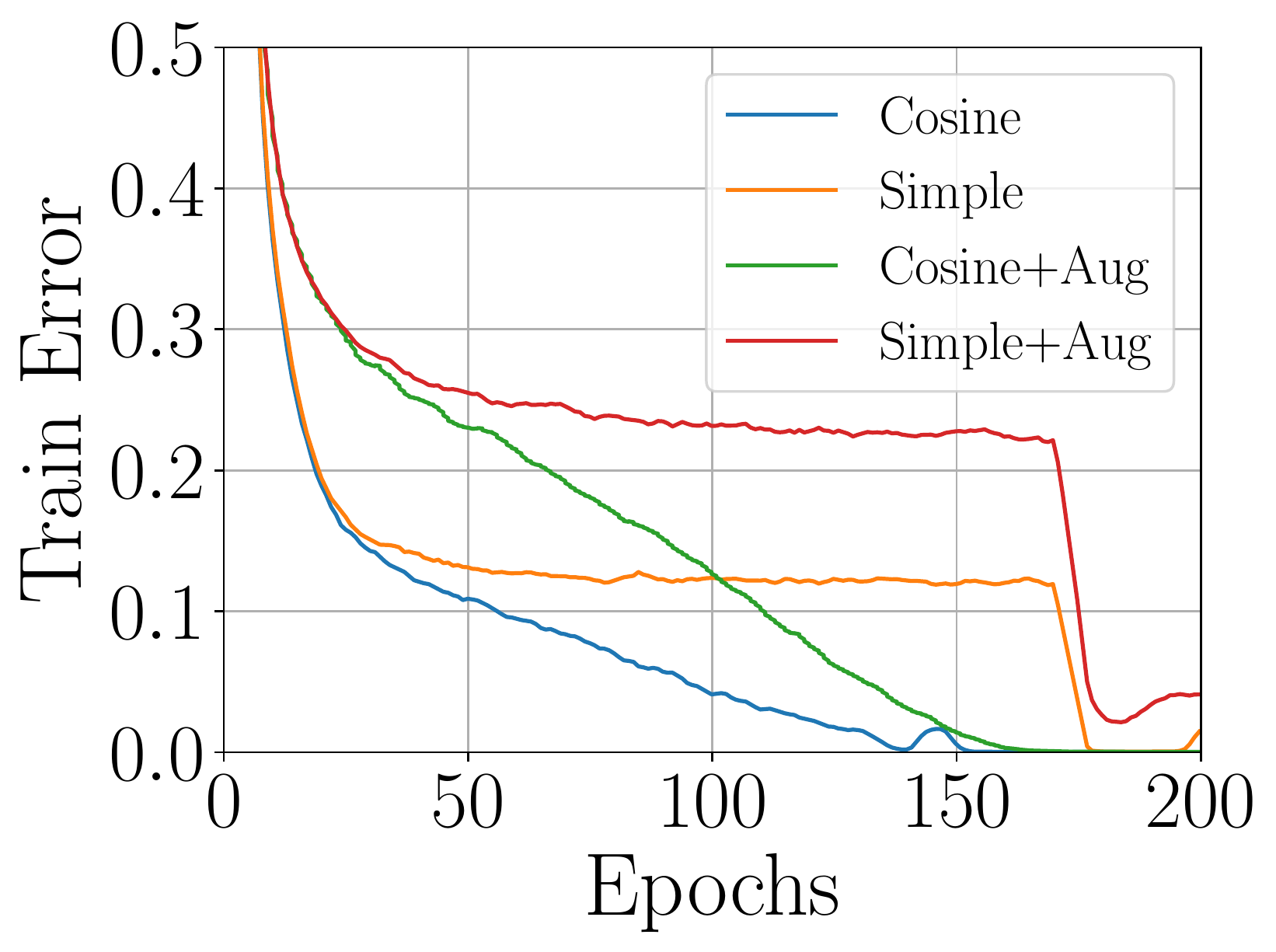}
    
  }
    \subfloat[Wide Resnet on CIFAR-100]{
    \includegraphics[width=0.33 \textwidth]{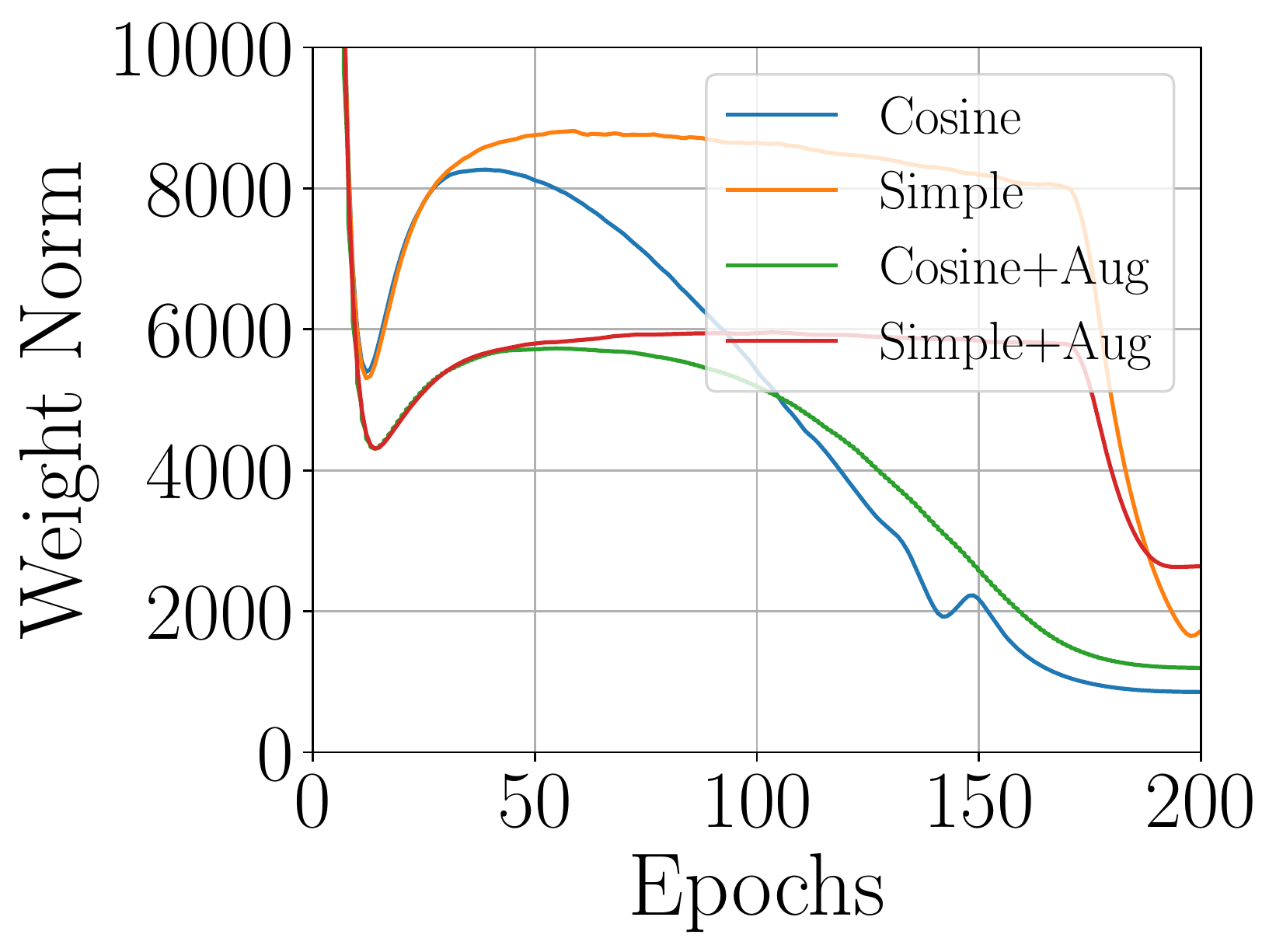}
    
  }
    \\
    \subfloat[Wide Resnet on CIFAR-10]{
    \includegraphics[width=0.33 \textwidth]{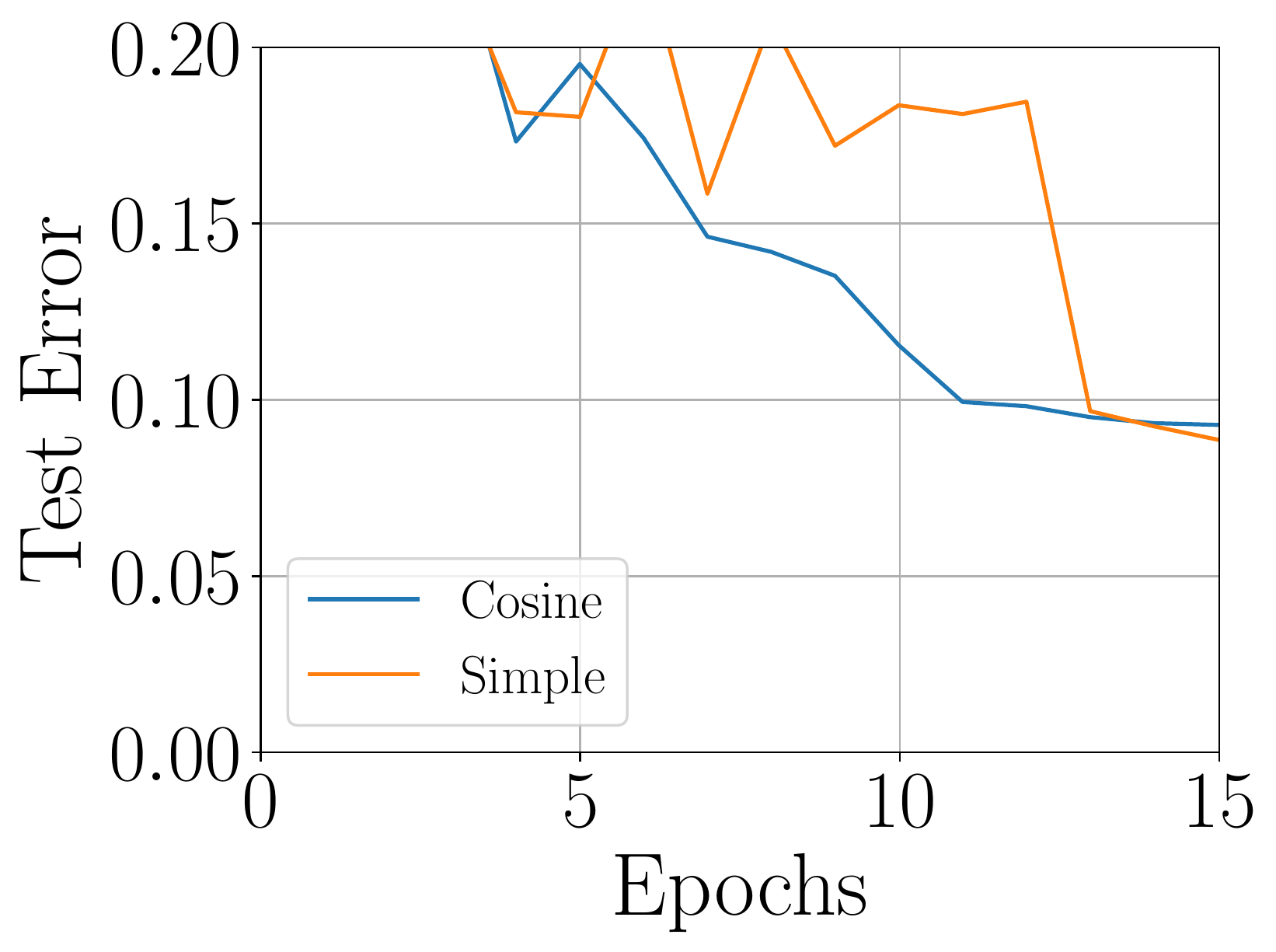}
  }
  \subfloat[Wide Resnet on CIFAR-10]{
    \includegraphics[width=0.33 \textwidth]{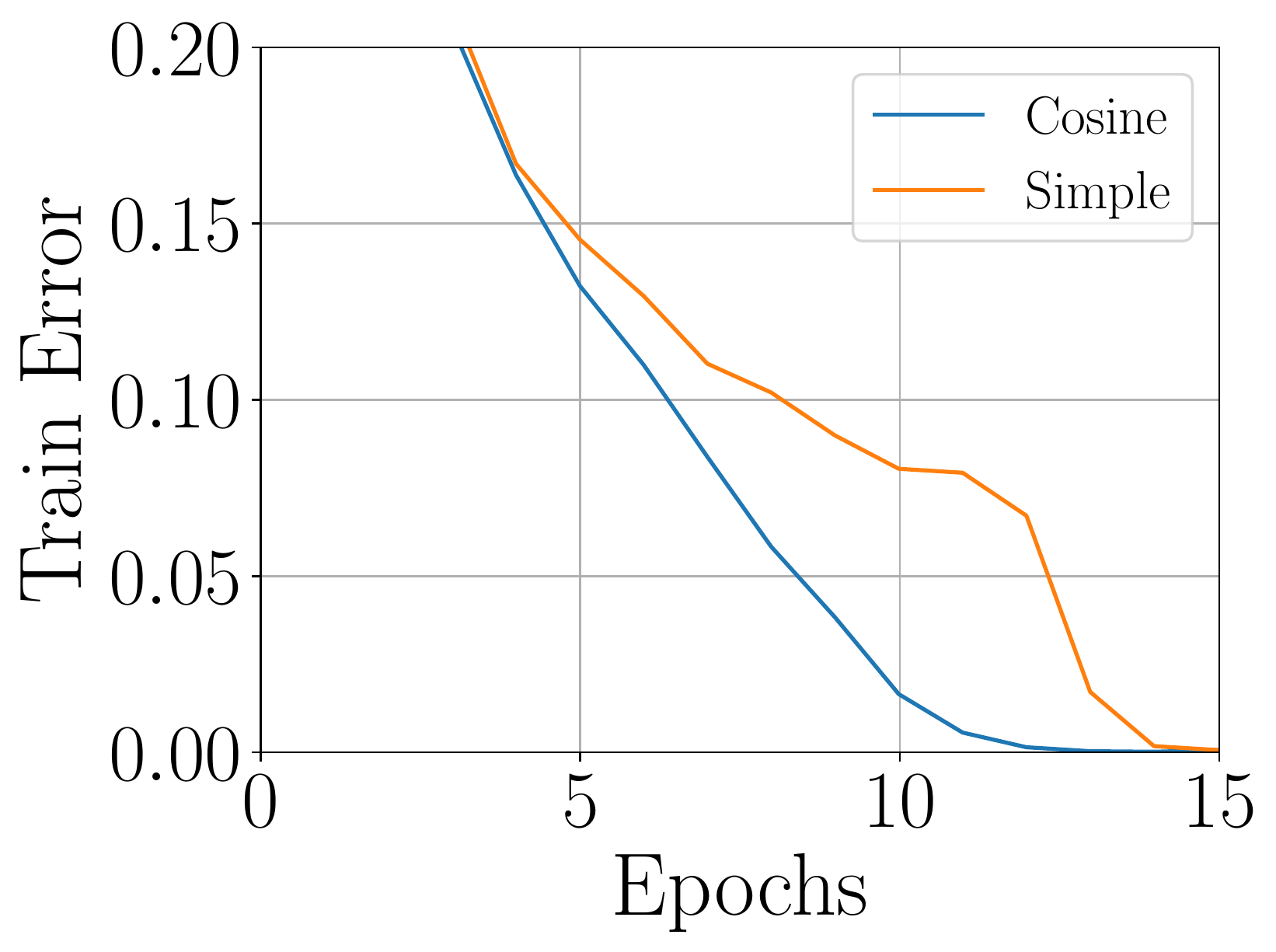}
    
  }
    \subfloat[Wide Resnet on CIFAR-10]{
    \includegraphics[width=0.33 \textwidth]{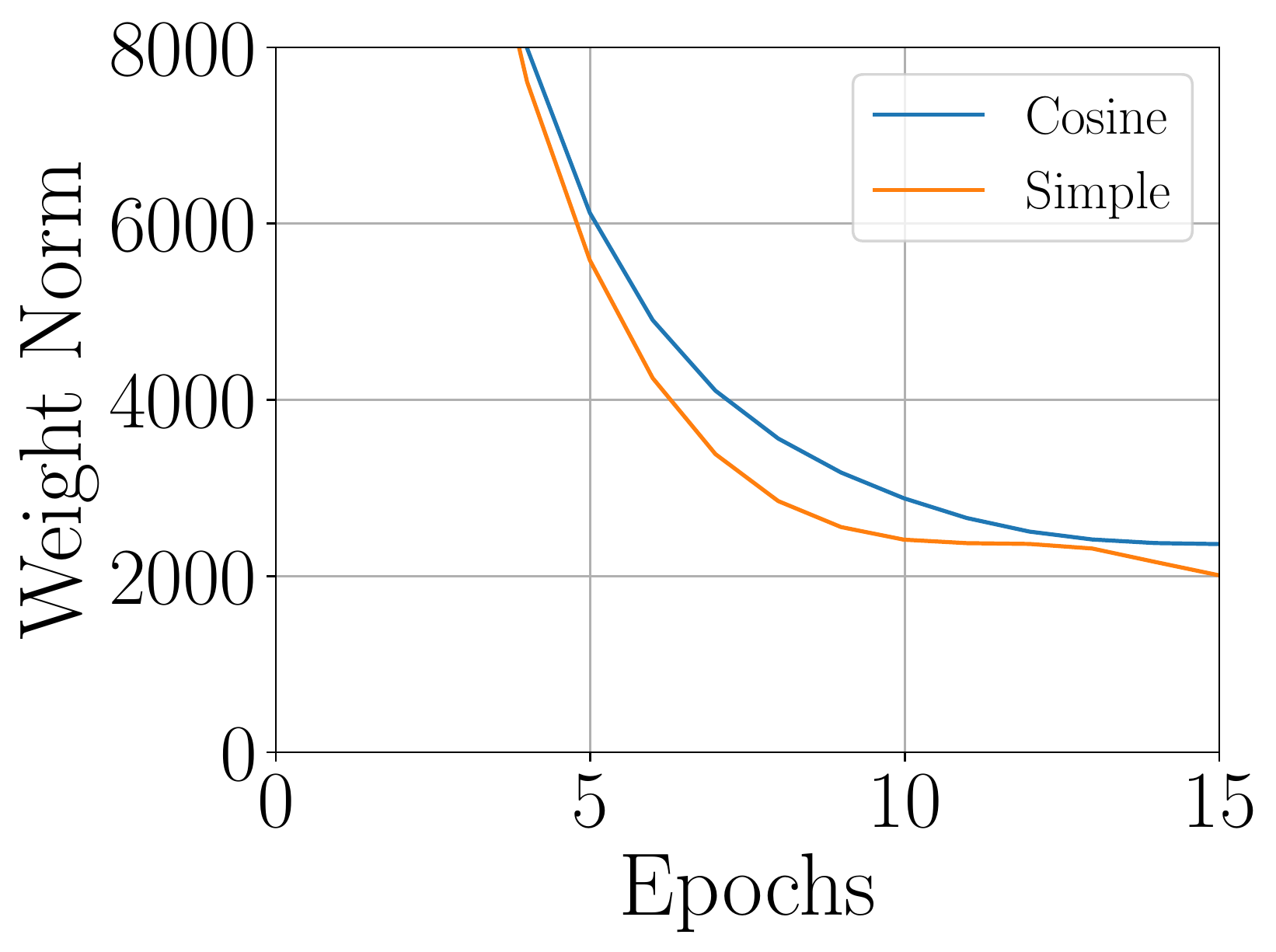}
    
  }
  \caption{Wide Resnets on CIFAR without data augmentation can reach zero training error with simple decay.  }
  \label{fig:noaug}
\end{figure}

\newpage
\subsection{Learning rate dependence of bouncing}
\label{SM:small_lr}
\begin{figure}[ht!]
  \centering
  \subfloat[VGG-16 on CIFAR-10]{
    \includegraphics[width=0.31 \textwidth]{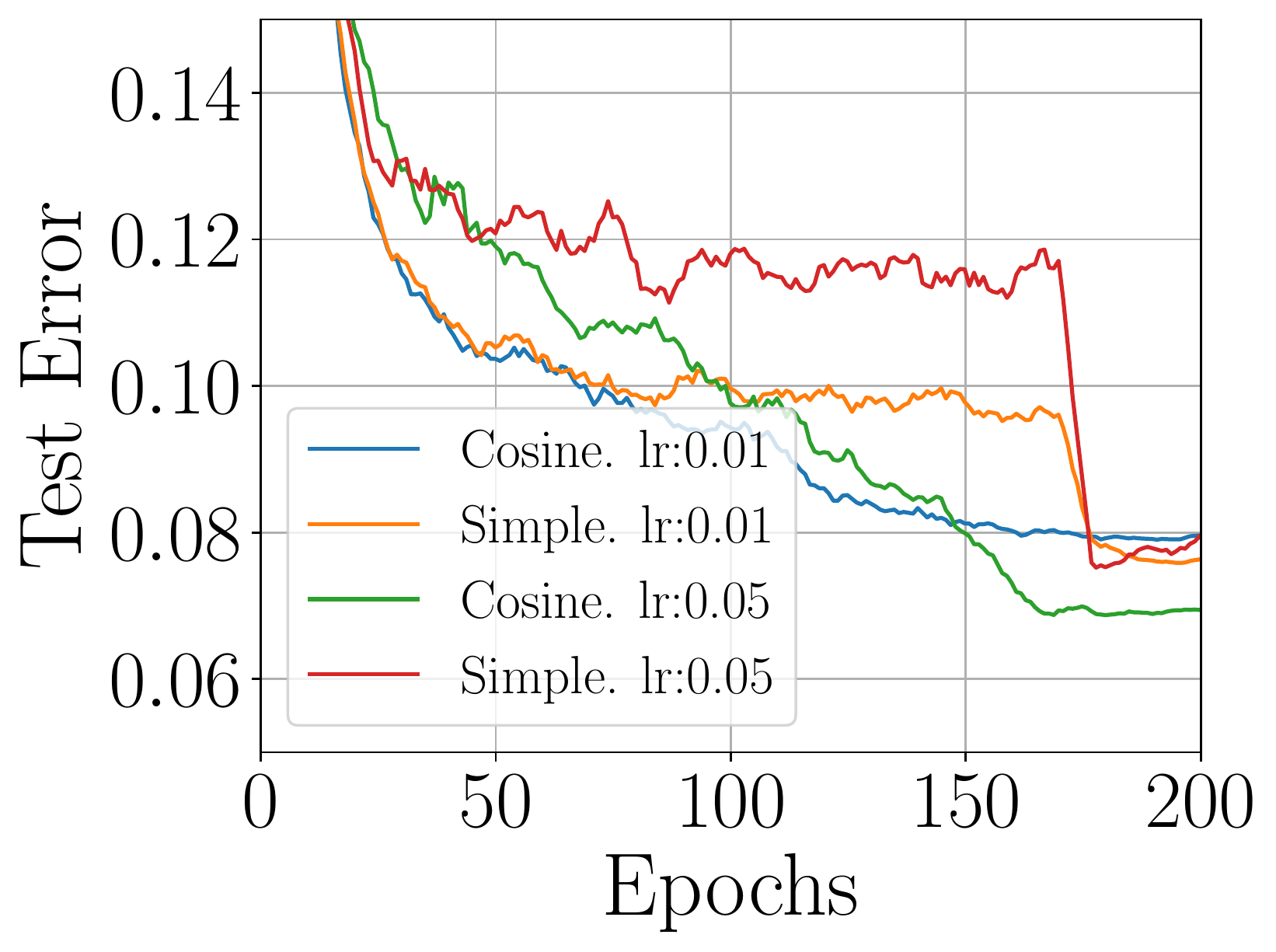}
  }
  \subfloat[VGG-16 on CIFAR-10]{
    \includegraphics[width=0.31 \textwidth]{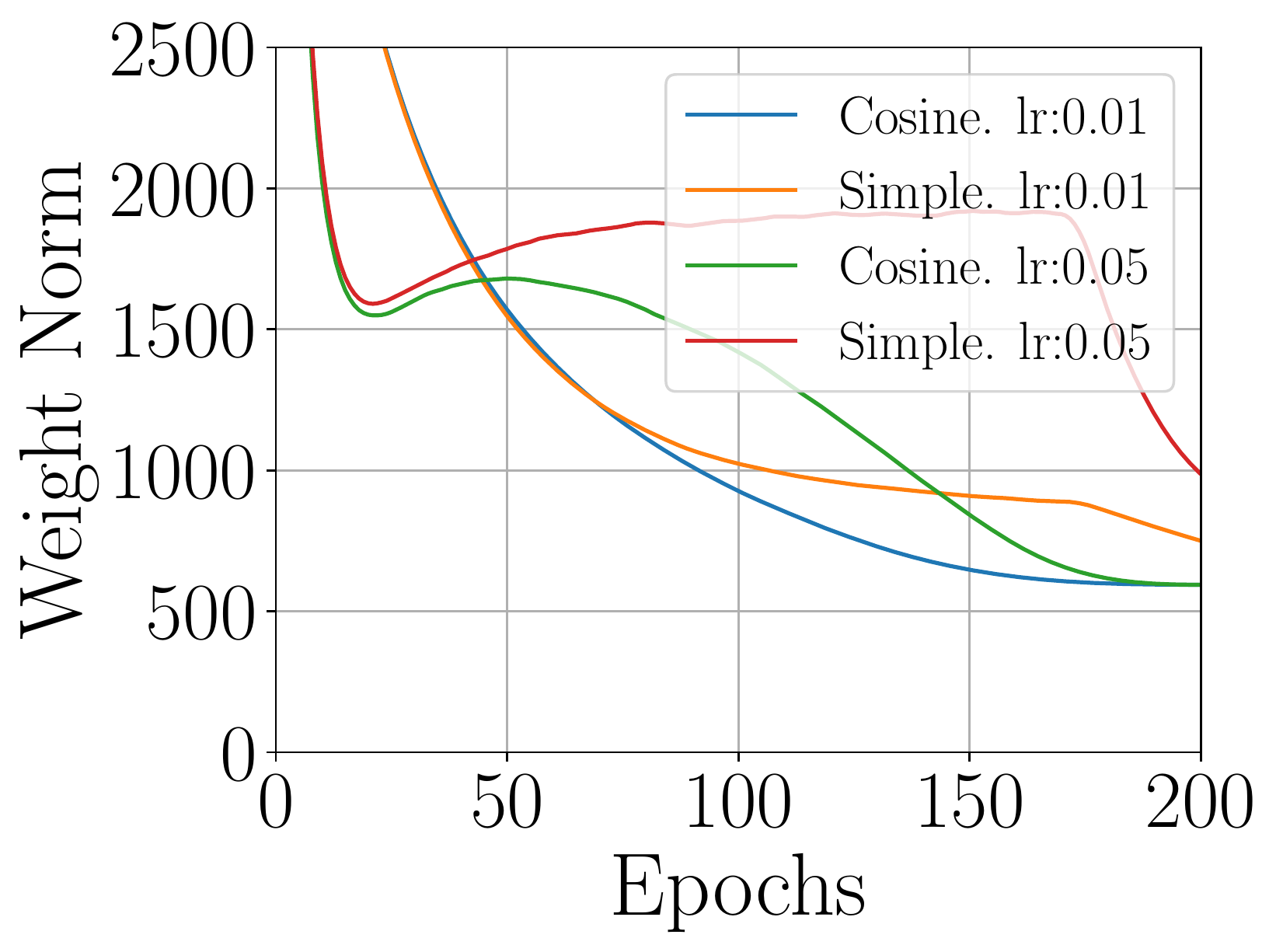}
}
    \\
   \subfloat[VGG-16 on CIFAR-10 1000epochs]{
    \includegraphics[width=0.31 \textwidth]{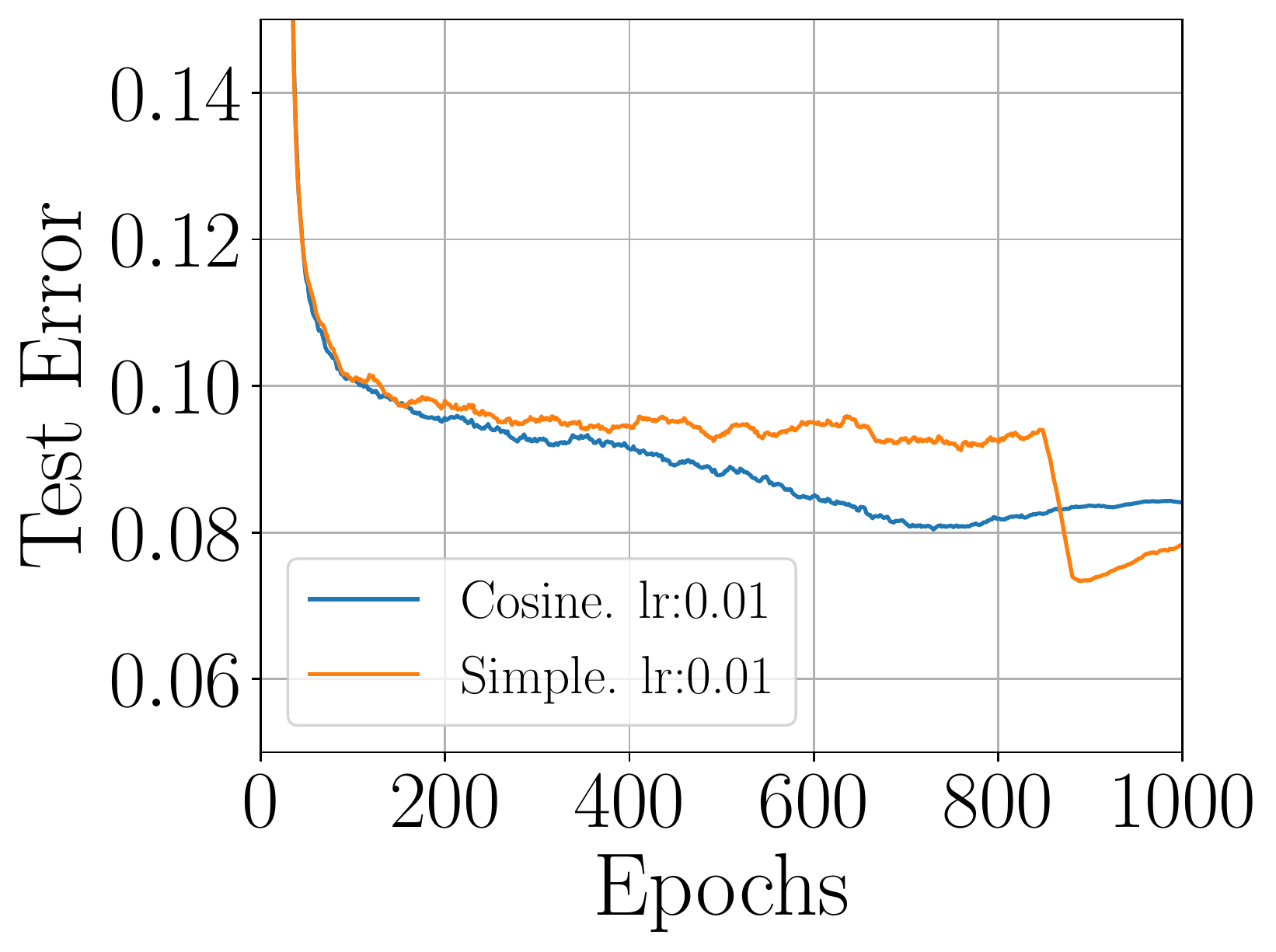}
  }
  \subfloat[VGG-16 on CIFAR-10 1000epochs]{
    \includegraphics[width=0.31 \textwidth]{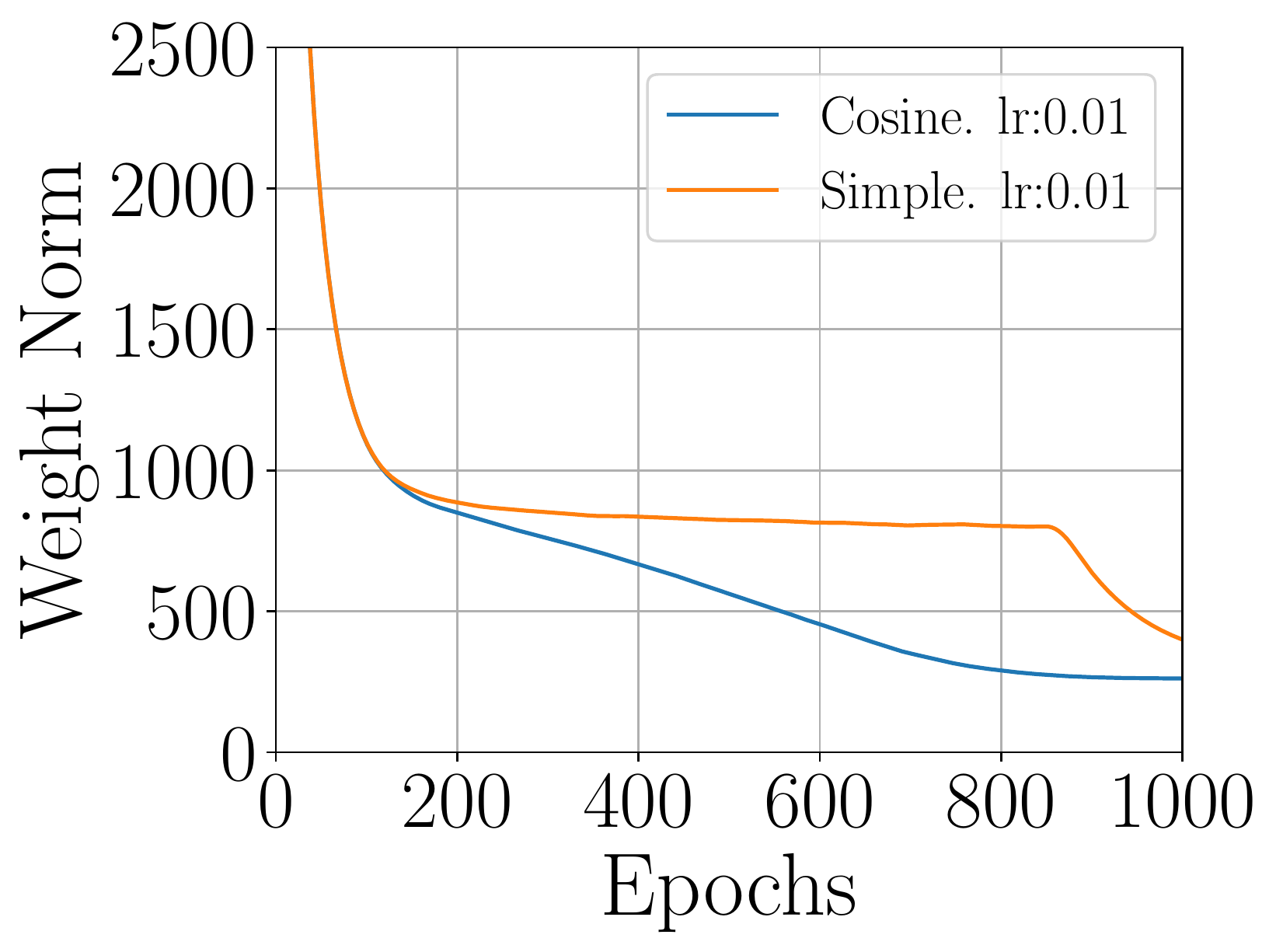}
    }
  \caption{In order for the weight norm to bounce in our training run, the learning rate has to be big enough. We see how for learning rate $0.01$ weight norm does not bounce. This is true even if we evolve for five times longer. }
  \label{fig:small_lr}
\end{figure}

We can study the effect of the learning rate. Large learning rates seem important for weight norm bouncing as we can see in figure \ref{fig:small_lr}, the small learning rate of $0.01$ does not have a bouncing norm even if we evolve for longer (since smaller learning often to take a longer time to train). We see that only when there is a bouncing norm are cosine schedules benefitial, see table \ref{table:VGG_small}.  
\begin{table}[htb]

 \centering
\begin{tabular}{ |l||r|r|  }
 \hline
   Setup  & Simple Decay & Cosine Decay    \\
 \hline
lr$=0.05$ & 7.4 & 6.7\\
lr$=0.01$ & 7.4 & 7.8 \\
lr$=0.01$ 1000 epochs & 7.2 & 7.6 \\
 \hline
\end{tabular}

\hspace*{0.5cm}
\caption{Test error for VGG-16 and CIFAR-10 for different small learning rate setups. Cosine decay is only superior at large learning rates when there is a bouncing weight norm.}
\label{table:VGG_small}
\end{table} 

\newpage
\subsection{Bouncing for different initialization scales.}
\label{sec:init_scales}
The bouncing behaviour suggests that maybe one can avoid bouncing at all by starting with the minimum weight norm to begin with. In order to check this, we run our CIFAR-100 WRN28-10 experiments with cosine decay for different initializations for all weights (including the batchnorm weights), which we denote $\sigma_w$, with $\sigma_w=1$ the standard experiments that we have run. We see how, even if we start with a weight norm which is below the minimum weight norm for $\sigma_w=1$, there is still bouncing. As we keep decreasing $\sigma_w$, it gets to a point where performance gets significantly degraded (and bouncing dissappears too). These small $\sigma_w$ are probably pathological, but it is interesting to see again the correlation between degraded performance and no bouncing. Before reaching this small weight initialization the final weight norm (and error rate) is very similar for the different initializations.

  \begin{figure}[h!]
  \centering
  \subfloat[WRN 28-10 on CIFAR-100]{
    \includegraphics[width=0.33 \textwidth]{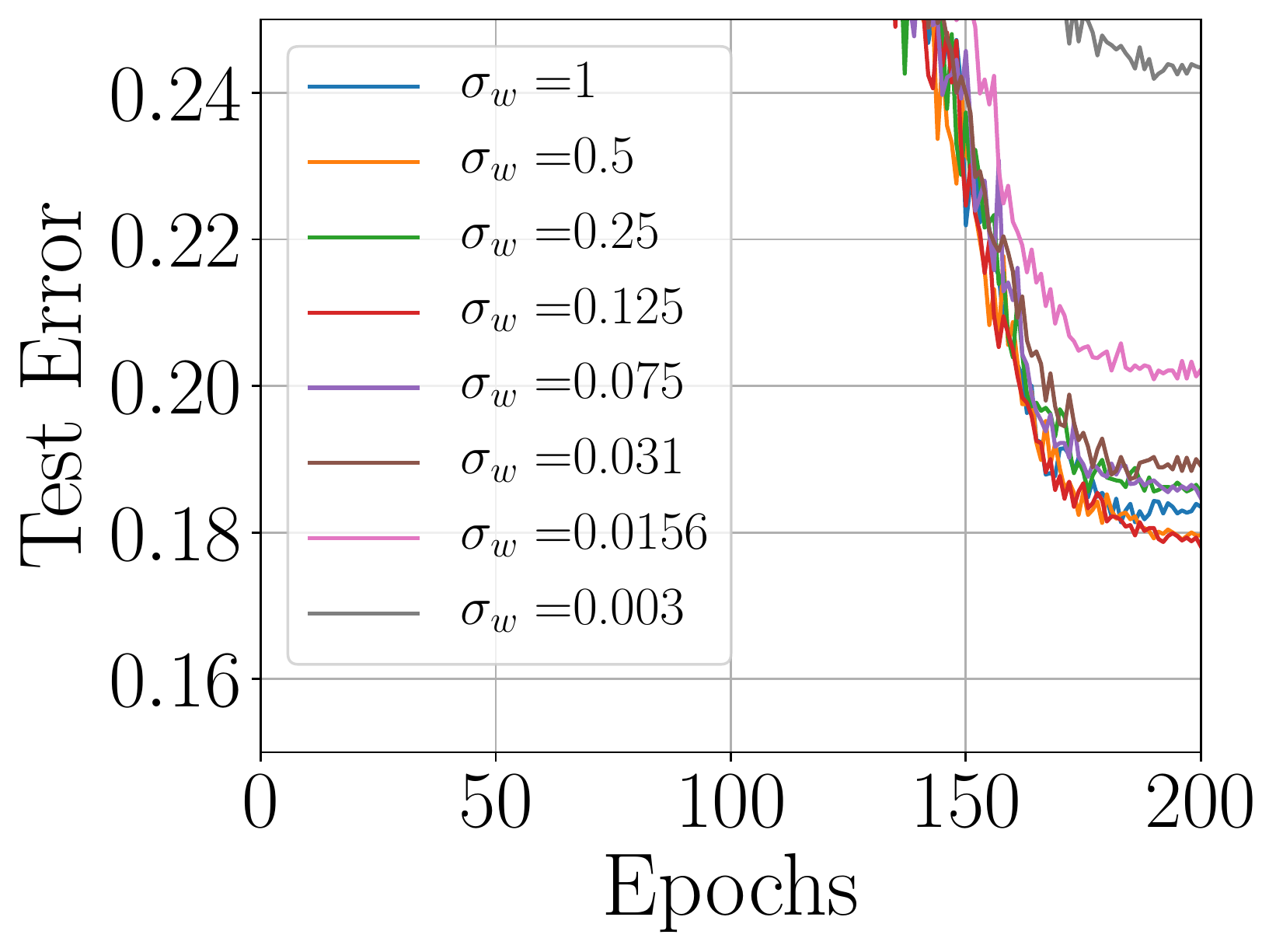}
  }
  \subfloat[WRN 28-10 on CIFAR-100]{
    \includegraphics[width=0.33 \textwidth]{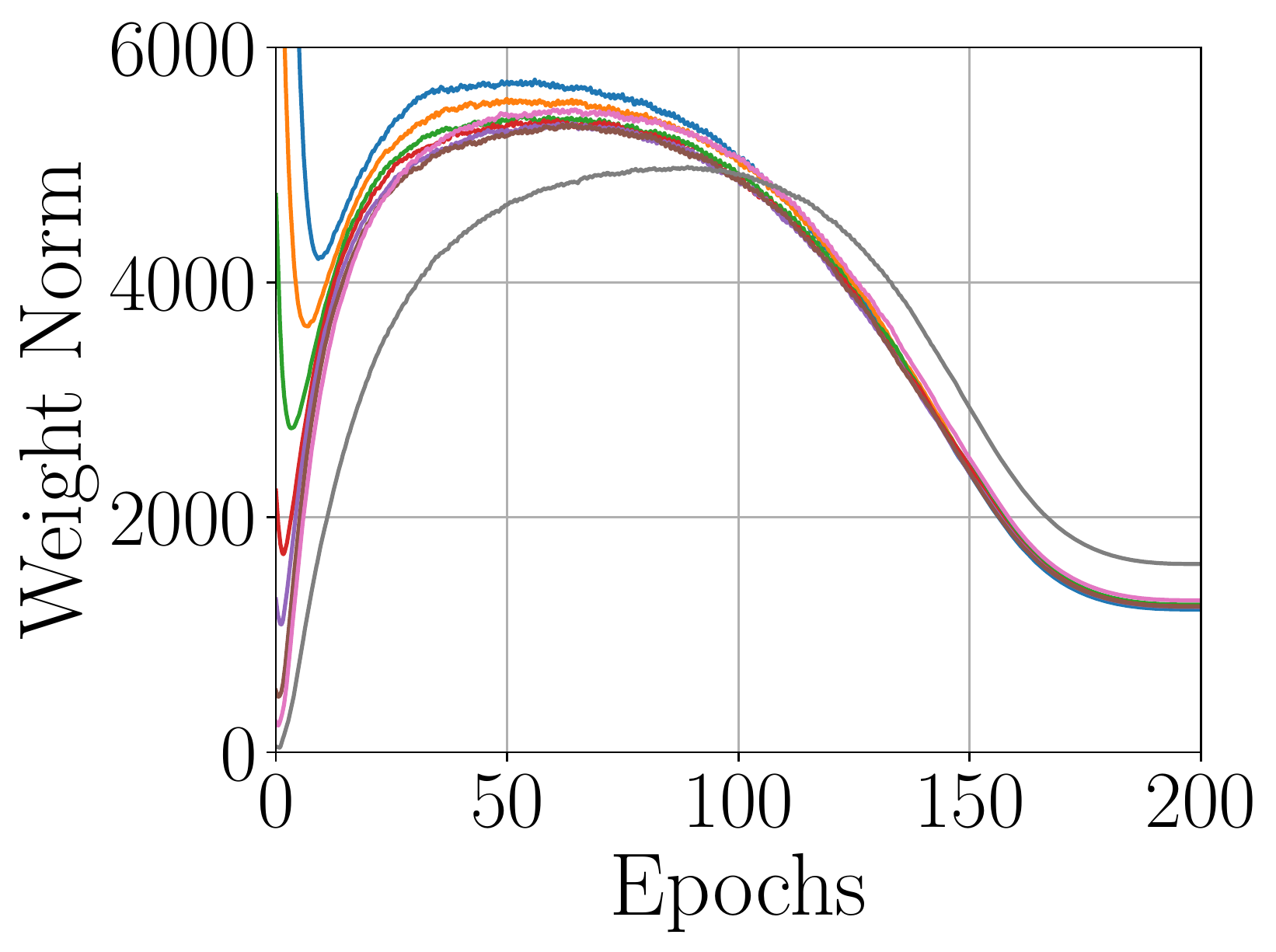}
  }
      \subfloat[WRN 28-10 on CIFAR-100]{
    \includegraphics[width=0.33 \textwidth]{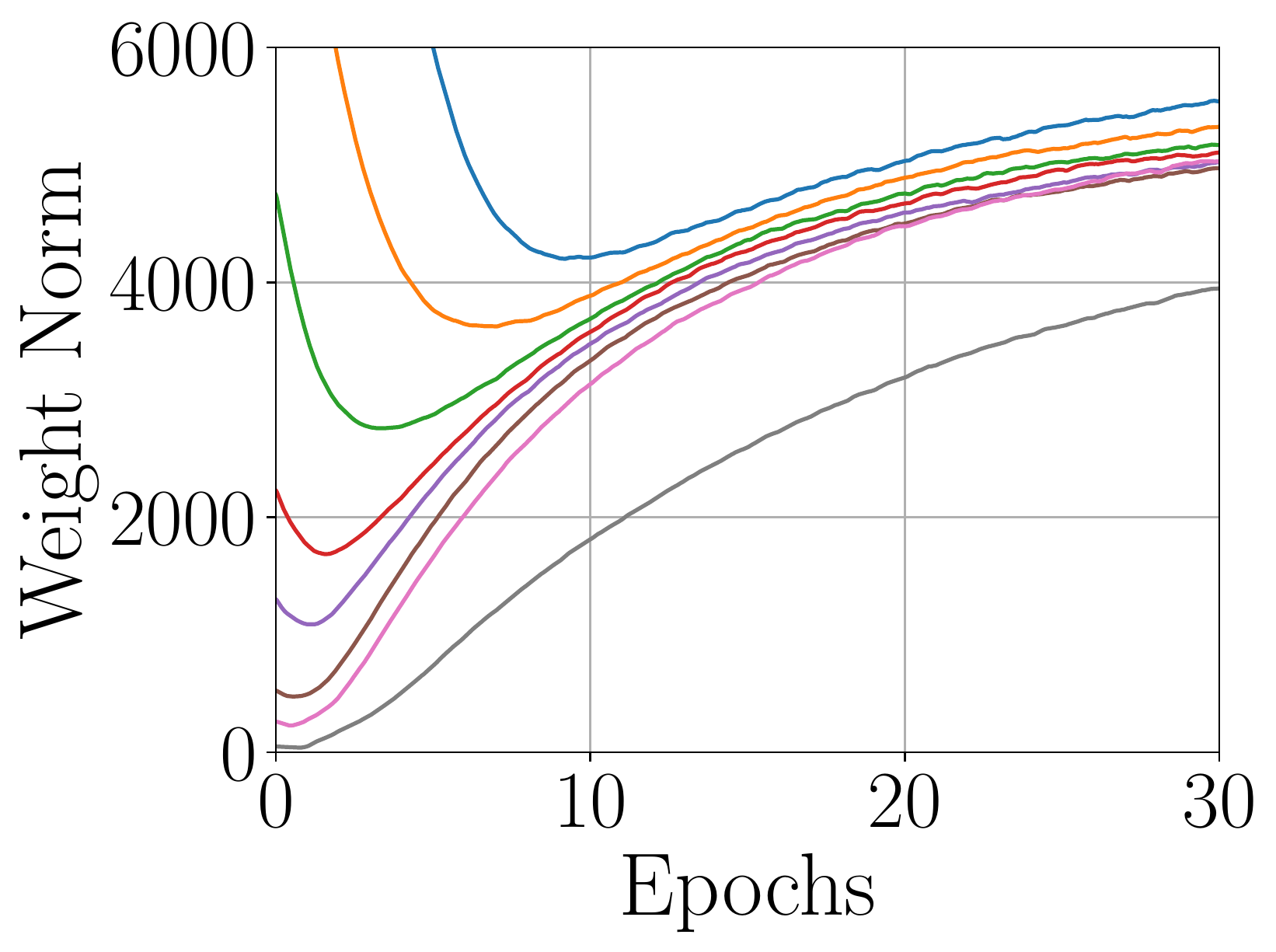}
  }
  \caption{Wide Resnet 28-10 on CIFAR100 expeirents with different initialization scale. a) Test error, we see that the two smallest initialization present substantial degradation of performance. b) Despite of the weight norm being so different at initialization, the final weight norm is the same for the different models. c) Zoomed version of b) we see that all the initializations but the smallest two exhibit clear bouncing. }
  \label{fig:bouncing_init}
\end{figure}

\clearpage
\section{More training curves}
\subsection{Training curves for Table \ref{table:table1} experiments}
\label{SM:training}
We show the remaining training curves for the Table 1 experiments.
\begin{figure}[ht!]
  \centering
  \subfloat[Wide Resnet on CIFAR-10]{
    \includegraphics[width=0.31 \textwidth]{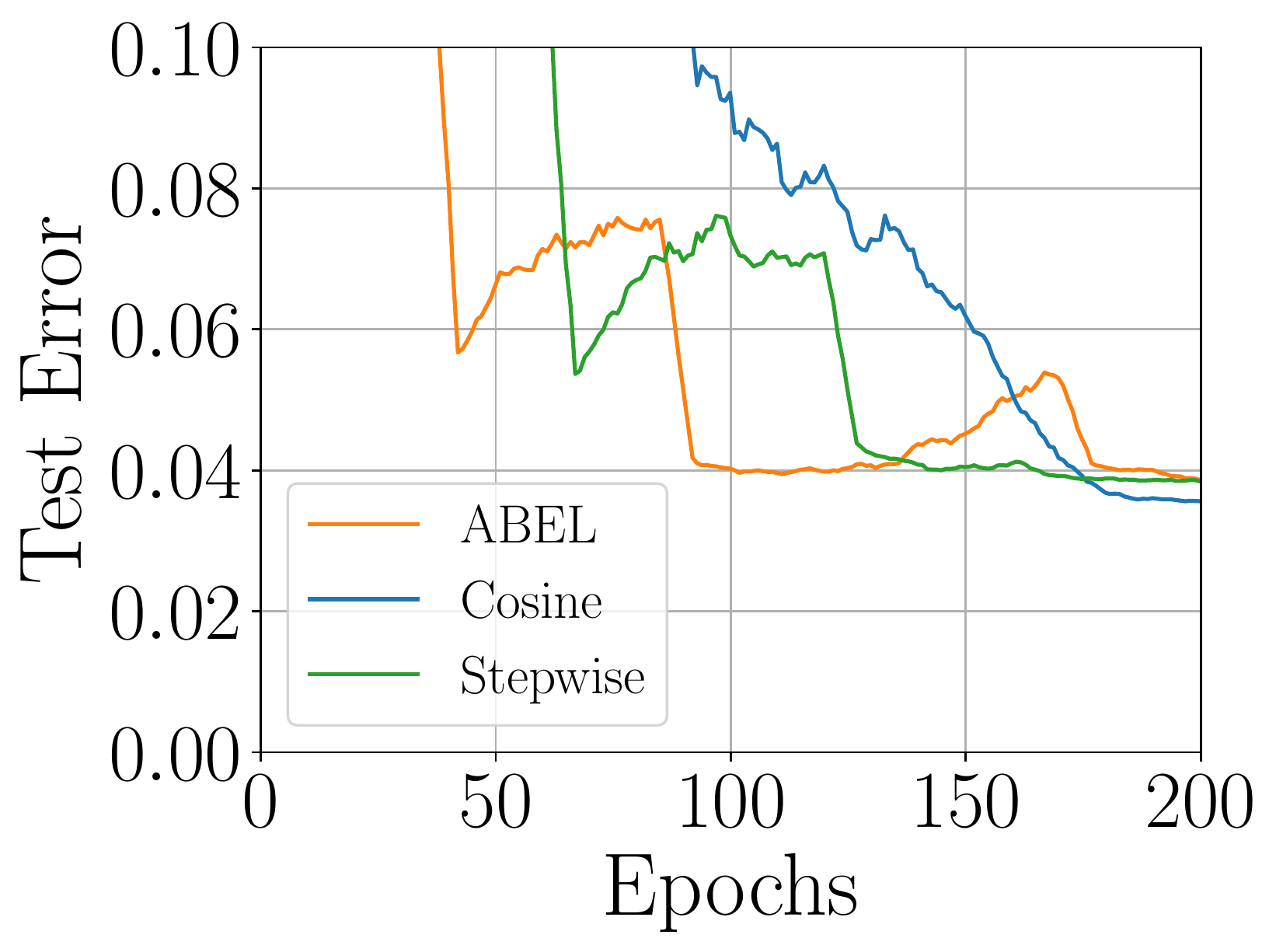}
  }
  \subfloat[Wide Resnet on CIFAR-10]{
    \includegraphics[width=0.31 \textwidth]{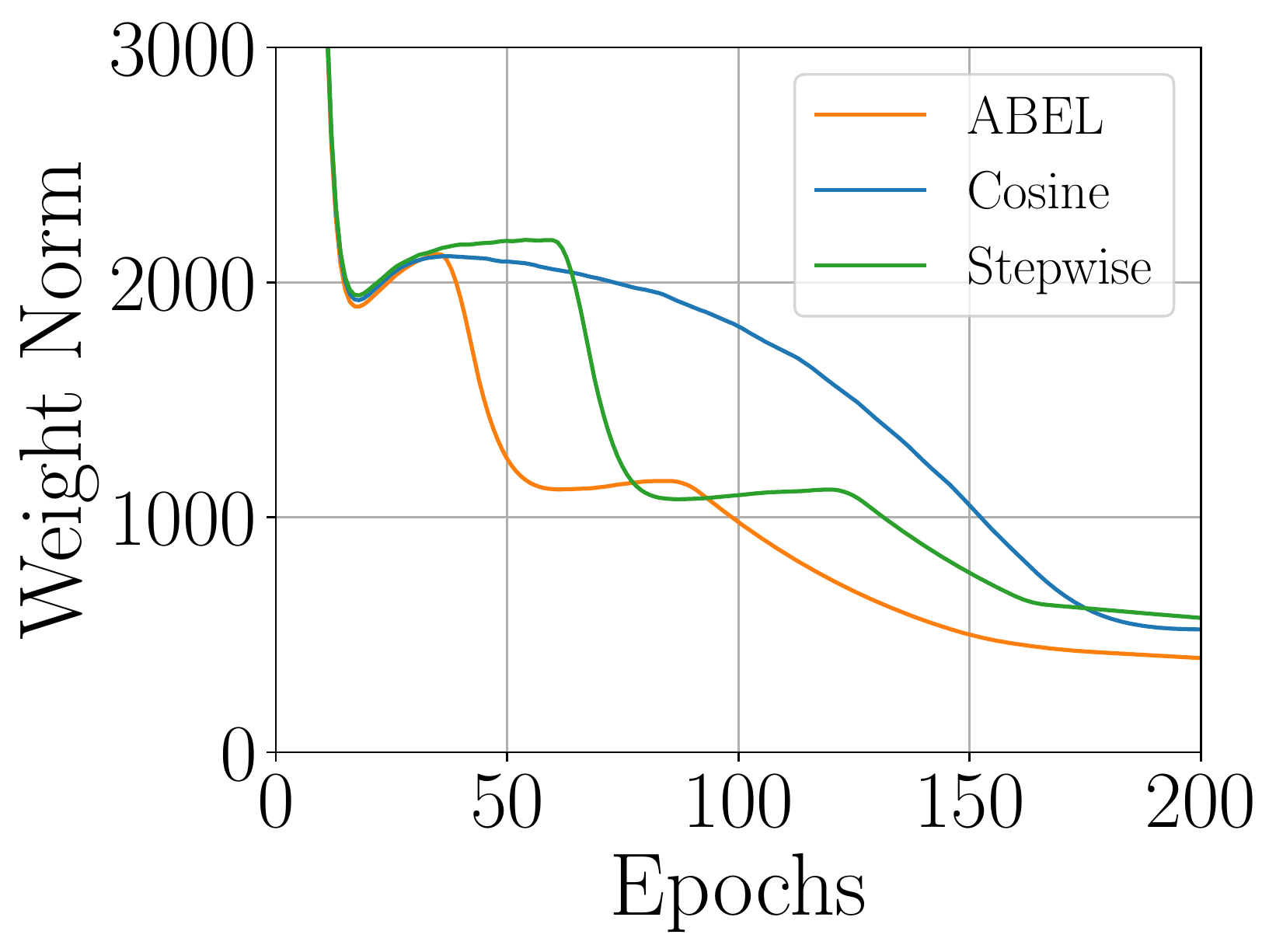}
    
  }
  \subfloat[Wide Resnet on CIFAR-10]{
    \includegraphics[width=0.31 \textwidth]{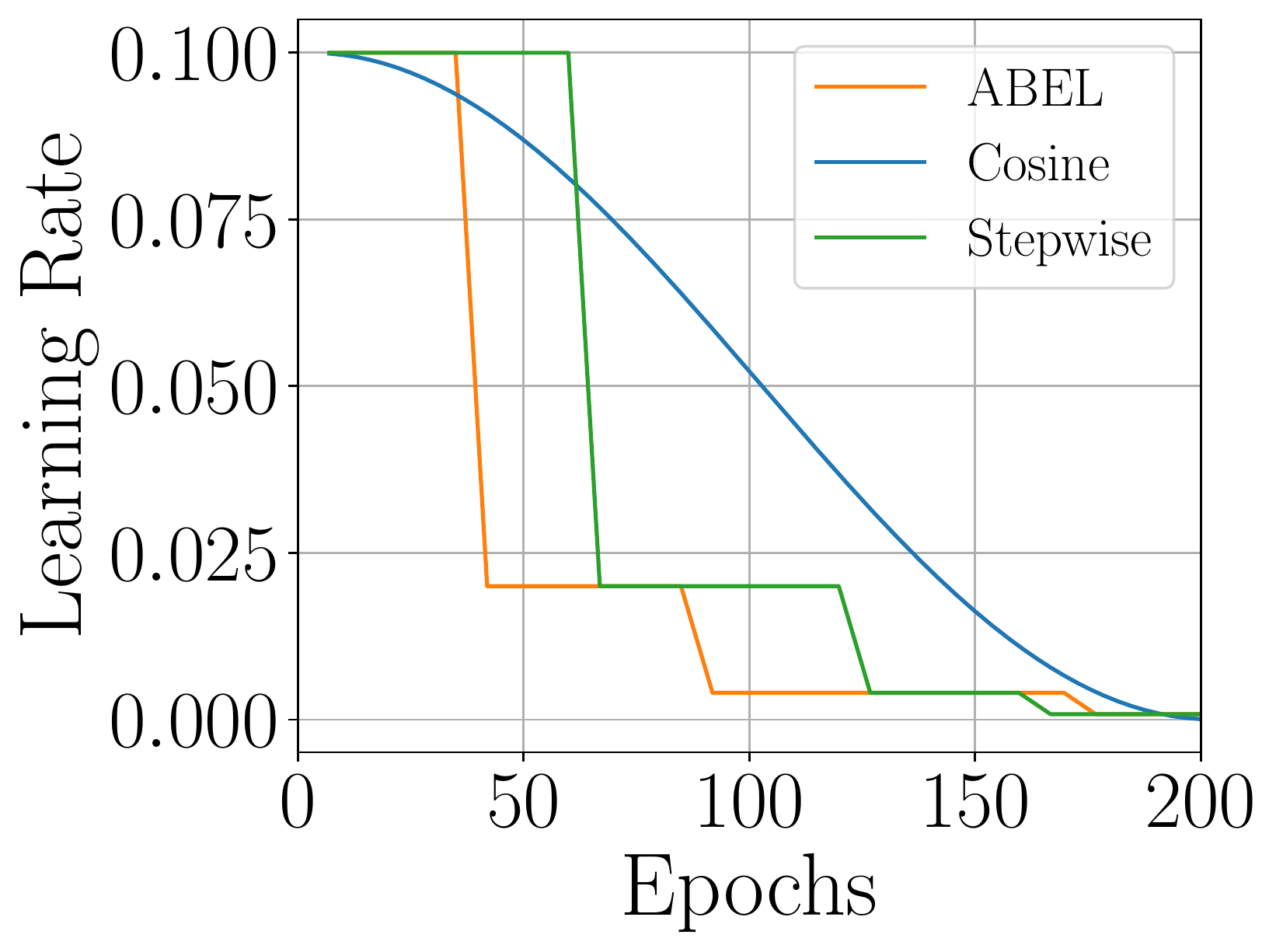}
  }
  \\
  
    \subfloat[VGG on CIFAR-10]{
    \includegraphics[width=0.31 \textwidth]{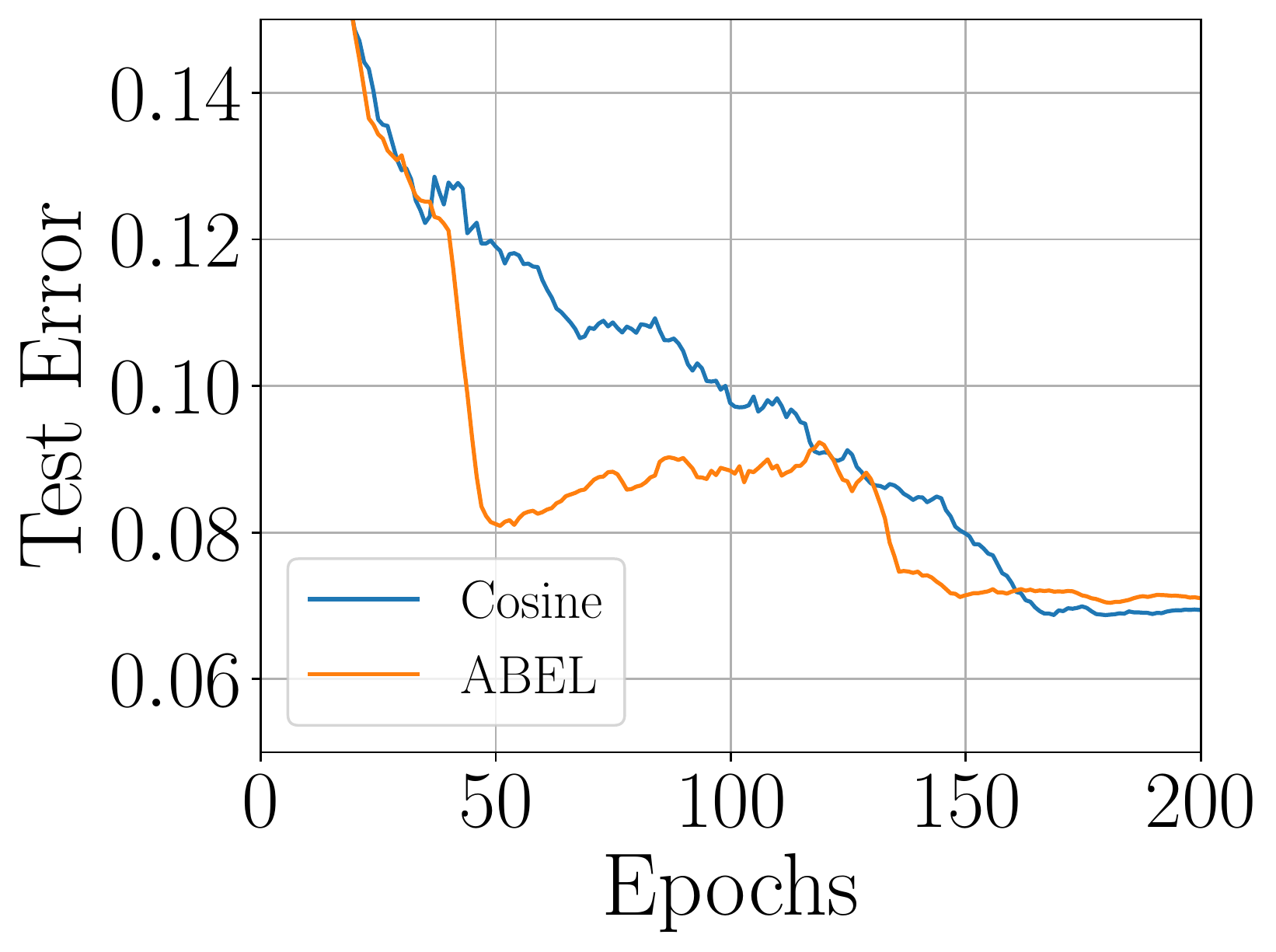}
  }
  \subfloat[VGG on CIFAR-10]{
    \includegraphics[width=0.31 
    \textwidth]{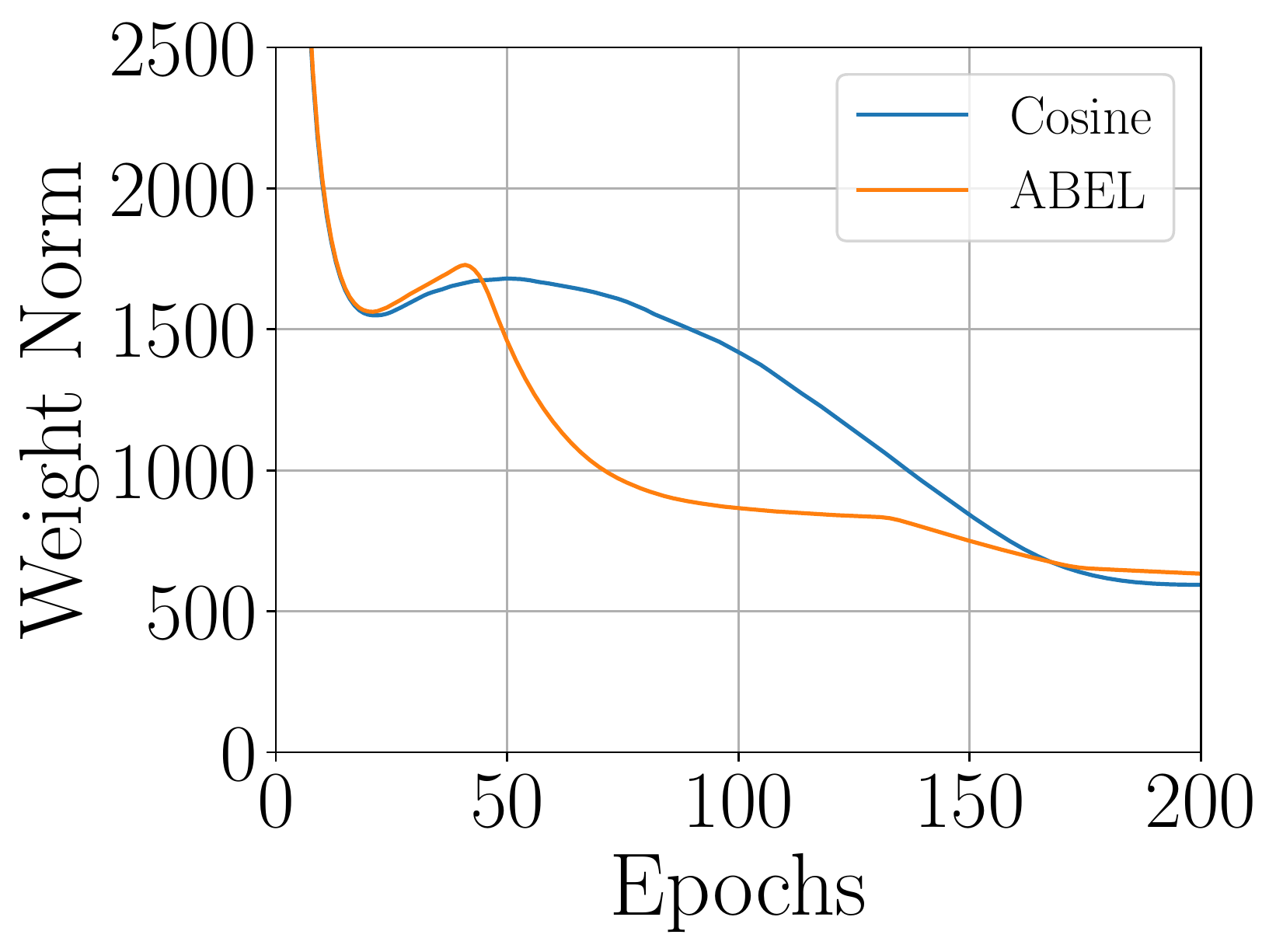}
    
  }
  \subfloat[VGG on CIFAR-10]{
    \includegraphics[width=0.31 \textwidth]{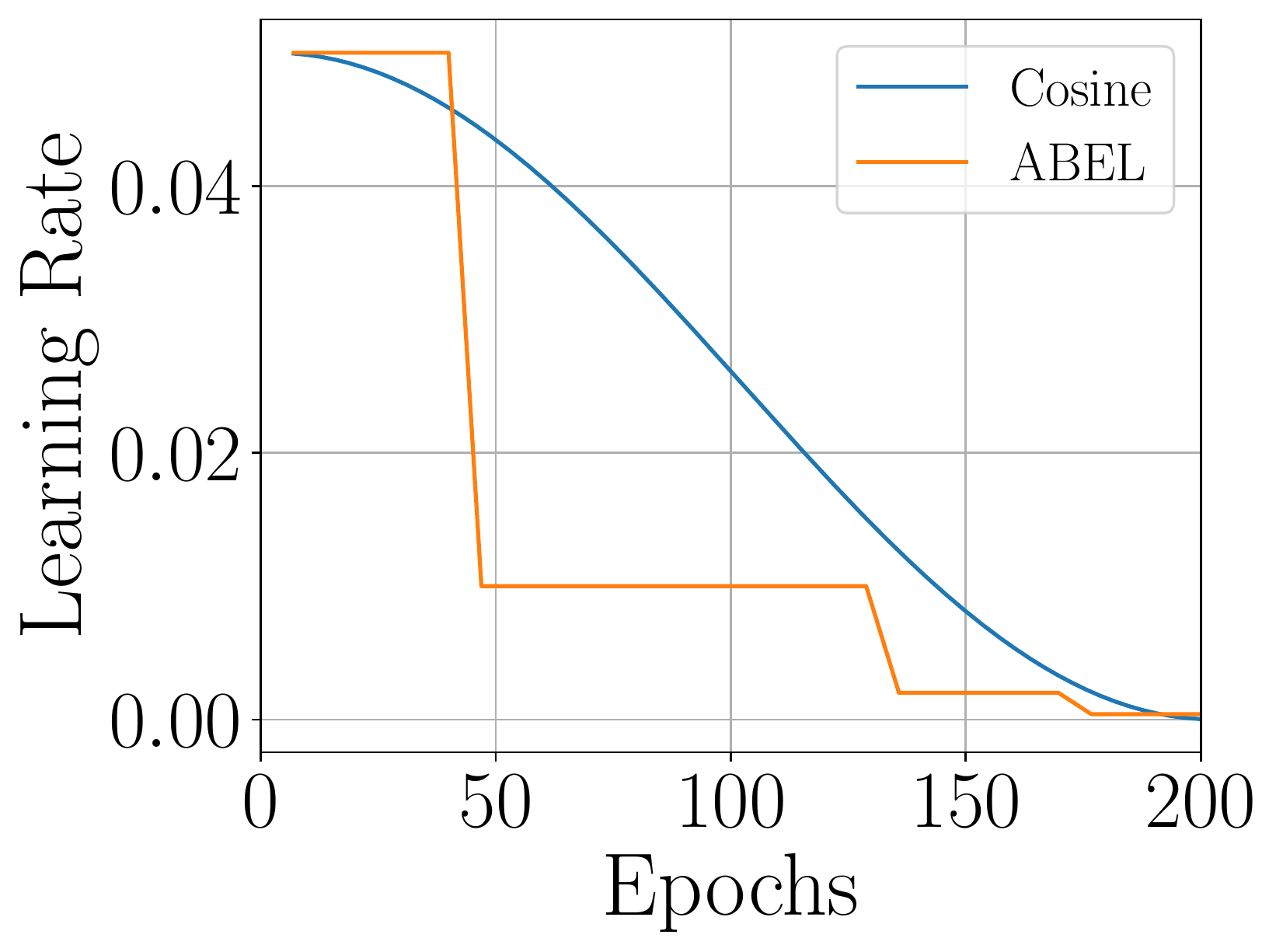}
    
  }
  \\
  \subfloat[Wide Resnet 16-8 on SVHN]{
    \includegraphics[width=0.31 \textwidth]{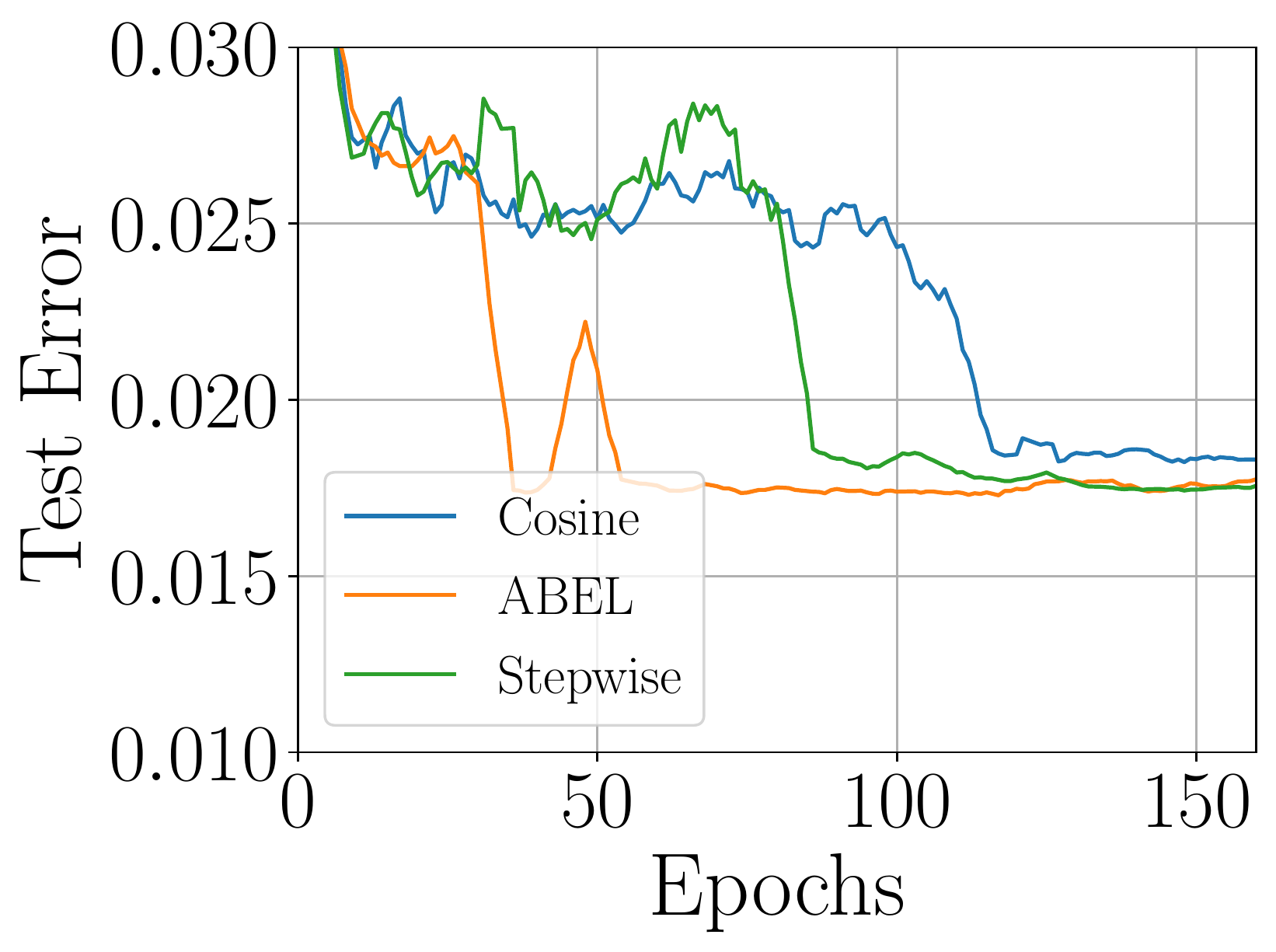}
    
  }
      \subfloat[Wide Resnet 16-8 on SVHN]{
    \includegraphics[width=0.31\textwidth]{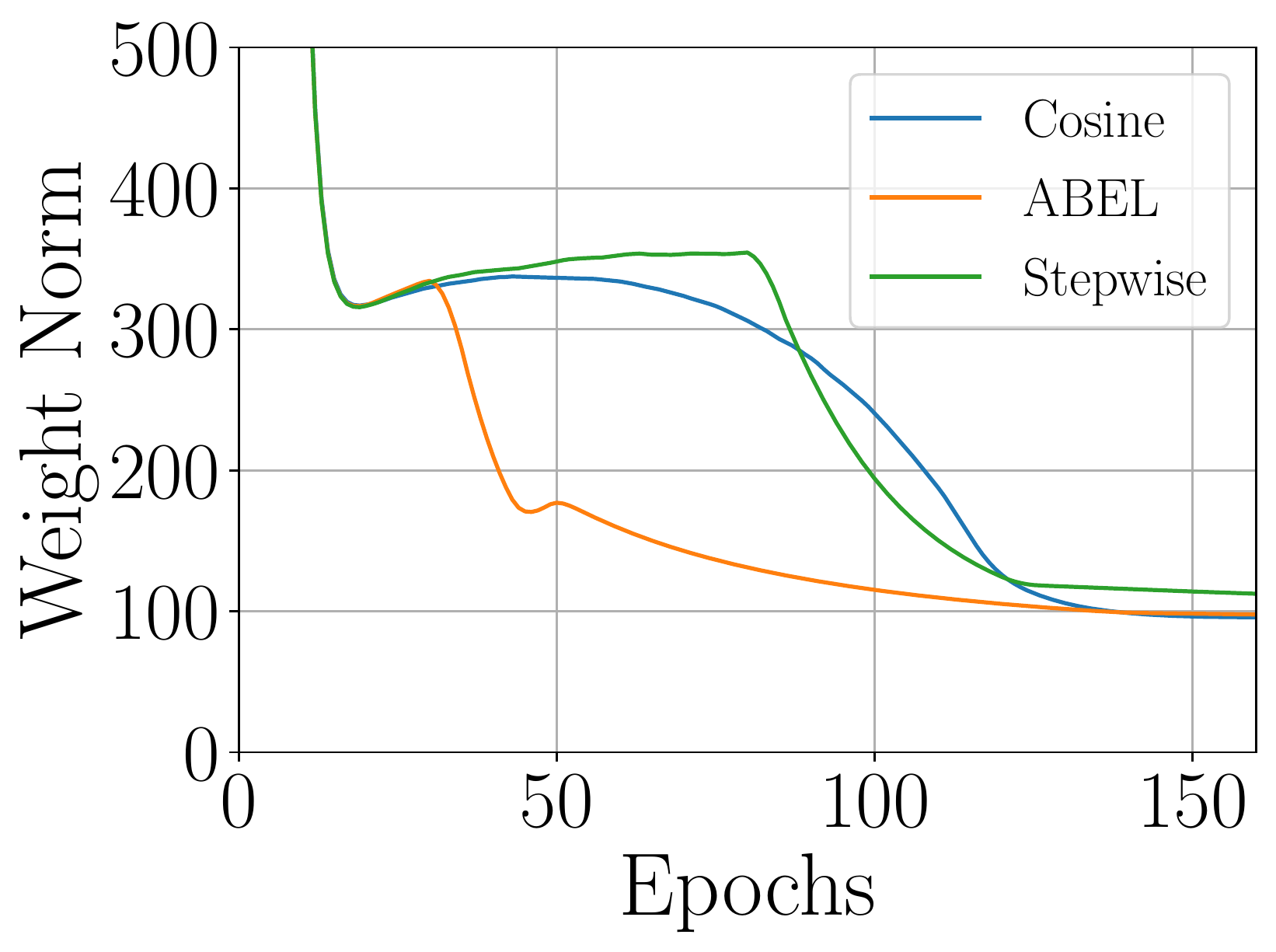}
    
  }
  \subfloat[Wide Resnet 16-8 on SVHN]{
    \includegraphics[width=0.31 \textwidth]{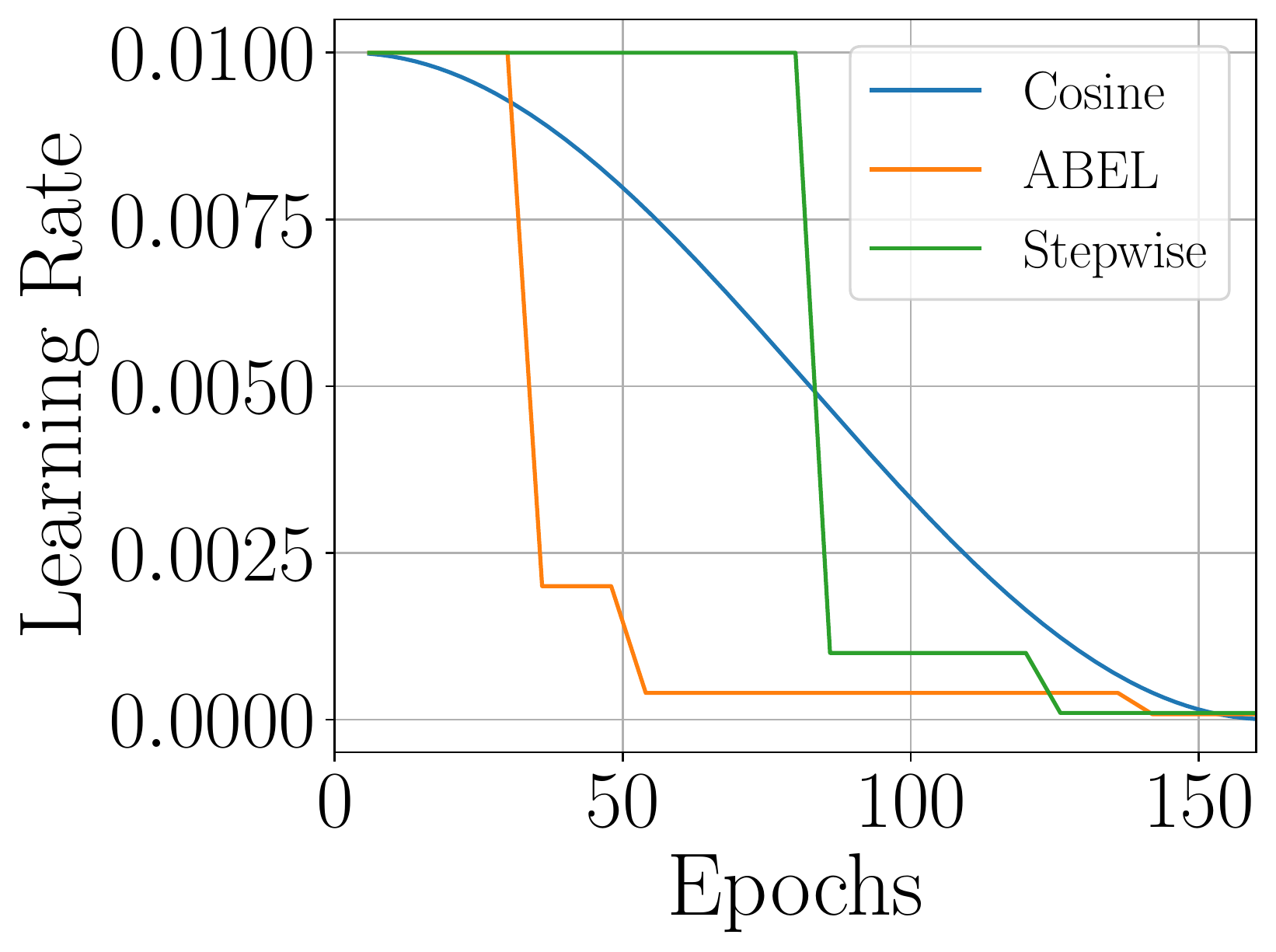}
    
  }
  \\
    \subfloat[PyramidNet on CIFAR100]{
    \includegraphics[width=0.31\textwidth]{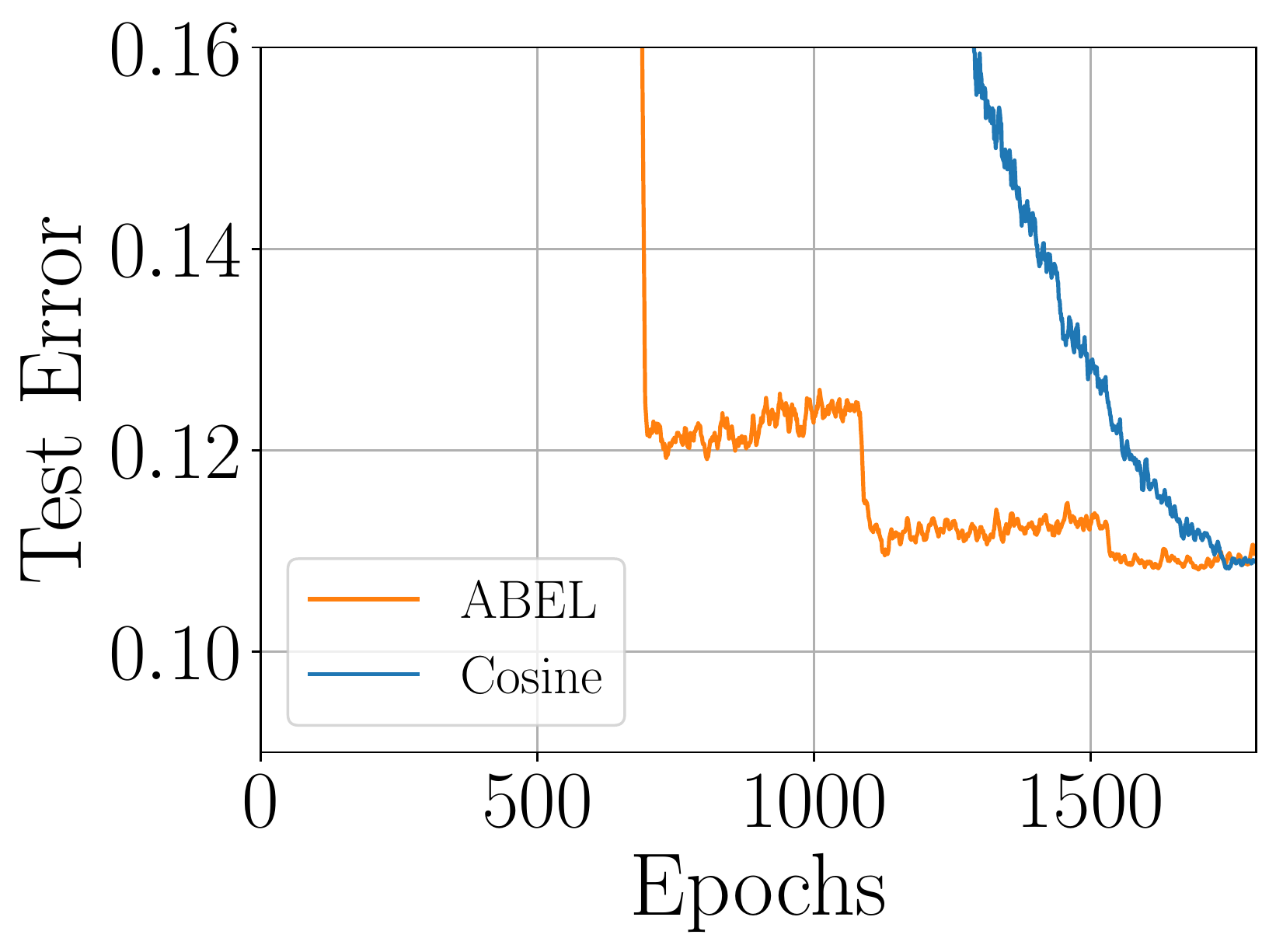}
    
  }
  \subfloat[PyramidNet on CIFAR100]{
    \includegraphics[width=0.31 \textwidth]{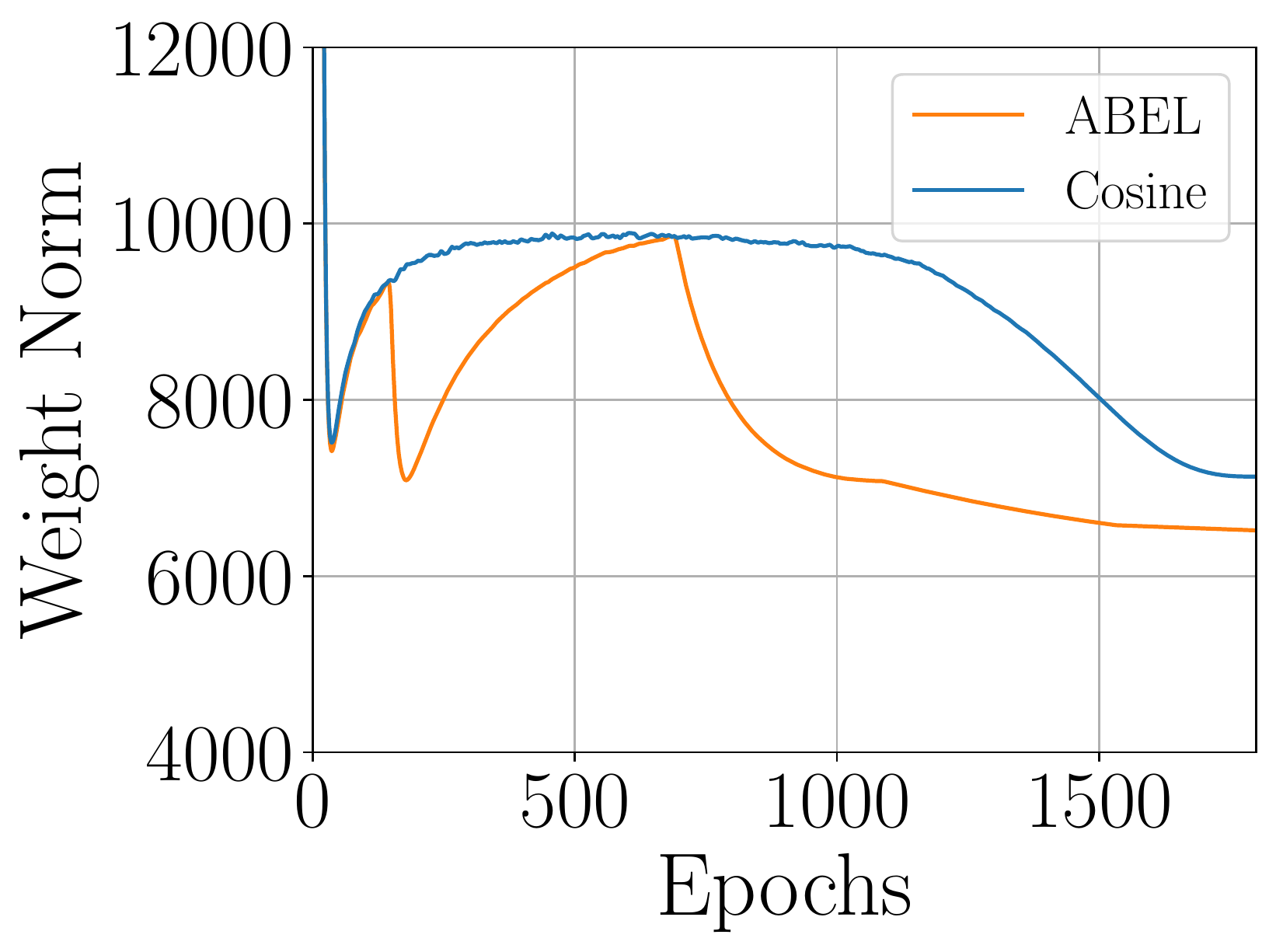}
    
  }
  \subfloat[PyramidNet on CIFAR100]{
    \includegraphics[width=0.31 \textwidth]{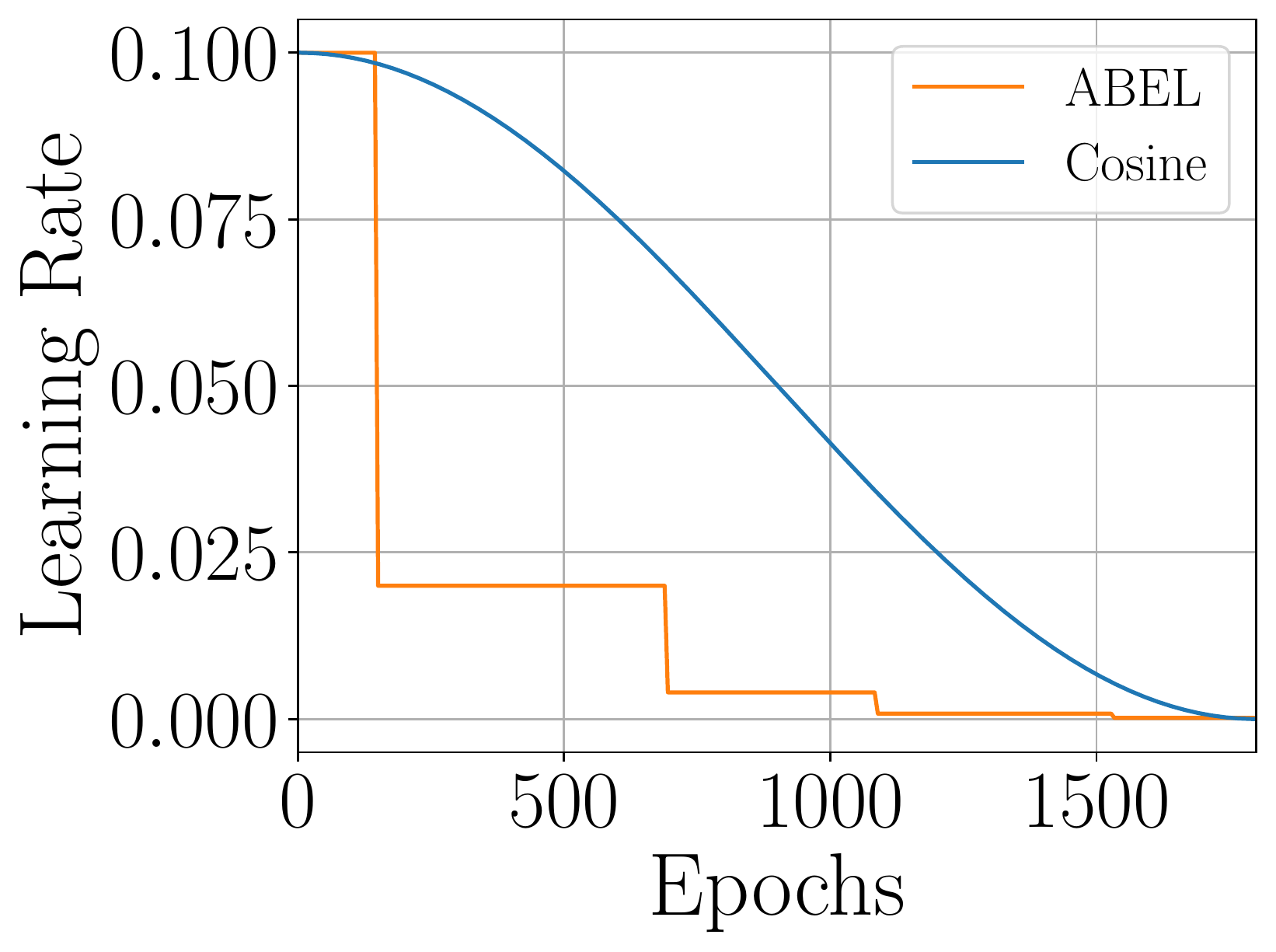}
    
  }
  \caption{Remaining training curves for experiments of table \ref{table:table1}. }
  \label{fig:more_training_curves}
\end{figure}
\clearpage
\subsection{Training curves for Table \ref{table:nobounce} experiments}
\label{SM:L20}
Here we show the training curves for some experiments in Table 2. 
\begin{figure}[ht!]
  \centering
  \subfloat[WRN 28-10 on CIFAR100]{
    \includegraphics[width=0.33 \textwidth]{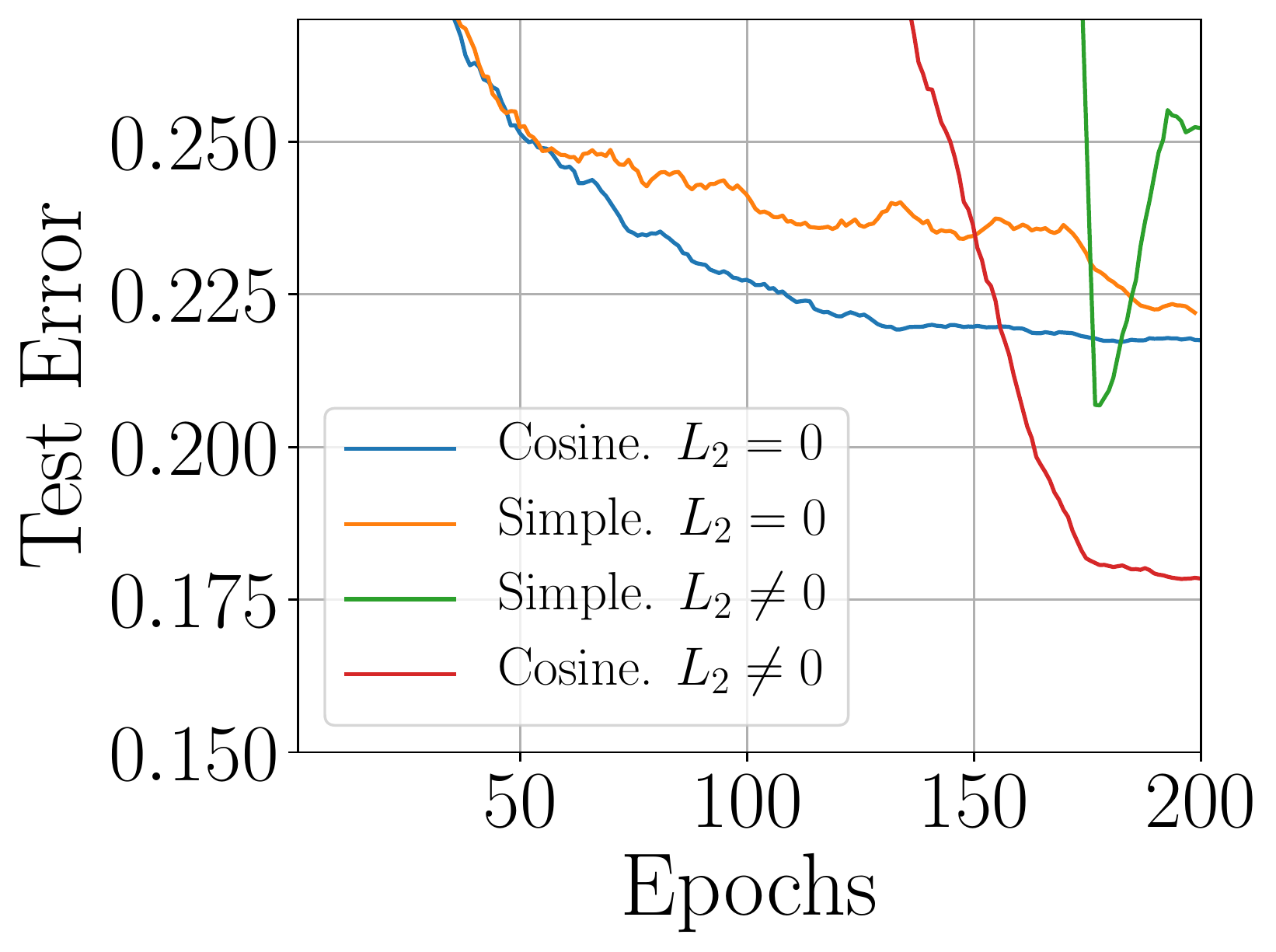}
  }
  \subfloat[WRN 28-10 on CIFAR100]{
    \includegraphics[width=0.33 \textwidth]{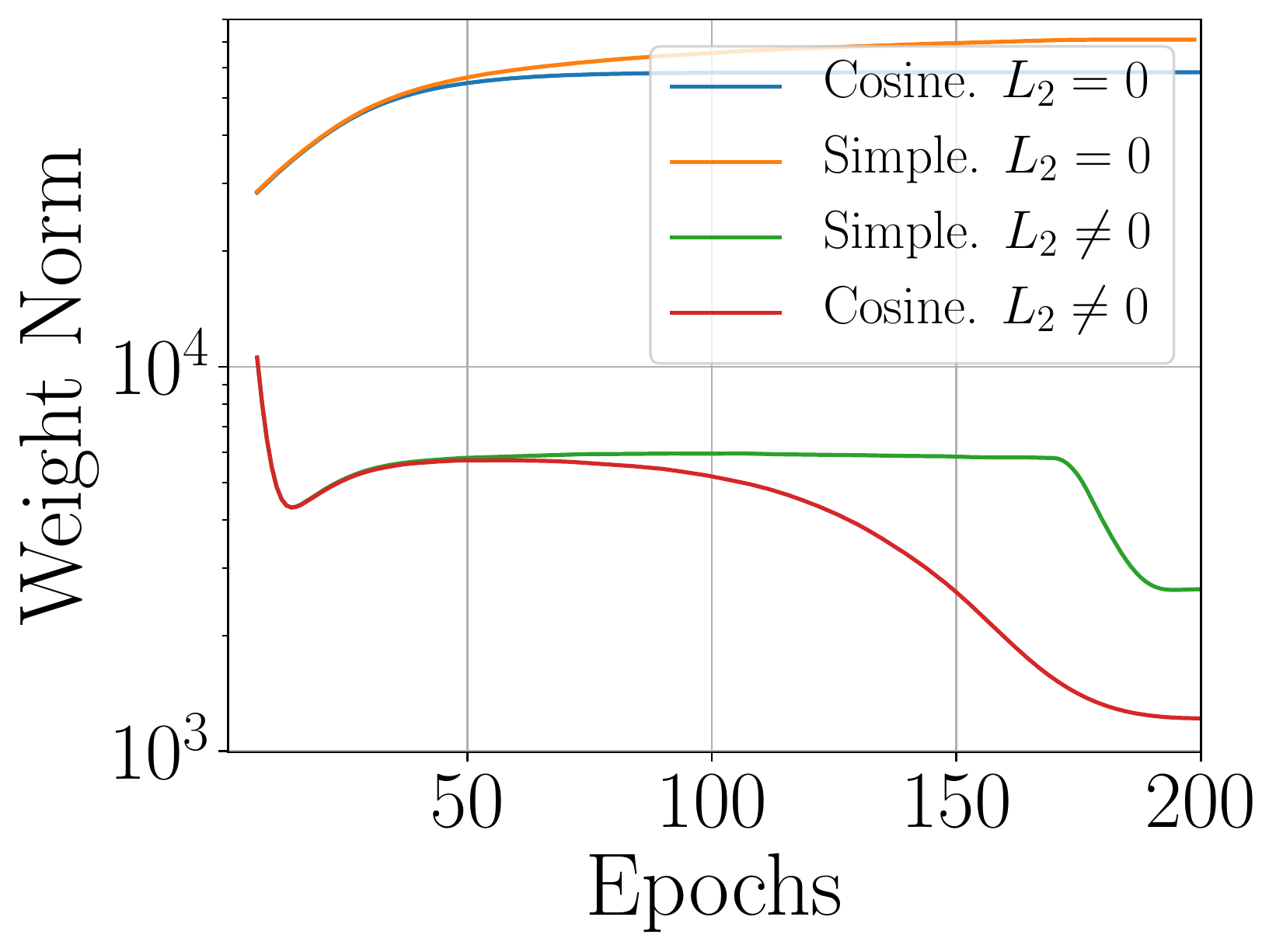}
    
  }
  \subfloat[WRN 28-10 on CIFAR100]{
    \includegraphics[width=0.33 \textwidth]{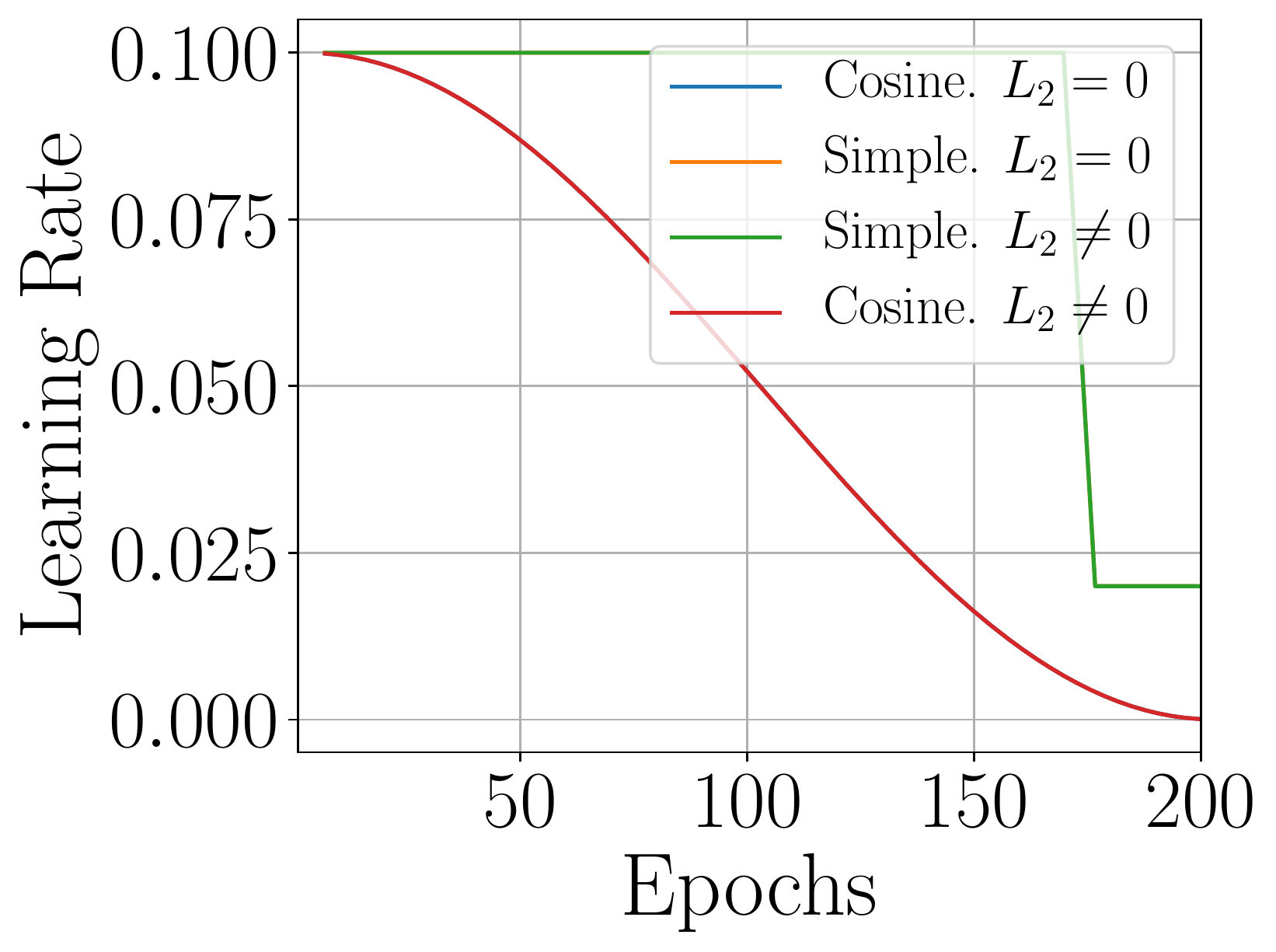}
  }
  \\
  \subfloat[Adam and Imagenet]{
    \includegraphics[width=0.33 \textwidth]{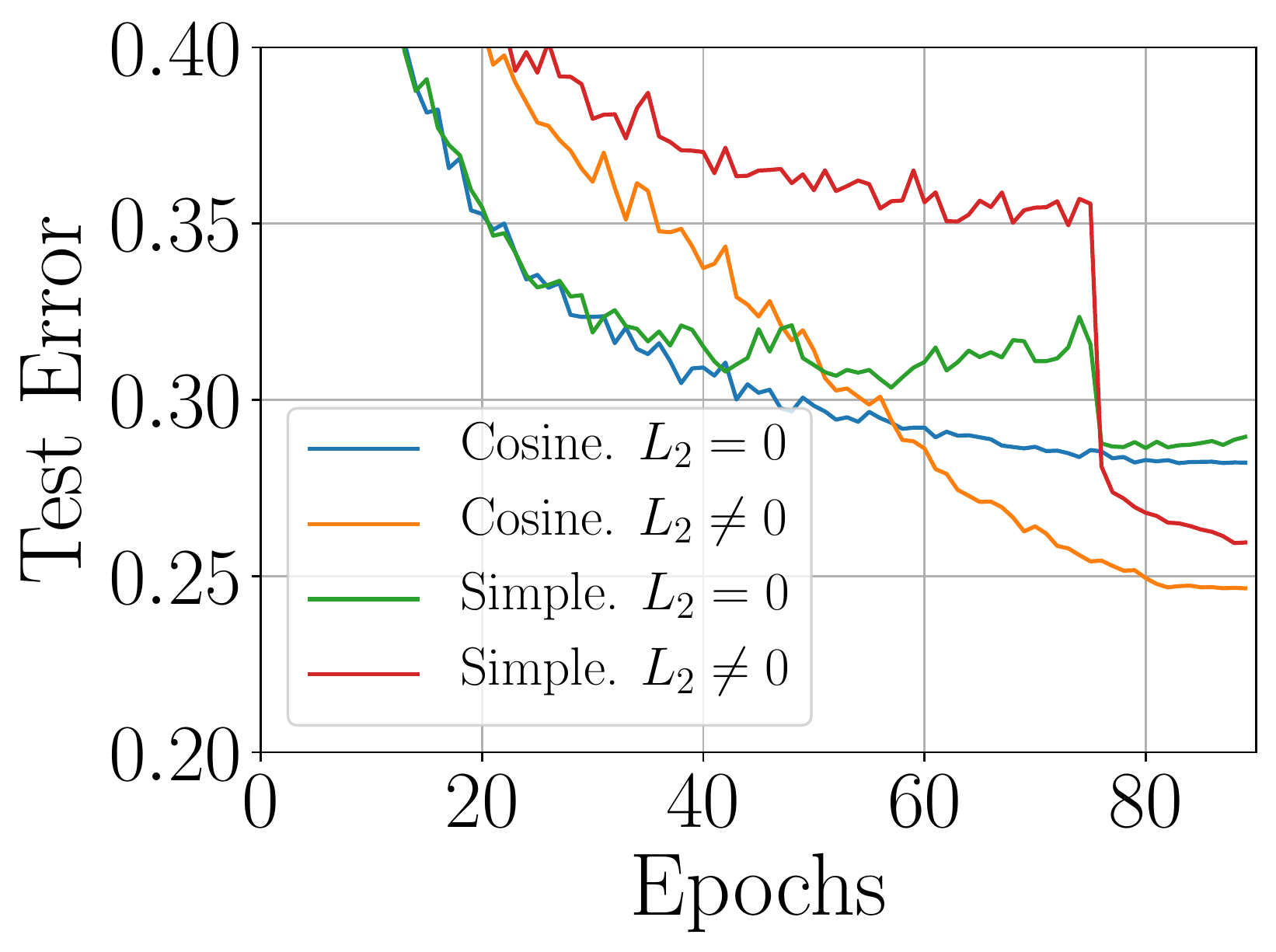}
  }
  \subfloat[Adam and Imagenet]{
    \includegraphics[width=0.33 \textwidth]{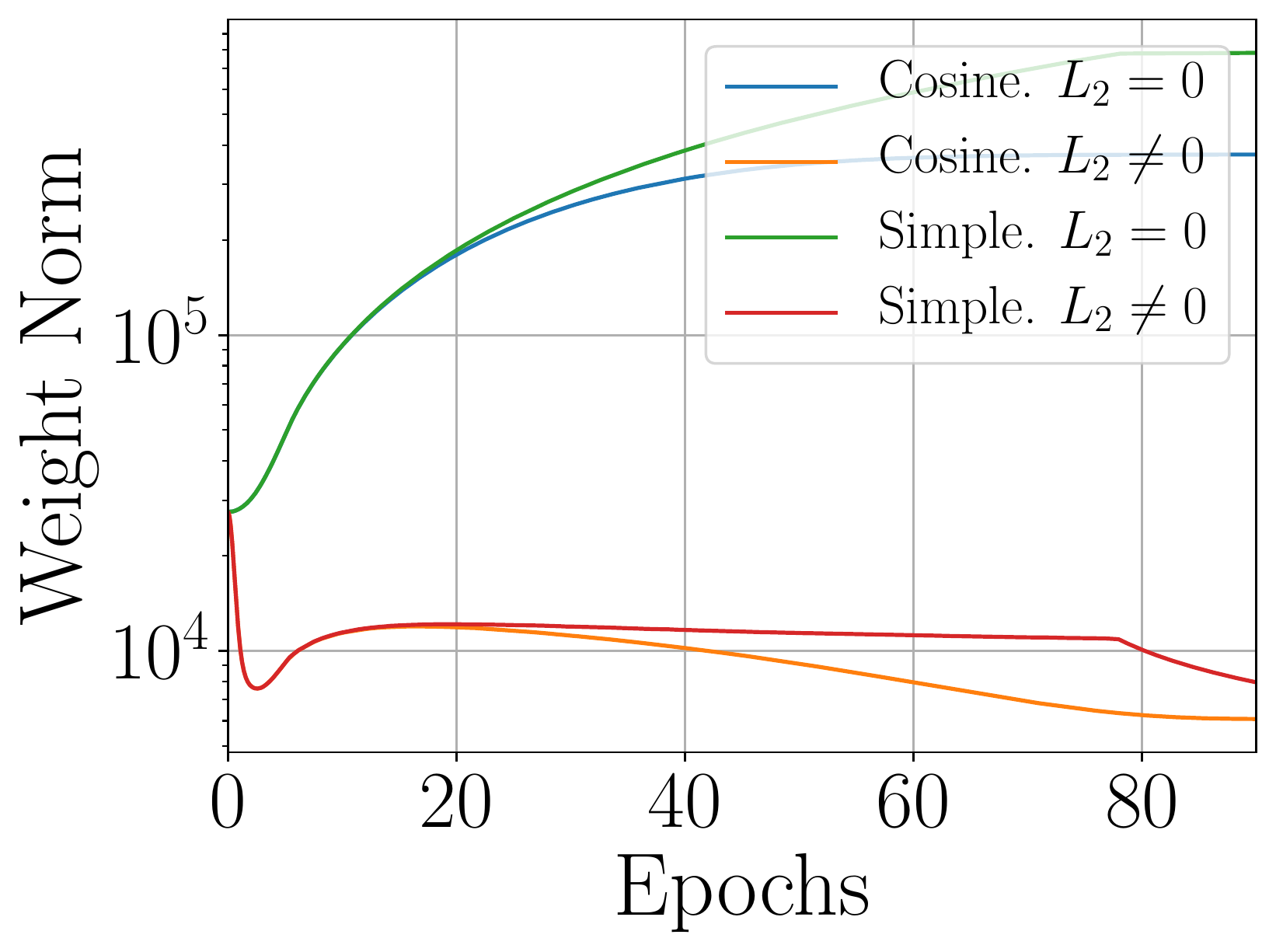}
    
  }
  \subfloat[Adam and Imagenet]{
    \includegraphics[width=0.33 \textwidth]{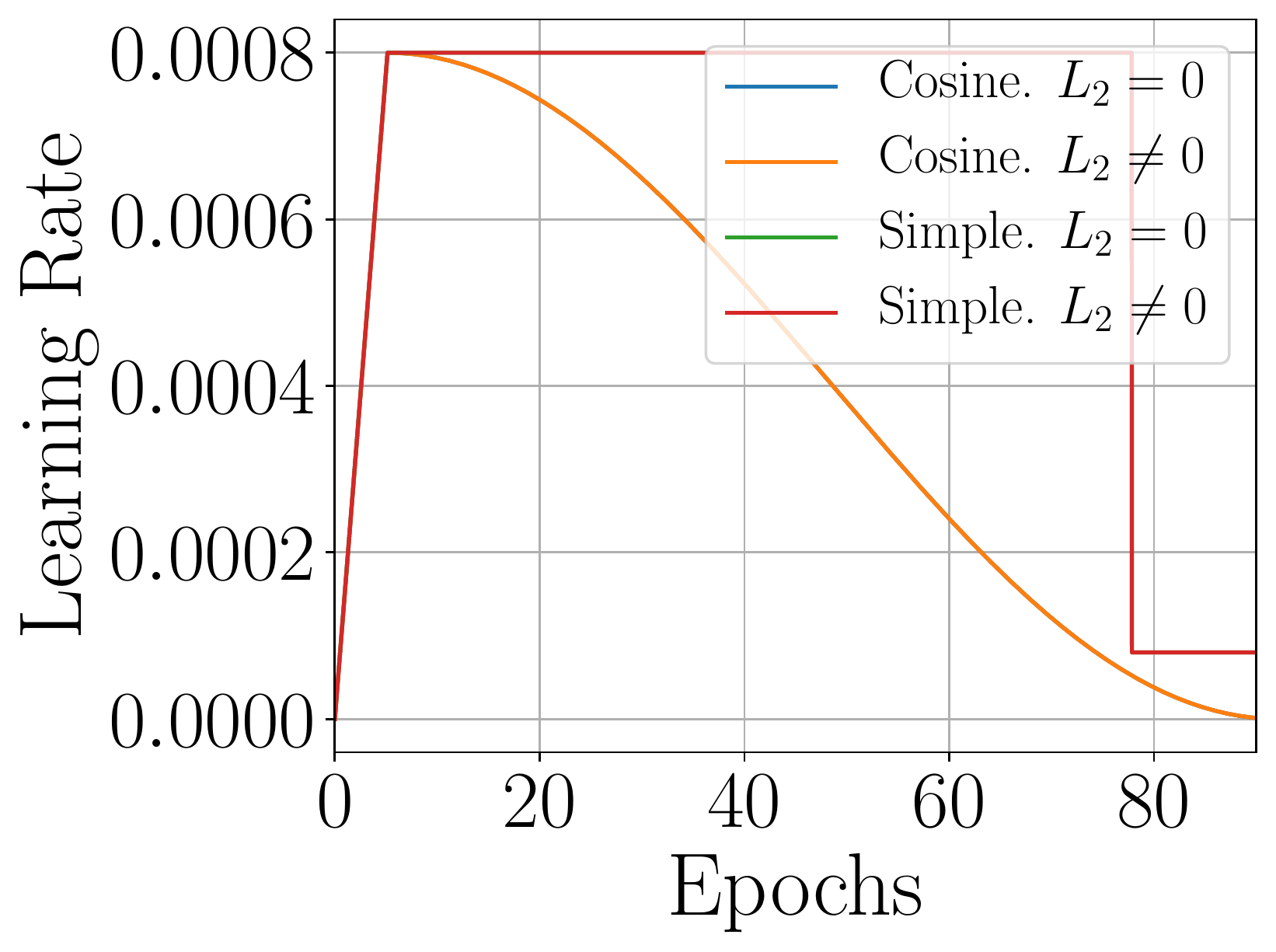}
  }

  \caption{A couple of vision tasks of table \ref{table:nobounce}. We see how in the absence of $L_2$, a non-trivial schedule does not affect the performance and the weight norm is usually monotonically increasing. d,e,f) Resnet-50 with Adam on Imagenet. We also see how weight norm bouncing also occurs despite of Adam having an implicit schedule.}
  \label{fig:L20_training_curves}
\end{figure}

\begin{figure}[ht!]
  \centering
  \subfloat[Transformer on En-De WMT]{
    \includegraphics[width=0.33 \textwidth]{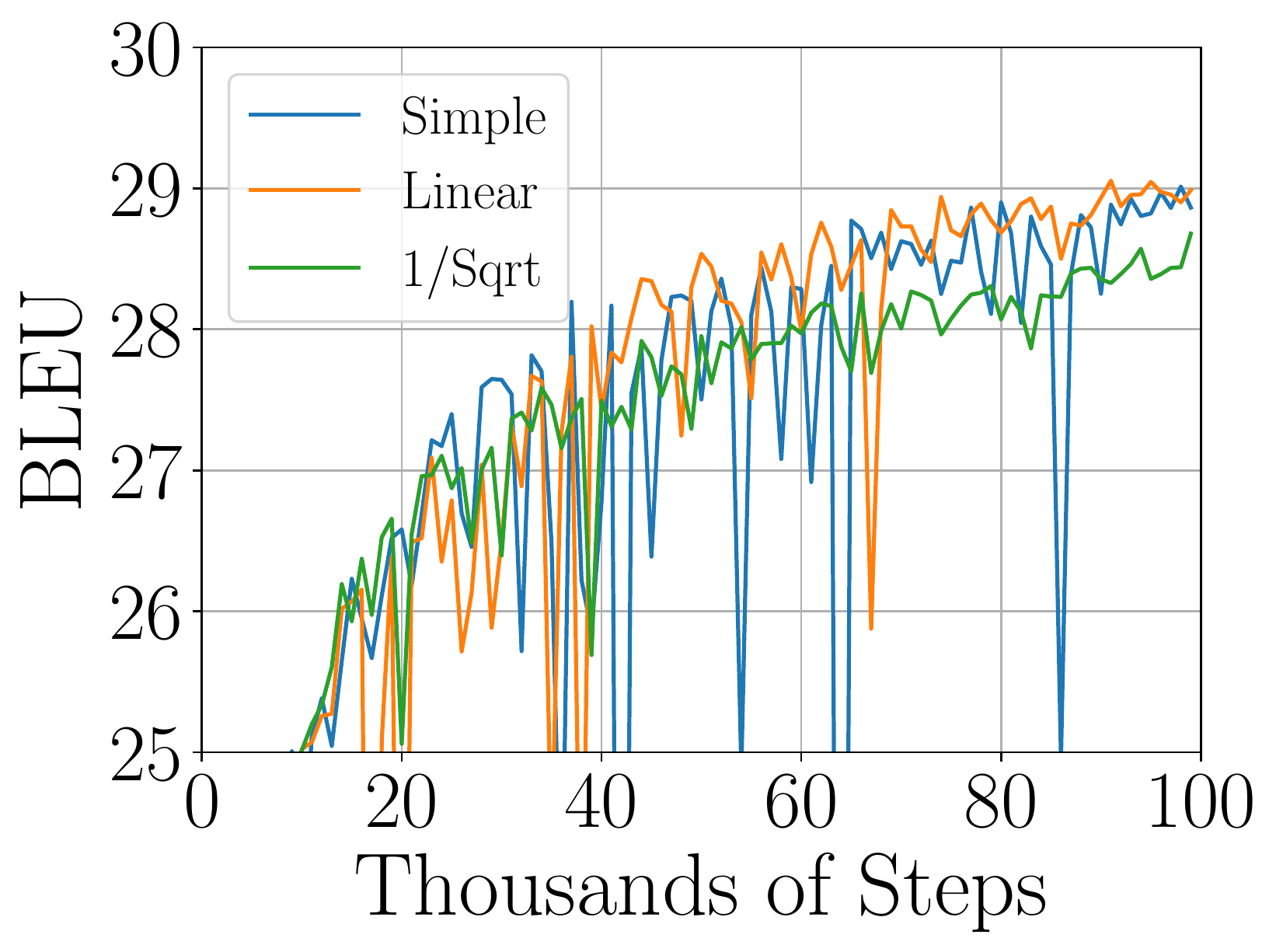}
  }
  \subfloat[Transformer on En-De WMT]{
    \includegraphics[width=0.33 \textwidth]{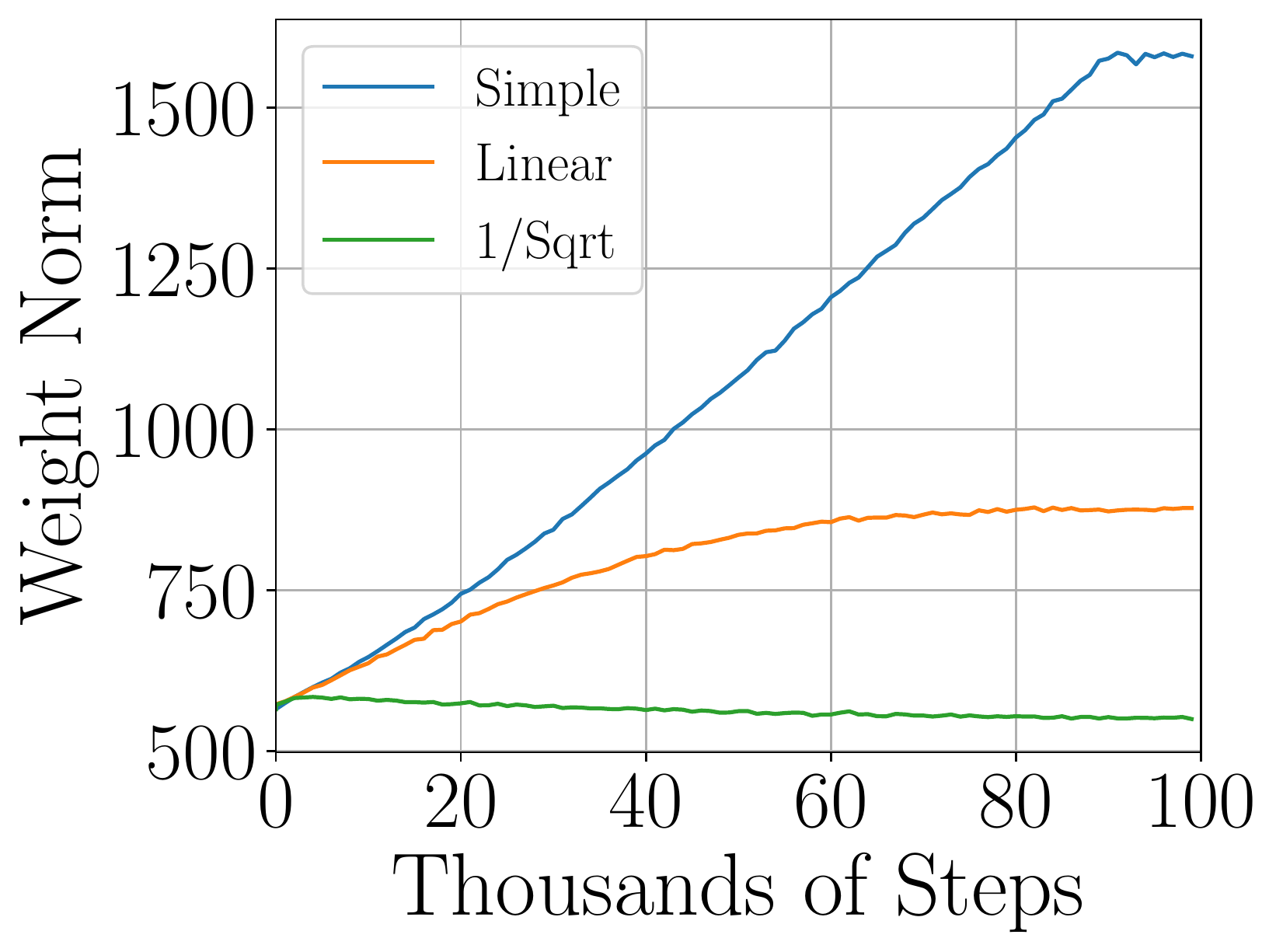}
    
  }
  \subfloat[WRN 28-10 on CIFAR100]{
    \includegraphics[width=0.33 \textwidth]{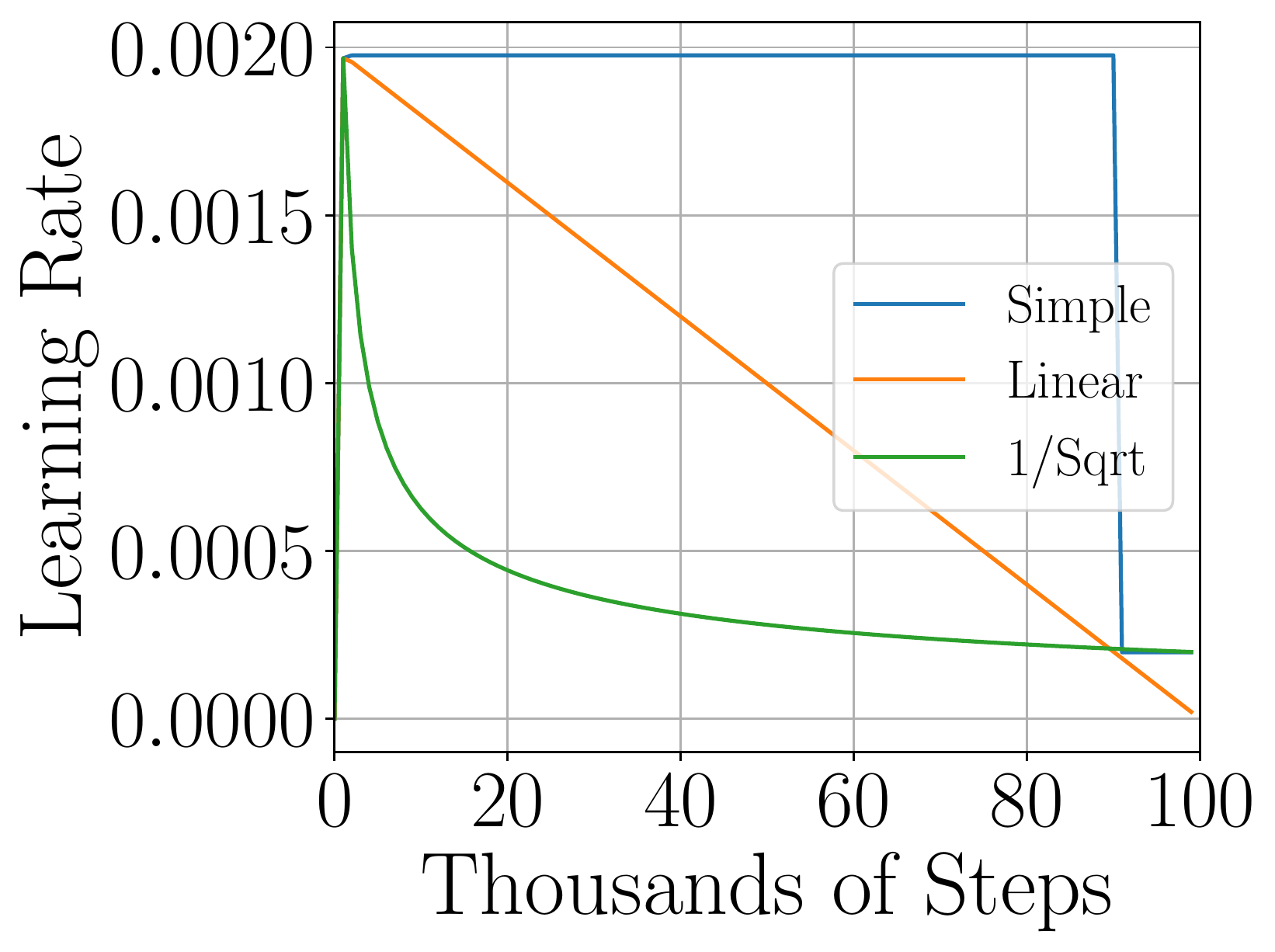}
  }

  \caption{Training curves for Transformer trained on the English-German translation task WMT'14. We see how there is no significant difference between differnet learning rate schedules.}
  \label{fig:NLP_training_curves}
\end{figure}

\begin{figure}[ht!]
  \centering
  \subfloat[PPO Pong]{
    \includegraphics[width=0.33 \textwidth]{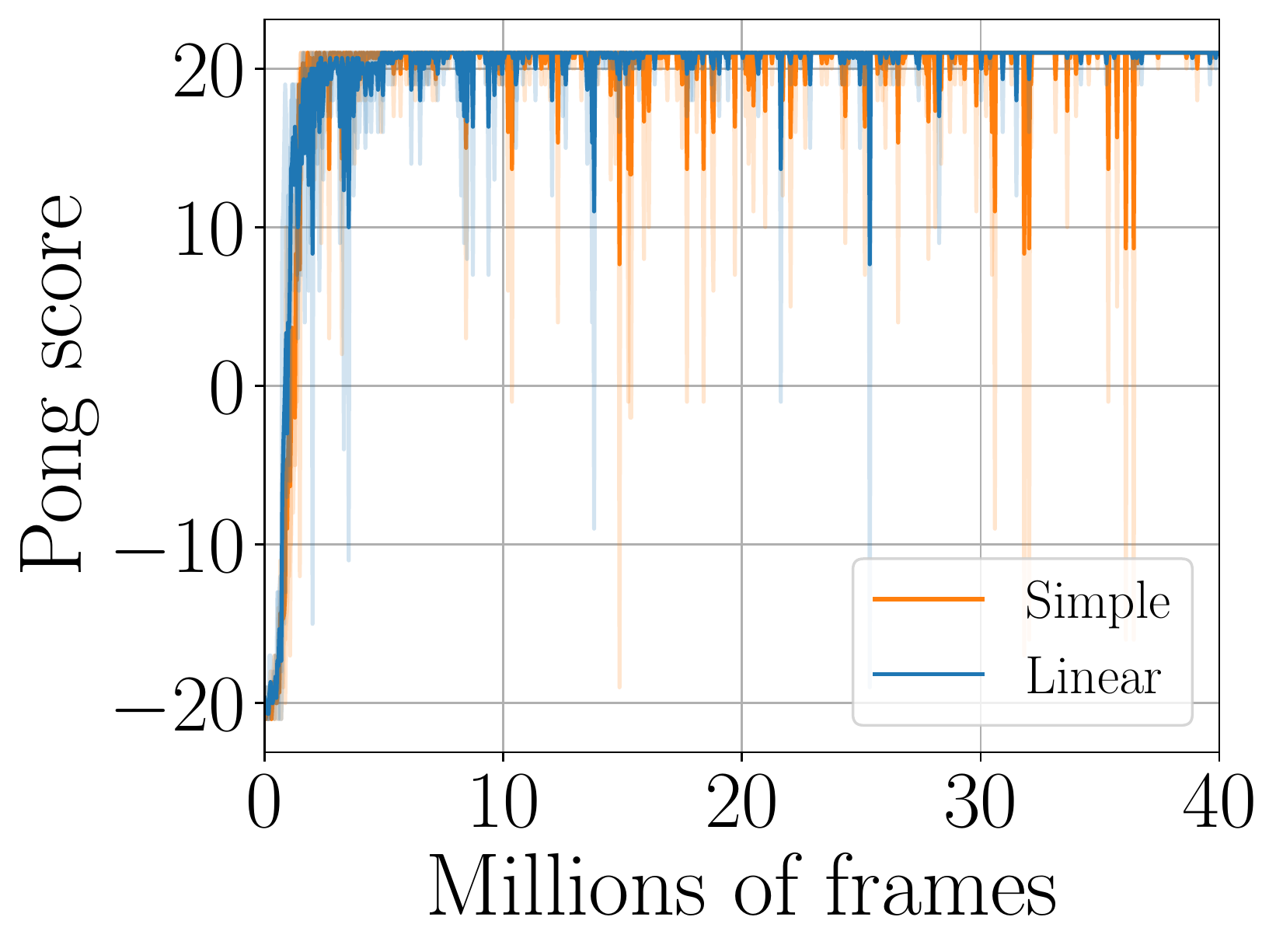}
  }
  \subfloat[PPO Qbert]{
    \includegraphics[width=0.33 \textwidth]{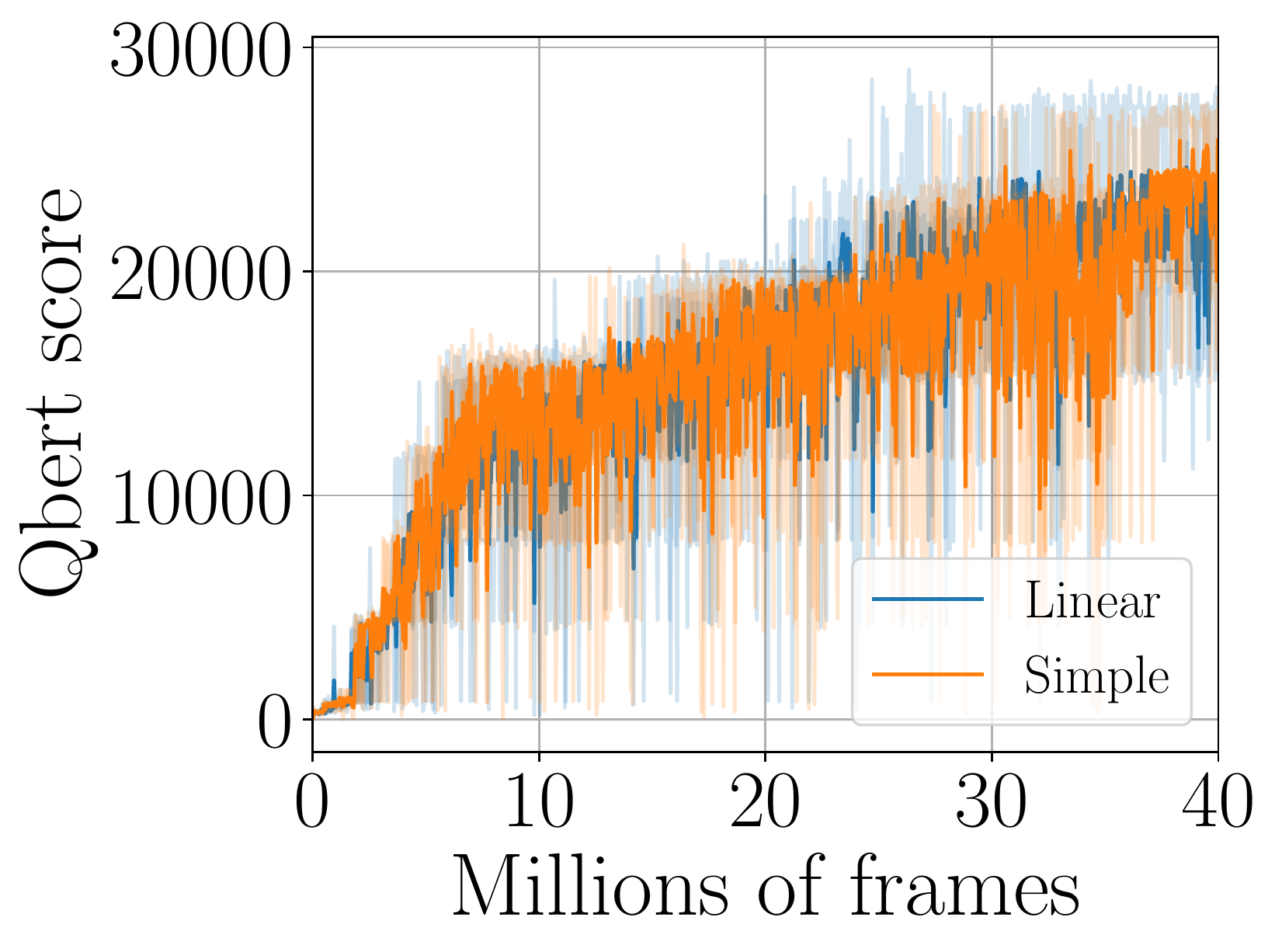}
    
  }
  \subfloat[PPO Seaquest]{
    \includegraphics[width=0.33 \textwidth]{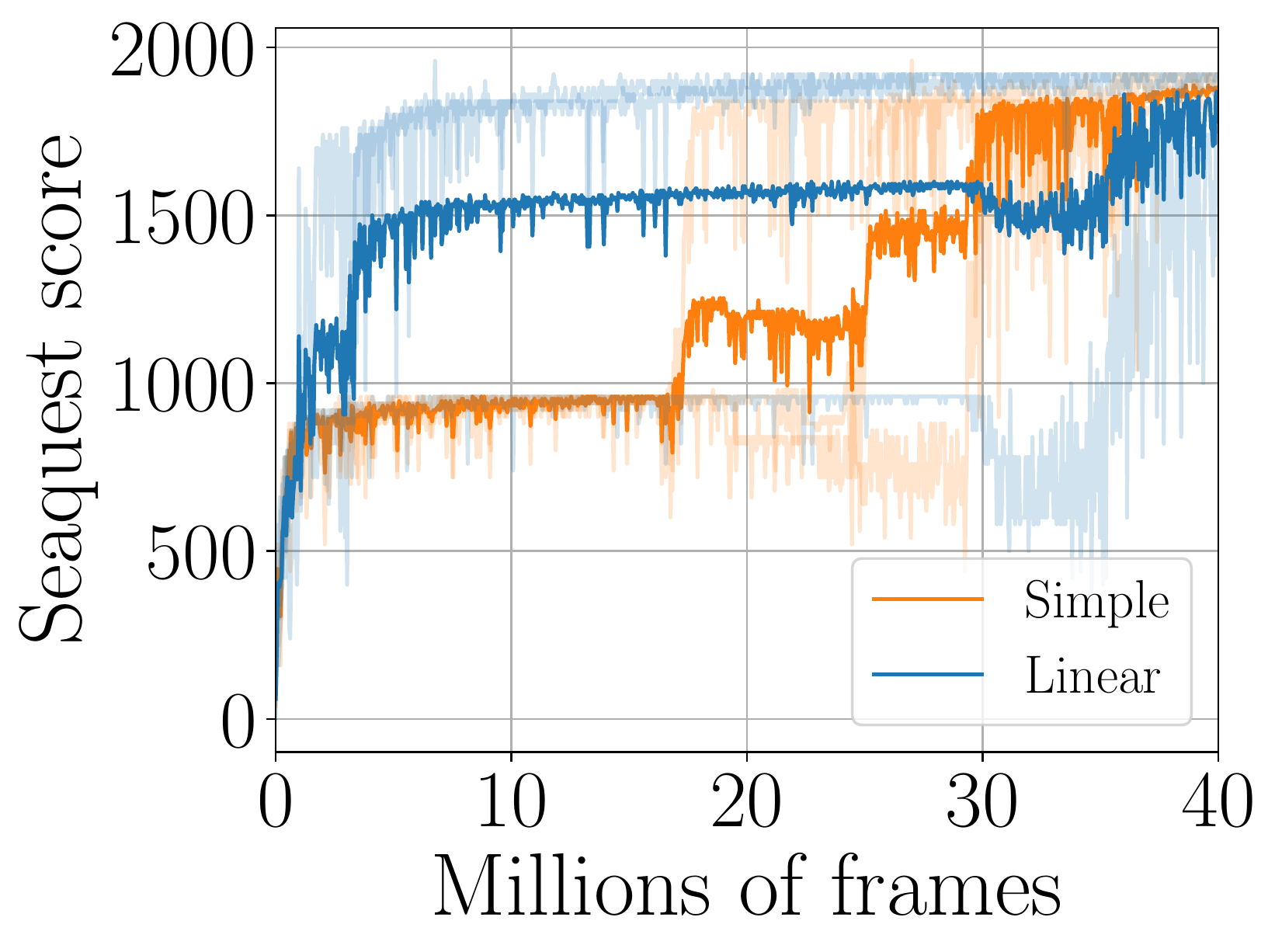}
  }
\\
  \subfloat[PPO Pong]{
    \includegraphics[width=0.33 \textwidth]{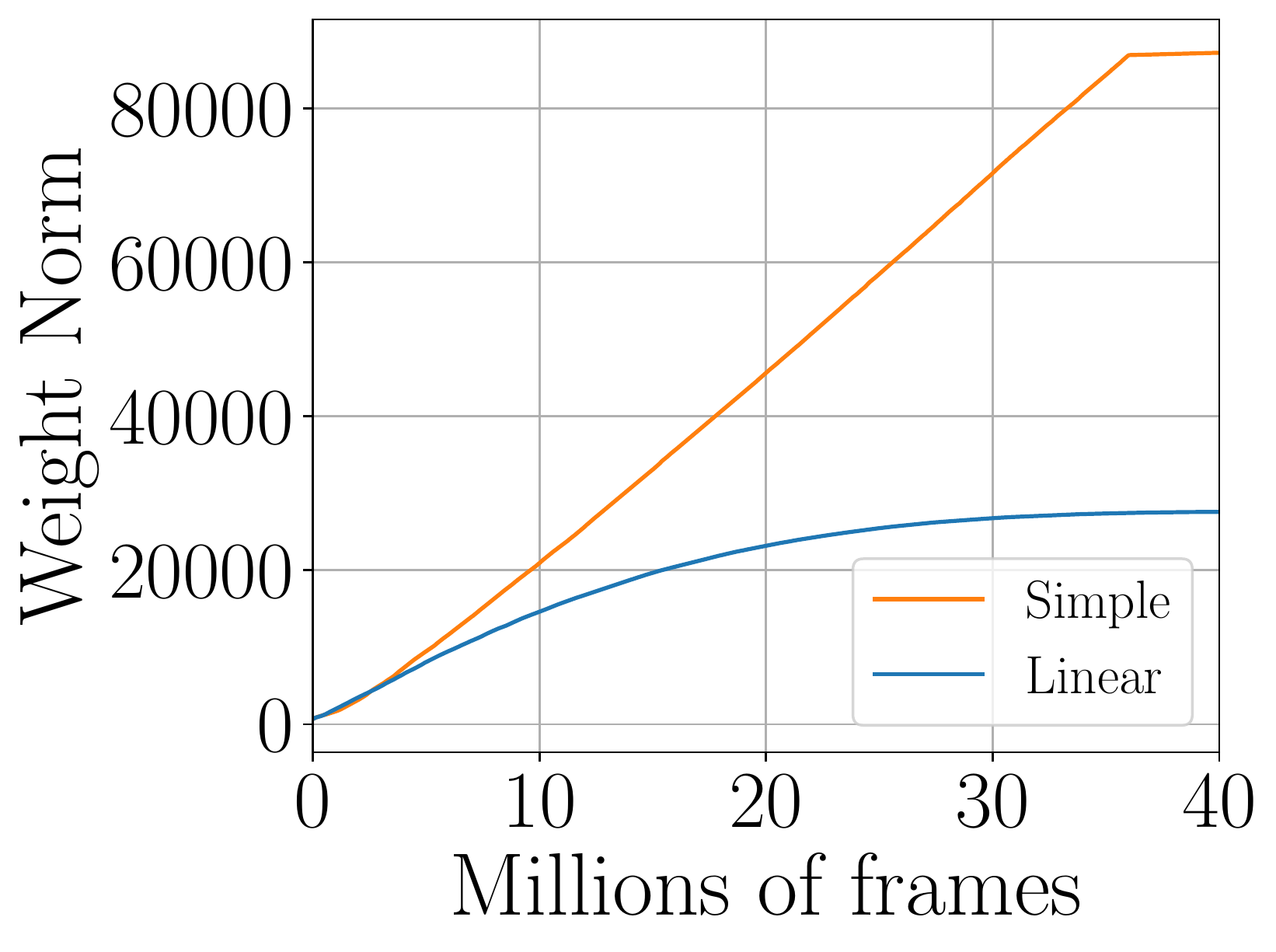}
  }
  \subfloat[PPO Qbert]{
    \includegraphics[width=0.33 \textwidth]{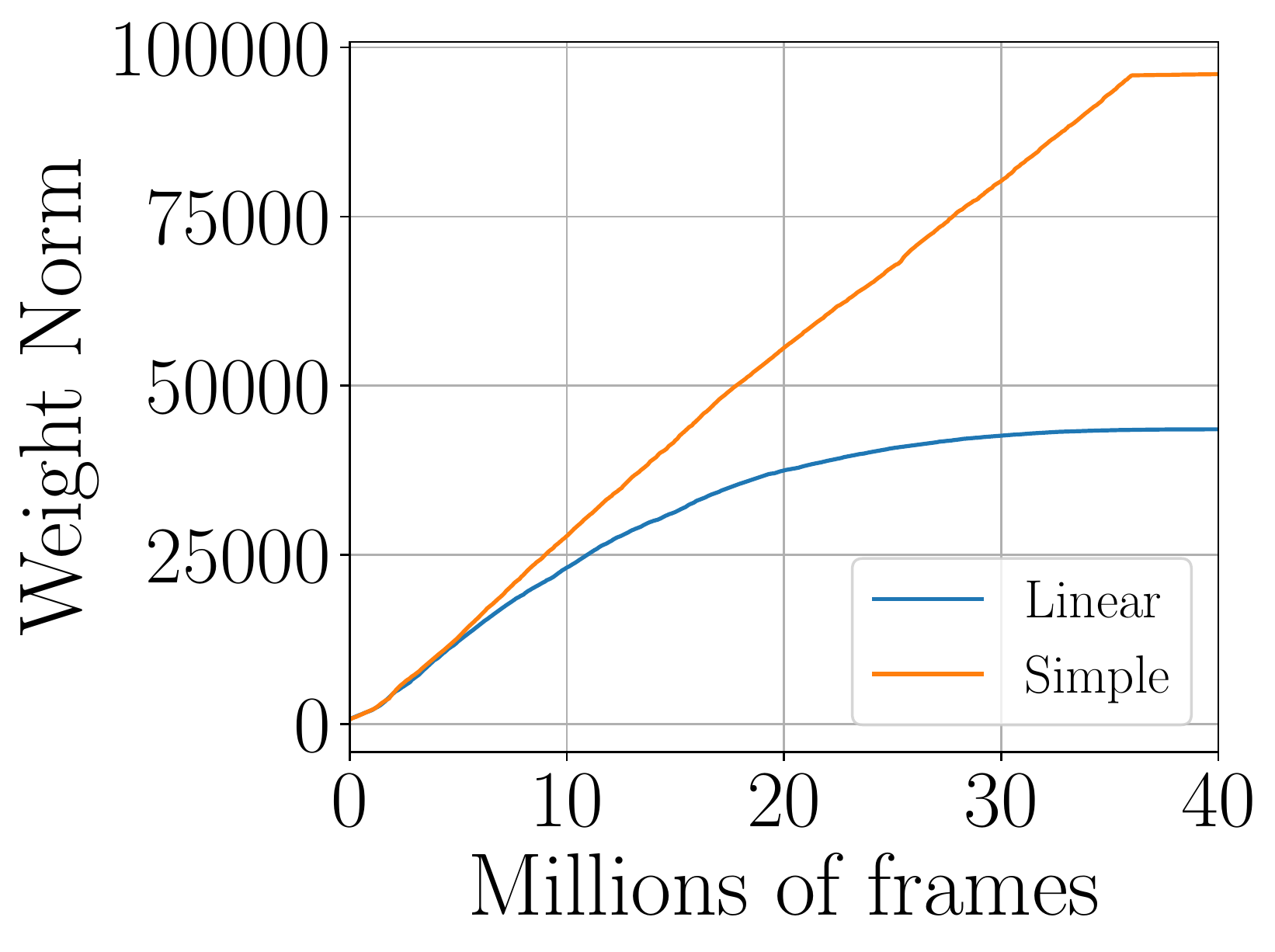}
    
  }
  \subfloat[PPO Seaquest]{
    \includegraphics[width=0.33 \textwidth]{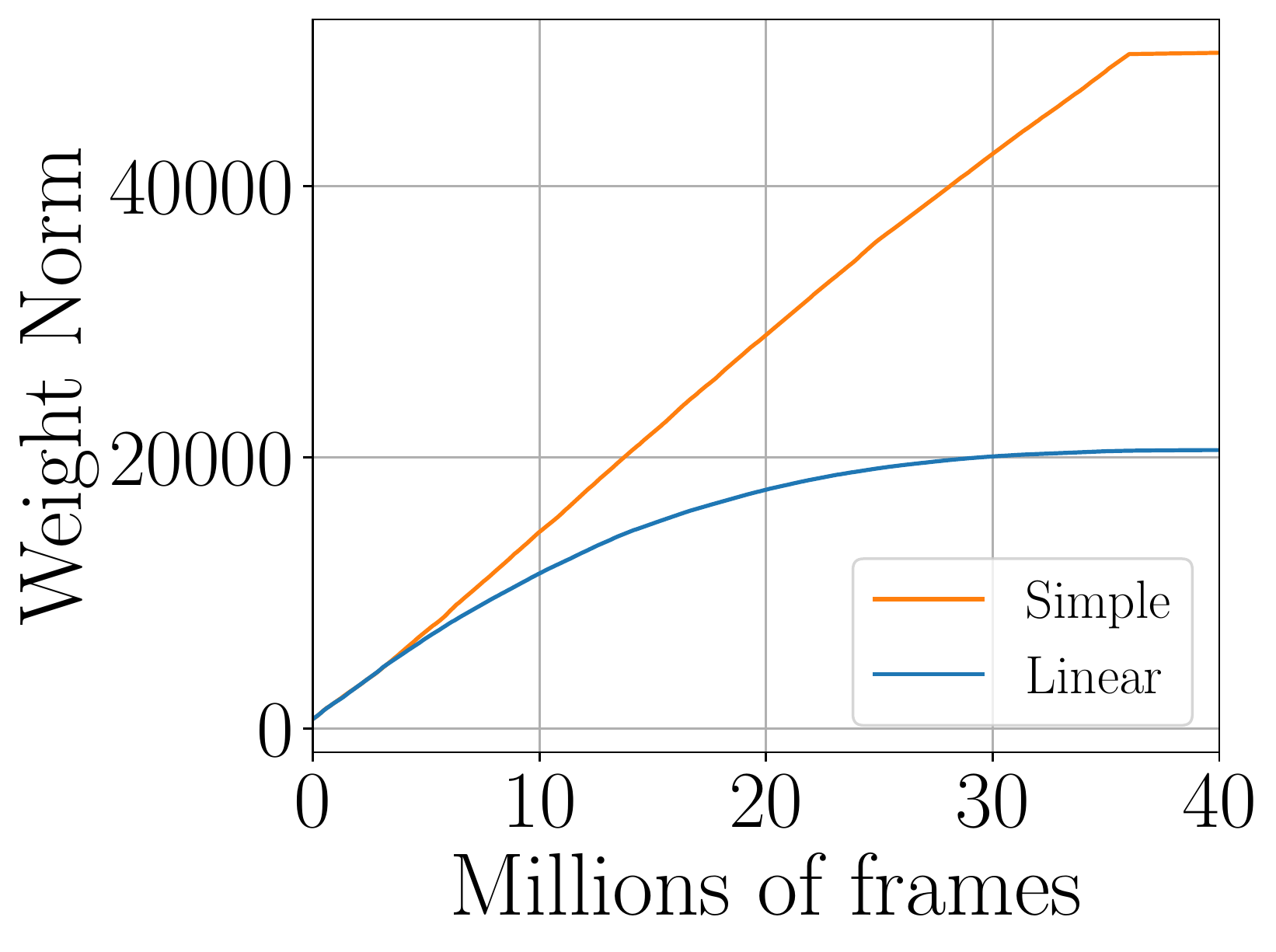}
  }
  \caption{Game score of different RL games for the two considered schedules. Mean score over three runs and individual scores in lighter color. Simple is just decaying the base learning rate by a factor of $10$ at $90\%$ of training and linear is linear decay.}
  \label{fig:PPO_training_curves}
\end{figure}

\clearpage
\subsection{Training curves for large learning rate experiment}
\label{SM:largelr}
We show the training curves for the ImageNet experiment with a large learning rate of $16$ of section \ref{sec:abel}, figure \ref{fig:comparison}. We see that Cosine and Stepwise schedules are basically not learning until the learning rate decays significantly, however, ABEL adapts quickly to the large learning rate and decays it fast, yielding better performance. 
\begin{figure}[ht!]
  \centering
  \subfloat[Resnet-50 on ImageNet]{
    \includegraphics[width=0.31 \textwidth]{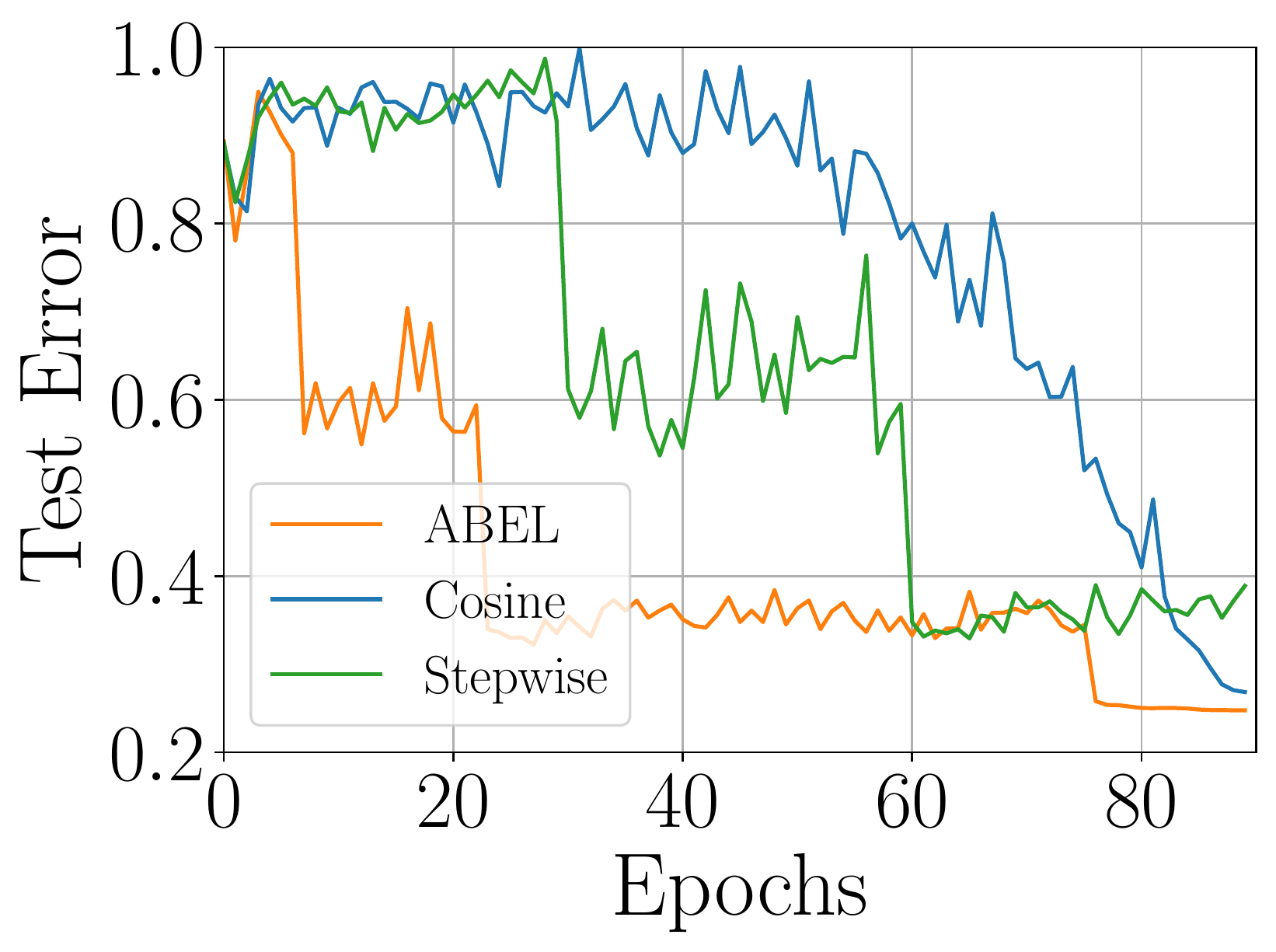}
  }
  \subfloat[Resnet-50 on ImageNet]{
    \includegraphics[width=0.31 \textwidth]{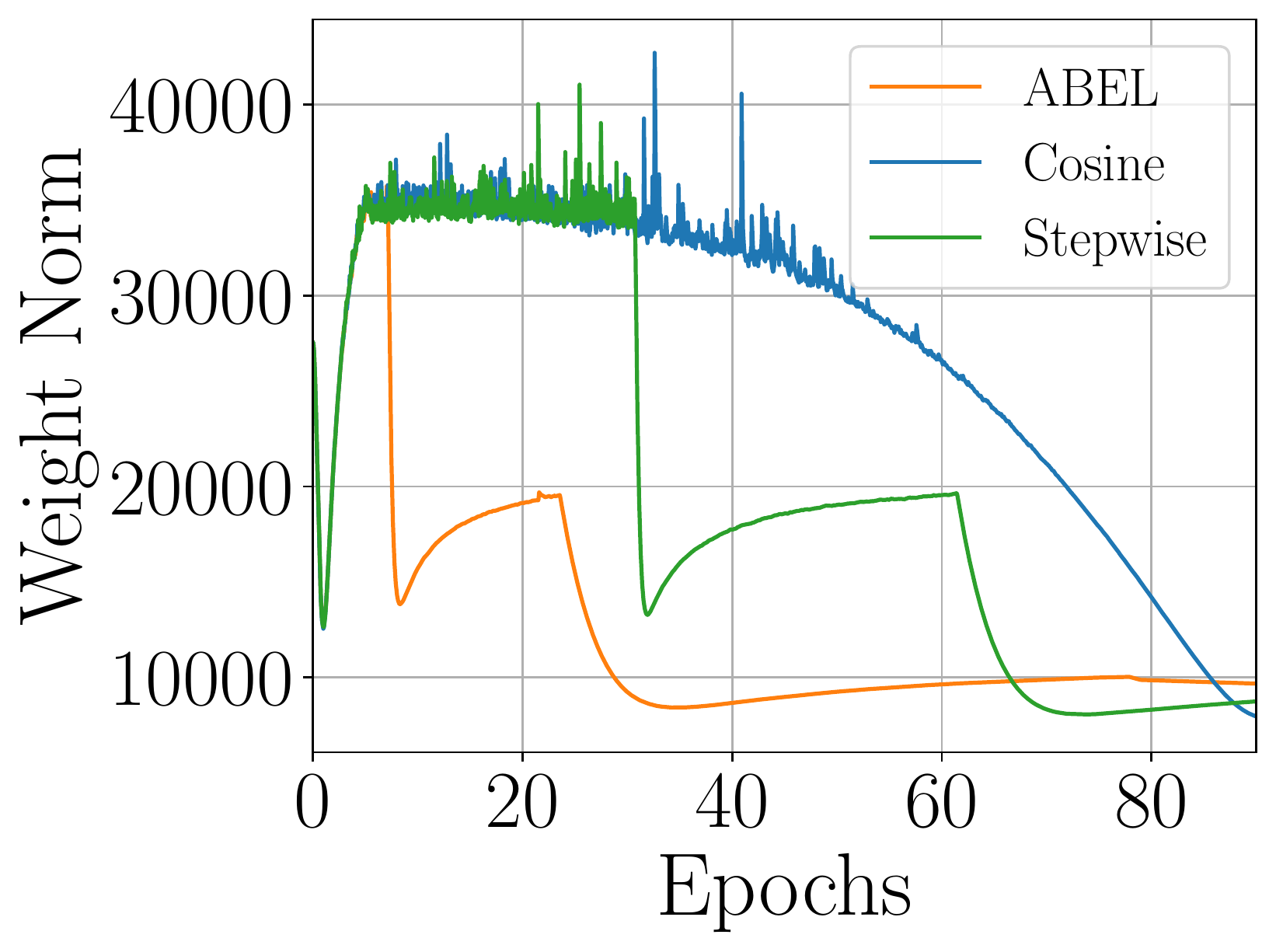}
    
  }
  \subfloat[Resnet-50 on ImageNet]{
    \includegraphics[width=0.31 \textwidth]{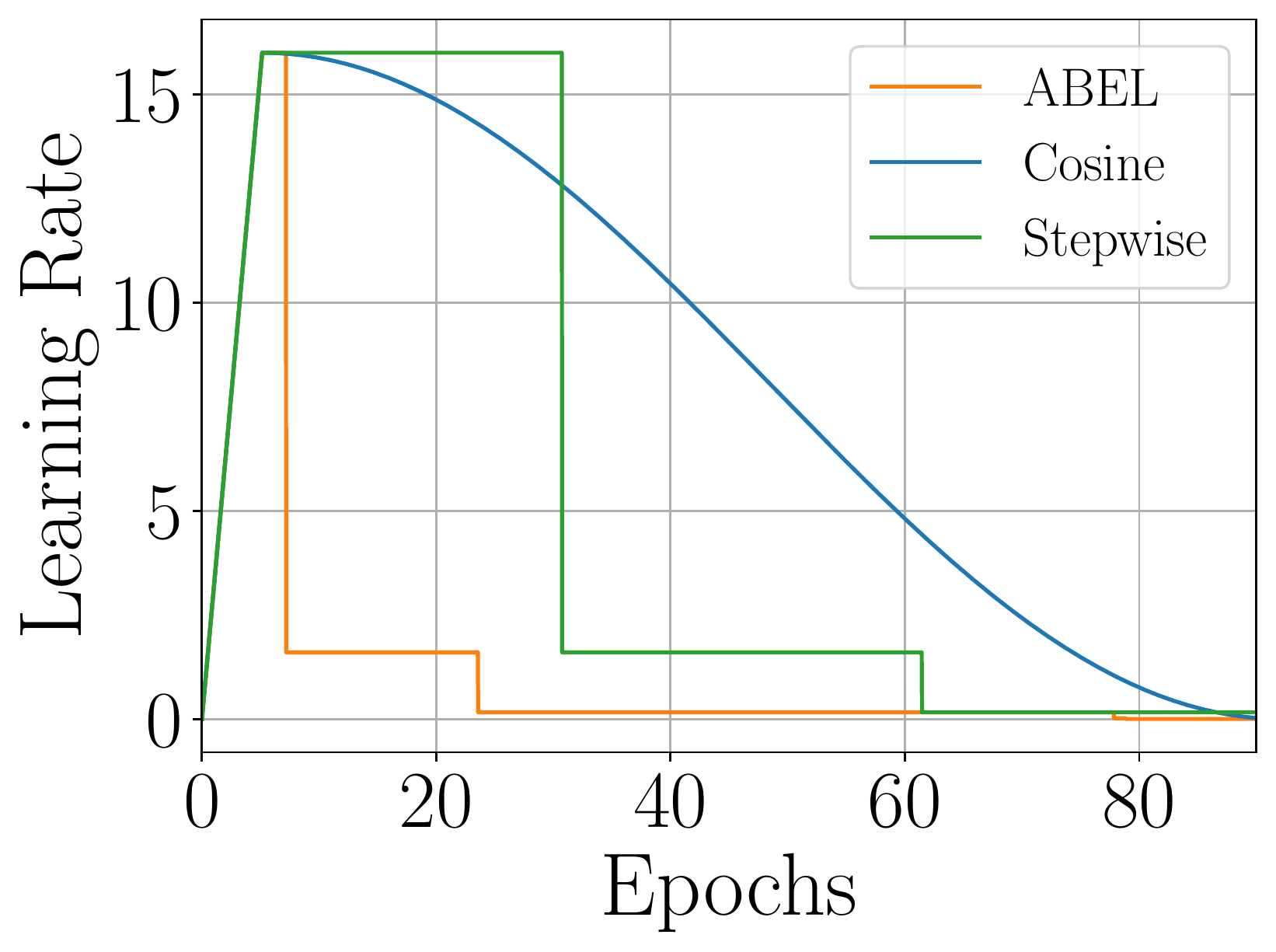}
  }

  \caption{Training curves for the Imagenet experiment of section \ref{sec:abel}, figure \ref{fig:comparison} with learning rate $16$. }

\end{figure}

\clearpage
\section{Derivation of equations of section \ref{sec:understand}}

\paragraph{Weight dynamics equations.}
Equation \ref{eq:weight_norm} follows directly from the SGD update:
\begin{equation}
    \Delta w_{t+1}=w_{t+1}-w_t= -\eta g_t - \eta \lambda w_t \\
\end{equation}
\begin{eqnarray}
    & \Delta |w_{t+1}|^2 \equiv |w_{t+1}|^2-|w_t|^2 = \Delta w_{t-1}.\Delta w_{t-1}+ 2 w_{t-1} \Delta w_{t-1} &  =  \eta^2 |g_t|^2- (2-\eta \lambda) \eta \lambda |w_t|^2-2(1-\eta \lambda)   g_t \cdot w_t  \nonumber\\
    & & \approx \eta^2 |g_t|^2-2 \eta \lambda |w_t|^2-2 \eta  g_t \cdot w_t ~~ 
\end{eqnarray}

The terms that we are dropping are subleading for our purposes because $\eta \lambda \ll 1$ practically. We expect these terms to be only relevant if the dominant terms cancel each other: $\eta^2 |g_t|^2-2 \eta \lambda |w_t|^2-2 \eta  g_t \cdot w_t \sim 0$

\paragraph{$g \cdot w=0$ for scale invariant layers.} A scale invariant layer is such that its network function satisfies $f(\alpha w)=f(w)$, the bare loss only depends on the weights through the network function. If we divide the weight into its norm and its direction: $w_a=|w|\hat{w}_a$ and use $\dfrac{d L}{d |w|}=0$, we get that:
\begin{equation}
   0= |w| \frac{d L}{d |w|}= |w| \sum_a \frac{dL}{d w_a} \frac{d w_a}{d |w|} =  \sum_a \frac{dL}{d w_a} |w| \hat{w}_a = g \cdot w
\end{equation}

\paragraph{For scale invariant networks without $L_2$, the change in the weights becomes smaller as the weight norm equilibrates.} Using the SGD equation we get that:
\begin{eqnarray}
 \cos \text{angle}(w_{t+1},w_t) & = & \dfrac{w_{t+1}\cdot w_t}{|w_{t+1}||w_t|}  =_{\Delta w_t=-\eta g} \frac{|w_t|}{|w_{t+1}|} \nonumber \\
 \sin \text{angle}(w_{t+1},w_t) & = &\sqrt{1- \cos^2 \text{angle}}  = \sqrt{\frac{\Delta |w_t|^2}{|w_{t+1}|^2}}
\end{eqnarray}
As $\Delta |w|^2 \rightarrow 0$, $\text{angle}(w_{t+1},w_t) \sim \sin \text{angle}(w_{t+1},w_t) \rightarrow 0$. 

\clearpage
\ifarxiv
\section{Flax implementation of ABEL}
\label{SM:flax}
For completeness we include our Flax implementation of ABEL.
\lstset{style=mystyle}

\begin{lstlisting}[language=Python]
# Copyright 2021 Google LLC.
# SPDX-License-Identifier: Apache-2.0

class ABELScheduler():
  """Implementation of ABEL scheduler."""

  def __init__(self,
              num_epochs: int,
              learning_rate: float,
              steps_per_epoch: int,
              decay_factor: float,
              train_fn: callable,
              warmup: int = 0):

    self.num_epochs = num_epochs
    self.learning_rate = learning_rate
    self.steps_per_epoch = steps_per_epoch
    self.decay_factor = decay_factor
    self.train_fn = train_fn
    self.warmup = warmup
    self.learning_rate_fn = self.get_learning_rate_fn(self.learning_rate)
    self.weight_list = []
    self.reached_minima = False
    self.epoch = 0

  def get_learning_rate_fn(self, lr):
    """Outputs a simple decay learning rate function from base learning rate."""
    lr_fn = flax.training.lr_schedule.create_stepped_learning_rate_schedule(
        lr, self.steps_per_epoch // jax.host_count(),
        [[int(self.num_epochs * 0.85), self.decay_factor]])
    if self.warmup:
      warmup_fn = lambda step: jax.numpy.minimum(
          1., step / self.steps_per_epoch / self.warmup)
    else:
      warmup_fn = lambda step: 1

    return lambda step: lr_fn(step) * warmup_fn(step)

  def update(self, step_fn, weight_norm):
    """Optimizer update rule for ABEL Scheduler. This is basically Algorithm 1."""
    self.weight_list.append(weight_norm)
    self.epoch += 1

    if len(self.weight_list) < 3:
      return step_fn

    if (self.weight_list[-1] - self.weight_list[-2]) * (
        self.weight_list[-2] - self.weight_list[-3]) < 0:
      if self.reached_minima:
        self.reached_minima = False
        self.learning_rate *= self.decay_factor
        step_fn = self.update_train_step(self.learning_rate)
      else:
        self.reached_minima = True

    return step_fn

  def update_train_step(self, learning_rate):
    learning_rate_fn = self.get_learning_rate_fn(learning_rate)
    return self.train_fn(learning_rate_fn=learning_rate_fn)
\end{lstlisting}
The main reason why this is a Flax implementation is because the \texttt{update\_step} takes a \texttt{train\_step} function and outputs a \texttt{train\_step} function. 
This can be easily added to the Flax examples/baselines, we only have to make two modifications.
Before starting the training loop, we initiate ABEL \footnote{Note that \texttt{ABELSchedule} takes \texttt{train\_step\_fn} as an argument. This is a function that takes a \texttt{learning\_rate\_fn} and outputs a \texttt{train\_step} (ie \texttt{train\_step\_fn = lambda lr\_fn:  functools.partial(
          train\_step,
          learning\_rate\_fn=lr\_fn)}}: 
\begin{lstlisting}[language=Python, numbers=none]
# Copyright 2021 Google LLC.
# SPDX-License-Identifier: Apache-2.0

# Before training starts.
scheduler = ABELScheduler(num_epochs, base_learning_rate, steps_per_epoch = steps_per_epoch, decay_factor=decay_factor, train_fn = train_step_fn)
learning_rate_fn = scheduler.learning_rate_fn
\end{lstlisting}

Only remaining modification is to add the ABEL update rule at the end of each epoch which takes the current \texttt{train\_step} function and mean weight norm and returns a (possibly updated) \texttt{train\_step} function. ABEL will update the optimizer if the learning rate has to be decayed. 
\begin{lstlisting}[language=Python, numbers=none]
# Copyright 2021 Google LLC.
# SPDX-License-Identifier: Apache-2.0

# At the end of each epoch.
train_step = scheduler.update(train_step, weight_norm)
\end{lstlisting}
\fi
\end{document}